\documentclass[preprint,12pt]{elsarticle}

\usepackage{amssymb}
\usepackage{amsmath}

\usepackage{lineno} 
\usepackage{enumitem}  
\usepackage{siunitx}   
\usepackage{tabularx}  
\usepackage{xcolor}
\usepackage{tabularx,booktabs}
\usepackage{url}
\usepackage{graphicx}
\usepackage{tabularx}
\usepackage{multirow}
\usepackage{colortbl}
\usepackage{caption}
\usepackage{hyperref}
\usepackage{float}     

\usepackage[ruled,vlined]{algorithm2e} 
\usepackage{algorithmic}

\usepackage{subcaption}

\begin{document}

\begin{frontmatter}

\title{EntON: Eigenentropy-Optimized Neighborhood Densification in 3D Gaussian Splatting}

\author{Miriam Jäger\corref{cor1}\fnref{label1}}
\ead{miriam.jaeger@kit.edu}
\author{Boris Jutzi\fnref{label1}}
\address[label1]{Institute of Photogrammetry and Remote Sensing (IPF), Karlsruhe Institute of Technology (KIT), Karlsruhe, Germany}
\cortext[cor1]{Corresponding author.}
\date{February 20, 2026}


\begin{abstract}  
We present a novel \emph{Eigen\textbf{ent}ropy-\textbf{o}ptimized \textbf{n}eighboorhood densification} strategy \textbf{EntON} in 3D Gaussian Splatting (3DGS) for geometrically accurate and high-quality rendered 3D reconstruction. While standard 3DGS produces Gaussians whose centers and surfaces are poorly aligned with the underlying object geometry, surface-focused reconstruction methods frequently sacrifice photometric accuracy.
In contrast to the conventional densification strategy, which relies on the magnitude of the view-space position gradient, our approach introduces a geometry-aware strategy to guide adaptive splitting and pruning. Specifically, we compute the 3D shape feature \textit{Eigenentropy} from the eigenvalues of the covariance matrix in the $k$-nearest neighborhood of each Gaussian center, which quantifies the local structural order.
These Eigenentropy values are integrated into an alternating optimization framework: During the optimization process, the algorithm alternates between (i) standard gradient-based densification, which refines regions via view-space gradients, and (ii) Eigenentropy-aware densification, which preferentially densifies Gaussians in low-Eigenentropy (ordered, flat) neighborhoods to better capture fine geometric details on the object surface, and prunes those in high-Eigenentropy (disordered, spherical) regions.
We provide quantitative and qualitative evaluations on two benchmark datasets: small-scale DTU dataset and large-scale TUM2TWIN dataset, covering man-made objects and urban scenes.
Experiments demonstrate that our Eigenentropy-aware alternating densification strategy improves geometric accuracy by up to 33\% and rendering quality by up to 7\%, while reducing the number of Gaussians by up to 50\% and training time by up to 23\%.
Overall, EnTON achieves a favorable balance between geometric accuracy, rendering quality and efficiency by avoiding unnecessary scene expansion.
\end{abstract}


\end{frontmatter}

\section{Introduction}
\label{sec:intro}

Recent advances in 3D scene reconstruction have been largely driven by the introduction of Neural Radiance Fields (NeRFs) \cite{mildenhall_et_al_2020}, which demonstrated impressive photorealistic rendering quality by learning continuous volumetric scene representations. Building on this progress, 3D Gaussian Splatting (3DGS) has emerged as an explicit scene representation that enables real-time rendering while maintaining competitive visual fidelity. Instead of implicitly encoding the scene in a neural network, 3DGS explicitly represents the scene where geometry is present by using a set of 3D Gaussians, each parameterized by mean (center), scale, rotation, color, and opacity. During the optimization process, the Gaussians are refined and adapted as they grow, shrink, and adjust in their color and opacity, to minimize the photometric error.

Despite its strong rendering performance, 3DGS can suffer from over-reconstruction issues \cite{absgs} or oversized Gaussians \cite{microsplatting}, particularly in scenes containing high-frequency geometric details. Such effects often lead to blurred areas in the rendered image, highlighting the importance of splitting big Gaussians to allow good reconstruction \cite{3DGS}. Moreover, it is evident that neither the centers nor the surfaces of Gaussians in 3DGS directly correspond to object surfaces, which makes their \textit{direct} use for point cloud and mesh reconstruction impractical. 
Although several splatting-based surface reconstruction approaches have demonstrated high geometric accuracy, they can involve a trade-off between image quality and geometry \cite{2DGS}, underscoring the need for densification strategies that are guided by the underlying 3D geometric structure of the scene.

\begin{figure*}[h!]
  \centering
   \includegraphics[width=1.0\textwidth]{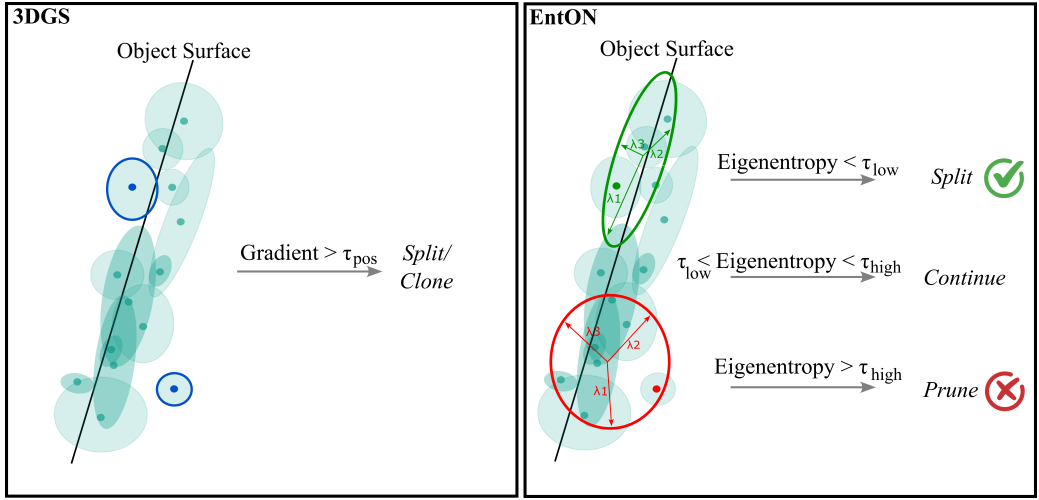}
   \caption{Methodology EntON. Gaussians are adapted based on the Eigenentropy of their local neighborhood: low Eigenentropy leads to splitting, medium Eigenentropy results in unchanged Gaussians, and high Eigenentropy triggers pruning. In constrast, 3DGS triggers densification based on the view-space position gradient: small Gaussians are cloned, large Gaussians are splitted.
   EntON uses the level of Eigenentropy to focus densification on object surfaces, avoiding unnecessary scene expansion and thus efficiently compressing the information content of the scene representation.}
   \label{fig:Methodology}
\end{figure*}

The standard 3DGS densification strategy relies primarily on view-space position gradients and does not explicitly account for geometric context information. As a result, the potential to align the splitting and pruning of Gaussians with their distribution in a local geometric 3D area remains unexploited. 
This limitation is particularly interesting when it comes to 3D reconstuction of man-made objects, as commonly encountered in photogrammetry and surveying applications of structures such as buildings. Those scenes are dominated by locally, piecewise planar surface structured areas that follow strong geometric regularities.
Therefore, we propose a guiding of the densification toward improved geometric alignment based on the local 3D distribution Gaussians. This can be done using 3D shape features from Gaussian neighborhoods. In the context of man-made, urban scenes, the shape feature Eigenentropy has been widely used in semantic segmentation and point cloud classification \cite{Weinmann_2015, Weinmann_2017_Feature_Relevance}, making it particularly well suited for this task. 

Our approach aims to explicitly exploit the local 3D structural geometry, by introducing a geometry-guided densification and pruning strategy EntON for 3D Gaussian Splatting (3DGS), and other splatting-based methods. We integrate the eigenvalue-based 3D shape feature \emph{Eigenentropy}, computed from the covariance matrix of the $k$-nearest neighboring Gaussians around each Gaussian center. 
During the densification process, the algorithm alternates between (i) standard gradient-based densification, which refines under- and over-reconstructed regions via high view-space gradients, and (ii) Eigenentropy-aware densification, which preferentially densifies Gaussians in low-Eigenentropy (ordered, anisotropic, flat) neighborhoods to better capture fine geometric details on the object surface, and prunes those in high-Eigenentropy (disordered, isotropic, spherical/scattered) regions. EntON guides the optimization toward improved geometric alignment, while compressing the information content of the scene to the object's surface to avoid unnecessary scene expansion.
 
We evaluate our method, EntON, on two different benchmark datasets: (i) small-scale DTU dataset using 15 diverse scenes, and (ii) large-scale TUM2TWIN dataset with urban scenes using two representative scenes. Geometric accuracy is measured using the Chamfer Distance (cloud-to-cloud). Photometric reconstruction quality is assessed via the Peak Signal-to-Noise Ratio (PSNR). To assess efficiency, we further report the final number of Gaussians and the total training time.

We demonstrate that EntON produces Gaussians that are better aligned with the underlying object surface geometry. By explicitly exploiting local 3D structural geometry, EntON improves geometric accuracy and reduces unnecessary model growth through targeted densification in planar, low-Eigenentropy areas, while simultaneously pruning Gaussians that contribute to locally scattered, high-Eigenentropy areas. Importantly, this geometry-guided optimization maintains high photometric quality and preserves a competitive model size and training time. Overall, EntON yields the following key outcomes:

\begin{itemize}
\item An improvement in \textit{geometric accuracy} of up to 32.7\% over 3DGS on average (and up to 39.8\% for the best neighborhood configuration) and 8.6\% over 2DGS, while remaining competitive with PGSR,

\item An improvement in \textit{rendering quality} of up to 6.8\% over 2DGS and PGSR on average (and up to 7.5\% for the best neighborhood configuration), while remaining competitive with 3DGS,

\item A reduction in the \textit{number of Gaussians} of up to 49.6\% compared to 3DGS on average (and up to 59.9\% for the most compact configuration), as well as reductions of 14.9\% compared to PGSR and 5.3\% compared to 2DGS,

\item A reduction in \textit{training time} of 22.7\% compared to 3DGS on average (and up to 29\% for the fastest configuration), as well as speedups of up to 45.0\% and 69.8\% over 2DGS and PGSR.

\end{itemize}

In the following, we first review related work in Section \ref{sec:RelatedWork}, providing an overview of 3D reconstruction techniques with a focus on Gaussian Splatting, including existing densification and pruning strategies. We also discuss different types of 3D shape features relevant for EntON. In Section \ref{sec:Methodology}, we then introduce our geometry-guided, Eigenentropy-aware densification strategy for 3DGS.
In Section \ref{sec:Experiments} we introduce the experimental setup, including datasets, evaluation metrics and implementation details.
In Section \ref{sec:Results}, we first demonstrate the effectiveness of EntON. We then present the results of our experiments on both small-scale and large-scale benchmark datasets, comparing EntON with 3DGS, 2DGS, and PGSR in terms of quantitative metrics and qualitative performance. Section \ref{sec:ablation} includes an ablation study for our method.
We analyze and discuss the outcomes in Section \ref{sec:Discussion}, highlighting the advantages of our approach with respect to geometric accuracy, rendering quality, and memory and time efficiency. Finally, Section \ref{sec:Conclusion} summarizes the contributions of our work and outlines potential directions for future optimizations and practical applications.

\section{Related Work}
\label{sec:RelatedWork}
We present an overview of novel view synthesis and 3D reconstruction techniques (Section \ref{sec:RelatedWork_c}), followed by an introduction to 3D reconstruction using Gaussian splats. Subsequently, we outline several densification strategies for 3D Gaussian Splatting (Section \ref{sec:RelatedWork_d}), and finally, we briefly introduce different types of 3D features (Section \ref{sec:3D_Features}), highlighting 3D shape features, such as Eigenentropy, which we use to steer our geometry-guided densification.

\subsection{3D Reconstruction}
\label{sec:RelatedWork_c}
The foundation for the rapid development of 3D scene reconstruction was laid with the introduction of Neural Radiance Fields (NeRFs) \cite{mildenhall_et_al_2020}, followed by numerous publications fostering the research and development of neural surface reconstructions, point cloud and mesh reconstructions \cite{unisurf, neus, volsdf, neuralangelo, densitygradient} in a variety of fields. NeRFs are neural networks with multilayer perceptrons (MLPs), that model the scene implicitly by estimating a color and volume density for each position and direction. These estimates are themselves subject to a certain degree of uncertainty \cite{uncertainty}.

Following the development of NeRFs, a novel concept for 3D scene reconstruction was introduced, in which scenes are explicitly described. 3D Gaussian Splatting (3DGS) \cite{3DGS} represents the scene using a large number of 3D Gaussians. To parameterize the scene, the Gaussians are initialized from a point cloud generated by Structure from Motion (SfM). This explicit representation avoids unnecessary computation in empty space, enables efficient GPU-based rasterization and allows real-time rendering with state-of-the-art visual quality \cite{3DGS}. Each Gaussians is defined by its mean, covariance, opacity, and spherical harmonics for color definition. The covariance is parameterized using scaling and rotation. These 3D Gaussians are projected onto the 2D image space, whereby their parameters (mean values for the Gaussian centers $\boldsymbol{\mu}$, scaling $S$, rotation $R$, opacity $\alpha$, and colors) are then refined during the optimization process to match the training images. This optimized process results in scenes with thousands to millions of Gaussians representing the 3D object geometry.
However, a huge amount of storage space is required, as 3DGS models the scene using a large number of Gaussians. 
Therefore, various methods focus on training speed or model capacity \cite{taming, speedysplat, HAC, lightgaussian, Gaussianpop, Gscodec, fastgs, fastergs}
In addition, the Gaussians do not take an ordered structure in general \cite{SuGaR}, indicating neither the centers nor the surfaces of the Gaussians are properly aligned with the actual object surface. And since the disorder of the Gaussians relies on the image reconstruction loss, it can result in incomplete geometric information \cite{PGSR}. 
To address this challenge, several methods \cite{2DGS, SuGaR, Surfels, PGSR, MVG_Splatting, Mip-Splatting, featuregs, 3dgs2pc, urbangs, objsplat, triangle} have been developed to achieve not only photometrically valid but also geometrically accurate 3D scene representations using 3DGS.
The concept of transforming 3D Gaussians into 2D ellipses or planar 3D ellipsoids in order to achieve higher geometric accuracy is widely used in many approaches.
SuGaR \cite{SuGaR} extracts meshes from 3DGS by introducing a regularization term that aligns Gaussians with the scene surface.
Surfels \cite{Surfels} combines 3D Gaussian points' optimization flexibility with the surface alignment of surfels by flattening 3D Gaussians into 2D ellipses, setting the z-scale to zero. 
2DGS \cite{2DGS} follows a similar approach and collapses 3D volumes directly into 2D planar Gaussian disks for view-consistent geometry, using perspective-accurate splatting with ray-splat intersection and depth and normal consistency terms. PGSR \cite{PGSR} flattens Gaussians into planes, using unbiased depth rendering to obtain precise depth information. MVG-Splatting \cite{MVG_Splatting} improves 2DGS by optimizing normal calculation and using an adaptive densification method guided by depth maps. MIP-Splatting \cite{Mip-Splatting} introduces a 3D smoothing filter to constrain Gaussian sizes based on the input views' sampling frequency, eliminating high-frequency artifacts. 3DGS-to-PC \cite{3dgs2pc} converts 3DGS scenes into accurate and dense point clouds by sampling points from each Gaussian’s multivariate distribution proportional to its scale and contribution, with outlier rejection to ensure geometric consistency. FeatureGS \cite{featuregs} incorporates an additional geometric loss term based on 3D shape features. Despite the high geometric accuracy, is achieves only moderate rendering quality, since the geometric loss affects all Gaussians equally.

\subsection{Densification and Pruning}
\label{sec:RelatedWork_d}

The optimization in 3DGS follows a successive adjustment of the Gaussians across the iterations based on rendering performance. The goal is to create based on the sparse SfM point set, a denser and better representation of the 3D scene. 
Thereby empty areas will be densified, focusing on under-reconstruction (areas with missing geometric features) and over-reconstruction (areas covered by large Gaussians). Gaussians in these areas supposedly have high view-space position gradients and are therefore candidates for densification. Thereby small Gaussians are cloned, while large Gaussians are split into two smaller Gaussians \cite{3DGS}.
Although this densification strategy is effective, it also causes certain issues. In particular, the 3DGS suffers from difficulties with over-reconstruction \cite{absgs} and oversized Gaussians \cite{microsplatting}, which leads to blurred rendered image areas, since splitting large Gaussians is important to allow good reconstruction \cite{3DGS}.


To overcome this issue, AbsGS \cite{absgs} proposes an homodirectional view-space positional gradient based on the sum of the absolute values of pixel-wise sub-gradients to identify large Gaussians as splitting candidates in over-reconstructed regions.
Gaussian Opacity Fields \cite{opacityfields} likewise incorporated a similar approach into their methodology to identify overly blurred areas.
Several works address the frequency of the gradients.
Micro-splatting \cite{microsplatting} changes densification in two key aspects: compact splats through covariance regularization by activating adaptive splitting when the trace exceeds a threshold, and adaptive stronger densification in high-frequency regions (high gradients). FreGS \cite{fregs} uses frequency regularization from high to low frequency signals following a coarse-to-fine-densification to overcome the over-reconstruction by analyzing the rendered images in the spectral space. AD-GS \cite{AD-GS} introduces an alternating densification, which combines high densification for fine details with low densification reduce artifacts. By applying geometry-constrained training, via edge-aware depth smoothness and pseudo-view consistency, it balances detail recovery with artifact mitigation. EFA-GS \cite{efa-gs} monitors the frequency behavior of gradients in floater artifacts in order to eliminate them. Another approach \cite{colorcued} replaces the view-space position gradient with a color-cued densification strategy, which leverages the 0th spherical harmonics gradient to identify regions with under-representing color.
To enhance training speedup, FastGS \cite{fastgs} assesses Gaussian importance via their multi-view consistency and employs corresponding densification and pruning strategies.

AbsGS \cite{absgs} shows that lowering the gradient threshold improves rendering quality, since fewer high-frequency Gaussians are split. However, when lower gradient thresholds are used, it leads to an excessive growth in the number of Gaussians. Moreover, the splitting decision of 3DGS relies on the view-space position gradients of the Gaussians. This means that densification is driven without considering the underlying local area of the 3D geometry. 

Motivated by this observation, we propose EntON, an alternating densification and pruning strategy that complements 3DGS densification scheme by explicitly exploiting geometric contextual information from the Gaussians of the reconstructed 3D scene. 
For man-made objects, as commonly encountered in the surveying of structures such as buildings, surface 3D points typically lie on structured, piecewise planar surfaces that follow strong geometric regularities. Instead of relying exclusively on view-space position gradients, we aim to split and prune Gaussians according to the geometric structure of their local neighborhoods in 3D space. The densification process then enhances a specific characterization of local 3D structures of man-made objects that is consistent with the Manhattan-World assumption~\cite{coughlan1999manhattan,coughlan2000manhattan}. We achieve this by introducing a geometry-guided denficiation decision process based on the Eigenentropy of Gaussian neighborhoods, which reinforces the dominance of structural entropy. To the best of our knowledge, EntON is the first method that targets geometric accuracy in combination with photometric quality directly within the 3D Gaussian Splatting densification process.

\subsection{3D Features}
\label{sec:3D_Features}

Several types of 3D features exist for point cloud-based applications such as classification, registration, or calibration.
Complex features, which cannot be interpreted directly include descriptors such as Shape Context 3D (SC3D) \cite{SC3D}, Signature of Histogram of OrienTations (SHOT) \cite{SHOT} or Fast Point Feature Histograms (FPFH) \cite{FPFH}. In contrast, interpretable features \cite{interpretable} are those that are directly interpretable, such as local 2D and 3D shape features. 
To describe the local structure around a 3D point, the spatial arrangement of other 3D points in the local neighborhood is often considered. Thereby the 3D covariance matrix, also known as the 3D structure tensor, is well-known and suitable for characterizing the shape properties of 3D data \cite{Weinmann_2015}. It is derived explicitly for each point from the point itself and its local neighbors. The three eigenvalues, \( \lambda_1 \geq \lambda_2 \geq \lambda_3 \geq 0 \), correspond to an orthogonal system of eigenvectors (\( \epsilon_1, \epsilon_2, \epsilon_3 \)), which indicate the direction (\textit{rotation}) of the three ellipsoid principal axes and correspond to the extent (\textit{scales}) of the 3D ellipsoid along the principal axes. Based on the behavior of the eigenvalues \( \lambda_1, \lambda_2 \), and \( \lambda_3 \), linear (\( \lambda_1 \gg \lambda_2, \lambda_3 \)), planar (\( \lambda_1 \approx \lambda_2 \gg \lambda_3 \)), and spherical (\( \lambda_1 \approx \lambda_2 \approx \lambda_3 \)) structures can be intuitively described.
The usage of geometric 3D shape features has led to thousands of publications in various fields over the past few decades, especially for the semantic segmentation and classification \cite{Weinmann_2015, Weinmann_2017_Feature_Relevance, weinmann20203d} of point clouds. But also for calibration \cite{hillemann2019automatic} or registration \cite{pointcloudregistration} of 3D point clouds.

\section{Methodology}
\label{sec:Methodology}

In this section, we present our method EntON (Figure \ref{fig:Methodology}), which introduces a geometry-guided densification and pruning strategy for 3D Gaussian Splatting. The proposed strategy guides densification and pruning based on local geometric Gaussian 3D neighborhoods, directly improves geometric accuracy and maintains photometric quality surface-focused through alignment of Gaussians. It leverages the 3D shape feature Eigenentropy, derived from the eigenvalues of the covariance matrix computed over local Gaussian neighborhoods, which enables targeted control over geometric regularity, promoting planar, low-entropy regions.

We first review the standard 3DGS densification strategy (Section \ref{sec:densification_3dgs}), which forms the basis of EntON. Then we introduce the background and scheme of our geometry-guided densification (Section \ref{sec:densification_geometry}), which is based on the Eigenentropy of local Gaussian neighborhoods. Finally, we present the resulting alternating densification strategy EntON (Section \ref{sec:densification_Enton}).

\subsection{3DGS Densification}\label{sec:densification_3dgs}

The optimization in 3D Gaussian Splatting (3DGS) \cite{3DGS} iteratively adjusts the Gaussians based on the rendering performance. This process starts from a sparse initial set of Gaussians, whose centers correspond to the 3D points of an SfM point cloud derived from SIFT \cite{Lowe} features. Starting from the sparse SfM point set, the objective is to adaptively controls the number and spatial density of 3D Gaussians to transform the initial sparse representation into a denser one, while removing nearly transparent Gaussians and populating previously empty regions \cite{3DGS}. Rendered images, obtained by projecting the 3D Gaussians onto the 2D image plane, are compared against the ground-truth training images. The Gaussians are then adapted by splitting/ cloning, or pruning them accordingly. 

In the standard adaptive densification and pruning strategy of 3DGS, Gaussians with opacity $\alpha$ below a predefined threshold are removed (pruned). Conversely, under-reconstructed regions (areas lacking sufficient geometric detail) and over-reconstructed regions (areas covered by large Gaussians) are targeted for densification.
Gaussians in these areas typically exhibit high view-space position gradients and thus serve as candidates for densification.
In 3D Gaussian Splatting~\cite{3DGS}, the view-space position gradient measures the sensitivity of the photometric loss to small displacements of the Gaussian center $\mu$ in 3D space, as observed through its 2D projection in each training view. This gradient is computed via backpropagation through the differentiable rasterizer. The magnitude corresponds to the norm of the gradient with respect to the projected 2D coordinates in view space, accumulated and averaged over contributing pixels and views as implemented in the official codebase\footnote{\url{https://github.com/graphdeco-inria/gaussian-splatting} \\(last access 07/21/2024)}. A mathematical interpretation (see also \cite{absgs, opacityfields}) of this accumulation is:

\begin{equation}
\nabla_{\boldsymbol{\mu}} L = \sum_{i} \frac{\partial L}{\partial \mathbf{p}_i} \cdot \frac{\partial \mathbf{p}_i}{\partial \boldsymbol{\mu}}.
\label{eq:gradient_accum}
\end{equation}

where the sum runs over all pixels $i$ to which the Gaussian contributes in a given view, and the full gradient is accumulated over multiple views across the training iterations. The key densification criterion is then the average magnitude:
\begin{equation}
\bar{g} = \frac{1}{N} \sum_{j=1}^{N} \left\| \nabla_{\boldsymbol{\mu}}^{(j)} L \right\|_2,
\label{eq:avg_grad}
\end{equation}
(or an equivalent per-component 2D norm averaging as in the referenced implementation), where $N$ is the number of accumulated gradient contributions. In practice, a Gaussian that appears in many different views (due to its visibility) will accumulate gradient contributions across numerous iterations. Consequently, the accumulated gradients implicitly stem from multiple views. Densification (cloning or splitting) is triggered when this average gradient $\bar{g}$ exceeds the threshold $\tau_{\mathrm{pos}}$, with $\tau_{\mathrm{pos}} = 0.0002$ by default \cite{3DGS}.

\subsection{Geometry-Guided Densification}\label{sec:densification_geometry}

3D Gaussian Splatting primarily relies on view-space position gradients to identify Gaussians whose 3D positional changes strongly affect the 2D rendered image, triggering the densification based on the magnitude of view-space position gradients. However, this criterion does not explicitly incorporate information of the underlying local geometry.
Specifically, whether a Gaussian lies on or near an actual object surface, where refinement for increased geometric detail is most beneficial. This can be done using 3D shape features from Gaussian neighborhoods.
In the task of semantic segmentation and point cloud classification of urban scenes in previous work \cite{Weinmann_2017_Feature_Relevance}, the 3D shape feature Eigenentropy has also proven to be particularly useful.
Eigenentropy has been shown to be an appropriate 3D feature for characterizing plane point cloud structures \cite{Dittrich} and a powerful tool for scale selection \cite{scaleselection}, making it particularly well suited for this task.

To address the limitation, we bias the densification process through an Eigenentropy-aware densification strategy. Particularly the splitting, toward Gaussians residing in locally flat, low-Eigenentropy neighborhoods, which are strong indicators of their alignment with the object surfaces. Specifically, we assess whether a Gaussian is located in a region exhibiting low to medium Eigenentropy and near-planar (flat) structure. Gaussians in such neighborhoods are preferentially split to enhance surface reconstruction fidelity and geometric accuracy. Conversely, Gaussians in high-Eigenentropy regions (disordered, isotropic, spherical/scattered) are densified more conservatively or become candidates for pruning. This directs the densification towards geometrically significant surface areas and is based on the Manhattan World assumption~\cite{coughlan1999manhattan,coughlan2000manhattan}, where artificial environments consist predominantly of flat structures.

Addressing the local geometry, we extract geometric feature Eigenentropy from a local 3D structure around each Gaussian center: This involves computing the covariance matrix from a local neighborhood (Section \ref{sec:geometry_covariance}), normalizing the eigenvalues (Section \ref{sec:geometry_normalize}), and finally extracting the feature Eigenentropy (Section \ref{sec:geometry_feature}). We further describe the typical geometric characteristics of the Eigenentropy (Section \ref{sec:geometry_feature_characteristics}) that are relevant for EntON, followed by the resulting Eigenentropy-aware densification strategy (Section \ref{sec:geometry_eigenetropy-aware}).

\subsubsection{Local Neighborhood of Gaussians}\label{sec:geometry_covariance}
To compute the interpretable geometric 3D shape features from the spatial arrangement of points within a local neighborhood around each 3D point (here: Gaussian center), a neighborhood must first be defined. The neighborhood size serves as a scale parameter and directly influences the resulting features, as only the selected neighboring points contribute to the eigenvalue analysis of the local covariance matrix. The 3D covariance matrix, also called 3D structure tensor \cite{structur_tensor}, is well-known to characterize such shape properties \cite{Weinmann_2015} and derived from a point and its local neighborhood. 
The three eigenvalues, \( \lambda_1 \geq \lambda_2 \geq \lambda_3 \geq 0 \), correspond to an orthogonal system of eigenvectors (\( \epsilon_1, \epsilon_2, \epsilon_3 \)), which indicate the direction (\textit{rotation}) of the three ellipsoid principal axes and correspond to the extent (\textit{scales}) of the 3D ellipsoid along the principal axes. 

Given a point \( p_0\) in the 3D space, i.e., the center of a Gaussian, we define its \(k\)-nearest neighbors \(\{p_1, p_2, \ldots, p_k\}\). The centroid \(\bar{p}\) (Equation \ref{equ:p}) of this neighborhood is computed as:

\begin{equation}\label{equ:p}
\bar{p} = \frac{1}{k+1} \sum_{i=0}^{k} p_i
\end{equation}

\begin{figure}[h!]
    \centering
        \includegraphics[width=0.3\linewidth]{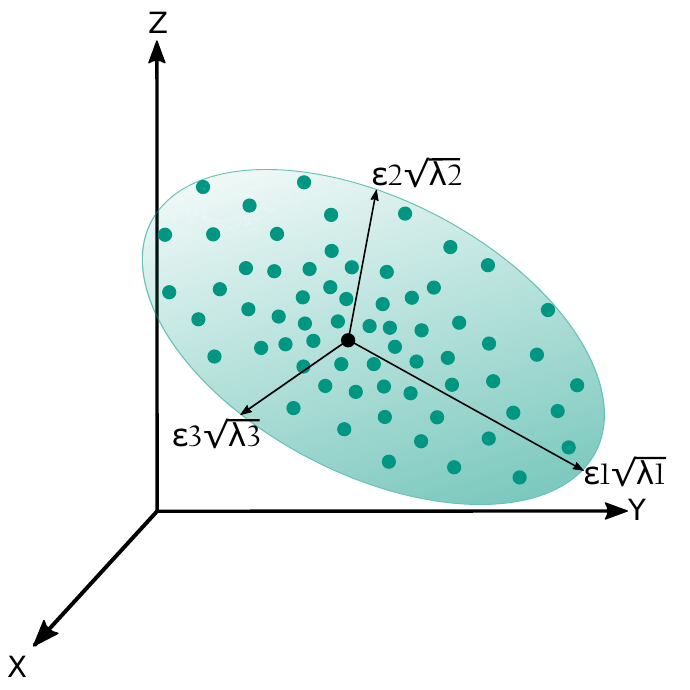}
 \caption{Representation of an ellipsoid from the neighborhood points represented by the Gaussian centers with the three eigenvectors (\(\mathbf{\epsilon_1}, \mathbf{\epsilon_2}, \mathbf{\epsilon_3}\)) and the corresponding eigenvalues (\(\lambda_1, \lambda_2, \lambda_3\)) in the three-dimensional coordinate system.}
    \label{fig:neighborhood_ellipsoid}
\end{figure}

The covariance matrix \(C\) (Equation \ref{equ:C}) \cite{weinmann_covariance} for the neighborhood (Figure \ref{fig:neighborhood_ellipsoid}) is then:

\begin{equation}\label{equ:C}
C = \frac{1}{k+1} \sum_{i=0}^{k} (p_i - \bar{p})(p_i - \bar{p})^T
\end{equation}

\subsubsection{Eigenvalue Normalization}\label{sec:geometry_normalize}
To ensure consistency, eigenvalues \(\lambda_1, \lambda_2, \lambda_3\) from the Gaussian neighborhood covariance matrix are normalized by dividing by the sum of the eigenvalues for each case by (Equation \ref{equ:C_knn}):
\begin{equation}\label{equ:C_knn}
\lambda'_i = \frac{\lambda_i}{\text{sum}(\mathbf{\lambda})} \quad \text{for} \enspace i \in \{1, 2, 3\},
\end{equation}
with
\begin{eqnarray}
 \text{sum}(\mathbf{\lambda}) = \sum_{i=1}^3 \lambda_i.
\end{eqnarray} 
 
The normalized eigenvalues \(\lambda'_1, \lambda'_2, \lambda'_3\) are then ordered in descending order for being used for the final geometric 3D feature computation:
\[
\quad \lambda'_1 \geq \lambda'_2 \geq \lambda'_3 \geq 0.
\]

\subsubsection{Feature Extraction}\label{sec:geometry_feature}

To enhance these structural properties that 3D point clouds of man-made objects exhibit, we incorporate the feature-aware densification und pruning using the \(k\)-nearest neighbors (kNN) of each point. This approach allows for the calculation of spatial features in the local neighborhood of each Gaussian. 

The feature Eigenentropy is defined according to \cite{weinmann_covariance}  as the Shannon entropy:

  \begin{equation}\label{equ:loss_eigen_knn}
\text{Eigenentropy}_{\text{kNN}} = - \sum_{i=1}^{3} \lambda'_i \log(\lambda'_i)
\end{equation}
and quantifies the order/disorder of points by measuring the entropy within their local 3D neighborhood.

\subsubsection{Geometric Characteristics of Eigenentropy}\label{sec:geometry_feature_characteristics}


The eigenvalues \(\lambda_1 \geq \lambda_2 \geq \lambda_3 \geq 0\) of this covariance matrix enable the characterization of dominant local structures, whereby the Eigenentropy exhibits characteristic values for local linear, planar, and spherical distributed structures:
\begin{itemize}
    \item For an ideal linear structure (\(\lambda'_1 \approx 1\), \(\lambda'_2 \approx \lambda'_3 \approx 0\)):
    \[
    E \approx -(1 \cdot \log 1 + 0 \cdot \log 0 + 0 \cdot \log 0) = 0.
    \]
    This corresponds to highly anisotropic, ordered distributions.
    \item For an ideal planar structure (\(\lambda'_1 = \lambda'_2 = 1/2\), \(\lambda'_3 = 0\)):
    \[
    E = -\left( \frac{1}{2} \log \frac{1}{2} + \frac{1}{2} \log \frac{1}{2} + 0 \cdot \log 0 \right) = \log 2 \approx 0.693.
    \]
    In practice, for near-planar distributions where \(\lambda'_3 \approx 0\) but \(\lambda'_1\) and \(\lambda'_2\) vary (while summing to 1), the Eigenentropy satisfies \(E \leq \log 2\), with the maximum achieved at the balanced case \(\lambda'_1 = \lambda'_2 = 1/2\). This behavior is particularly relevant in idealized settings, such as when Gaussians lie perfectly flat on surfaces (e.g., walls or floors in Manhattan-world scenes).
    \item For isotropic/spherical structures (\(\lambda'_1 \approx \lambda'_2 \approx \lambda'_3 \approx 1/3\)):
    \[
    E \approx -\log \frac{1}{3} \approx 1.099,
    \]
    reflecting maximal local disorder.
\end{itemize}

\begin{figure}[h!]
    \centering
    \includegraphics[width=1.1\linewidth]{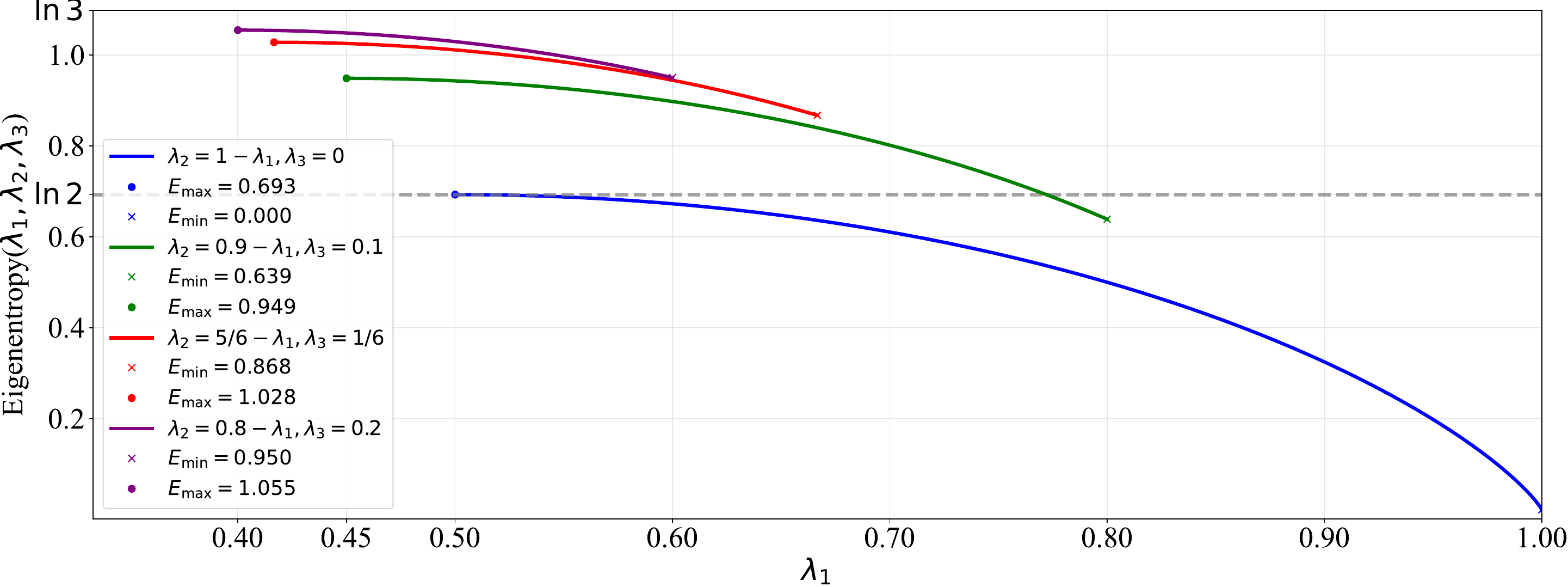}
    \caption{Eigenentropy $E(\lambda_1, \lambda_2, \lambda_3)$ as a function of the largest normalized eigenvalue $\lambda_1$ (with $\lambda_1 \geq \lambda_2 \geq \lambda_3 \geq 0$, $\sum_{i=1}^3 \lambda_i = 1$). 
    Curves represent different fixed values of $\lambda_3$ (ideal planar: $\lambda_3=0$; near-planar: $\lambda_3=0.1$; transitional between planar and spherical: $\lambda_3=1/6\approx0.1667$; and higher spherical distribution).
    Markers indicate minimum and maximum Eigenentropy for each case. The dashed line at $\ln 2 \approx 0.693$ corresponds to the ideal planar case ($\lambda_1 = \lambda_2 = 0.5$, $\lambda_3 = 0$).}
    \label{fig:Eigenentropy_behavior}
\end{figure}

The Eigenentropy provides a scalar measure of the local order versus disorder (structural entropy) in the 3D neighborhood \cite{Weinmann_2015}.
Low Eigenentropy thus indicates highly structured, anisotropic local geometry (favoring flat, linear or planar areas), while high values signal disordered or volumetric, spherical distributions. Figure~\ref{fig:Eigenentropy_behavior} illustrates this behavior of Eigenentropy for various normalized eigenvalue configurations. 
Splitting Gaussians when their local Eigenentropy falls below a threshold of $\ln 2 \approx 0.693$ favors the presence of flat, anisotropic, low-entropy structures, including both strongly linear and planar flat geometries. The plotted examples include $\lambda_3 = 0.0$ (ideal planar), $0.1$, $0.15$, and $0.2$ (increasing deviation from ideal planarity). The aim is to focus the attention within the densification process to Gaussian neighborhoods in flat, low-Eigenentropy regions.

\textit{Why not use planarity?} 
Planarity $P$ as a 3D feature ($P = \frac{\lambda_2 - \lambda_3}{\lambda_1}$) assigns high values to ideal planar structures ($P \approx 1$). 
However, low planarity values ($P \approx 0$) are ambiguous, as they occur for both highly linear ($\lambda_1 \gg \lambda_2 \approx \lambda_3$) and isotropic, spherical neighborhoods ($\lambda_1 \approx \lambda_2 \approx \lambda_3$). As a result, a single planarity threshold cannot reliably distinguish between flat, surface-aligned structures and disordered volumetric regions, making robust splitting and pruning decisions difficult.
In contrast, Eigenentropy resolves this ambiguity by explicitly quantifying the degree of local structural order/ disorder, enabling a more stable geometry-guided densification strategy.

\subsubsection{Eigenentropy-Aware Densification}\label{sec:geometry_eigenetropy-aware}

\begin{figure}[h!]
  \centering
   \includegraphics[width=0.65\textwidth]{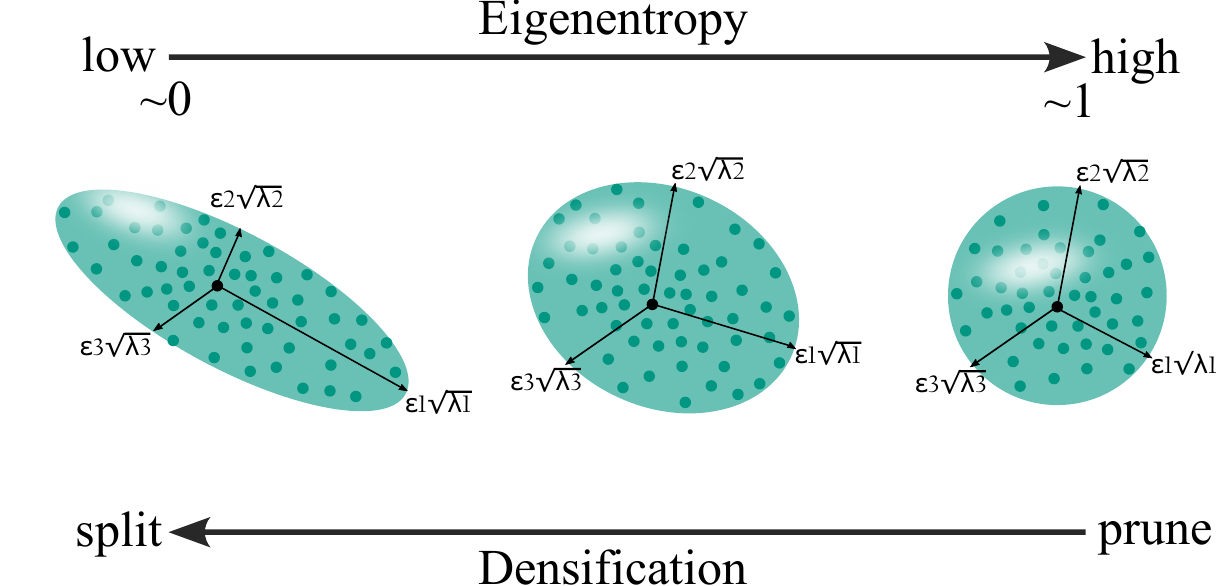}
   \caption{Low Eigenentropy leads to splitting, medium Eigenentropy results in unchanged Gaussians, and high Eigenentropy triggers pruning. Representation of the ellipsoids based on neighboring Gaussian centers with the three eigenvectors (\(\mathbf{\epsilon_1}, \mathbf{\epsilon_2}, \mathbf{\epsilon_3}\)) and the corresponding eigenvalues (\(\lambda_1, \lambda_2, \lambda_3\)) in the three-dimensional coordinate system.}
   \label{fig:Eigenentropy_ellipsoid}
\end{figure}

Our method leverages the eigenvalue-derived 3D shape feature Eigenentropy to assess and enhance local 3D geometric shape properties around each Gaussian center in a local neighborhood (Figure~\ref{fig:Methodology}): We integrate a geometry-guided densification process (illustrated in Figure~\ref{fig:Eigenentropy_ellipsoid}) based on the 3D shape feature Eigenentropy computed from the covariance matrix of the $k$-nearest neighbors (kNN) around each Gaussian center. During optimization, Gaussians are split, or pruned according to their local neighborhood characteristics. This strategy enables a geometry-guided control of the Gaussian distribution, promoting structured, low-entropy local neighborhoods. Specifically:
\begin{itemize}
    \item Gaussians exhibiting \textbf{low Eigenentropy} (indicating more ordered, anisotropic, flat regions) are preferentially split to increase local density in these areas and better capture fine geometric details on the objects surface.
    \item Gaussians with \textbf{high Eigenentropy} (indicating disordered, isotropic, spherical/scattered regions) are preferentially pruned.
    \item Gaussians falling within a \textbf{transitional Eigenentropy interval} are left unchanged to preserve stability and prevent unnecessary splitting or premature pruning.
\end{itemize}

\subsection{Alternating Densification}\label{sec:densification_Enton} 

To effectively combine the complementary strengths of solely gradient-based and geometry-guided densification, we adopt an \textit{alternating} densification strategy EntON (Algorithm~\ref{alg:Eigenentropy_densification}): it alternates between two complementary strategies at regular intervals during optimization, (i) gradient-based densification of 3DGS and (ii) Eigenentropy-aware, geometry-guided densification.

The rationale is as follows: gradient-based densification effectively refines regions with high view-dependent gradients, while Eigenentropy-aware densification reinforces a specific geometric characteristic of local spatial structures. It focuses the attention of the densification on the object surface and avoids 'unnecessary' scene expansion in high-Eigenentropy regions, while compressing the information content of the scene to the object's surface.
Alternating these two strategies prevents both over-reconstruction and under-reconstruction, leading to a scene representation with photometric fidelity and geometrically accurate, structured Gaussians. By alternating reverting to gradient-based densification, we safeguard high-contribution Gaussians using the default gradient threshold. This maintains an overall reconstruction quality, while the Eigenentropy-aware steps guide the distribution of the Gaussians towards a geometrically accurate structures.

Starting, after a pre-training, from iteration 3000, the training alternates every 100 iterations between two densification modes:
\begin{itemize}
     \item \textbf{Gradient-aware densification}  (3DGS): This strategy ensures that Gaussians with persistently high view-space position gradients, which indicate strong photometric contribution, are reliably cloned/split, even if their local neighborhood does not yet exhibit the Eigenentropy criteria.
    \item \textbf{Eigenentropy-aware densification:} This strategy relies on the extracted Eigenentropy of local Gaussian neighborhoods. It selectively reinforces geometric accuracy by preferentially splitting Gaussians (even with lower view-space position gradients) in low-Eigenentropy (ordered, anisotropic, flat) neighborhoods and pruning those in high-Eigenentropy (disordered, isotropic, spherical/scattered) regions.
\end{itemize}

The first 3000 iterations are pure gradient-based densification (3DGS) to establish a sufficiently dense initial representation. This pre-densification phase is essential, as the reliability of the eigenvalue-derived features strongly depend on adequate local point density and neighborhood definition. After this phase, densification is performed every 100 iterations, following the default densification schedule used in 3DGS, and we therefore keep the same step size. The Gaussians are divided further depending on their size into 2, 4 or 8 Gaussians to allow a more uniform spatial distribution, which is inspired by \cite{microsplatting}.

\begin{algorithm}[H]
\caption{Alternating Densification with EntON}
\label{alg:Eigenentropy_densification}
\For{each training iteration $t$}{

    \If{$t \bmod 100 = 0$}{
        \If{$t < 3000$}{
            \tcp{Pre-training: Gradient-based 3DGS densification}
            clone or split Gaussians with high view-space gradient magnitude\;
        }
        \Else{
            \tcp{Alternating strategy: switch every 100 iterations}
            \If{$(t / 100) \bmod 2 = 0$}{
                \tcp{Gradient-based 3DGS densification}
                clone or split Gaussians with high view-space gradient magnitude\;
            }
            \Else{
                \tcp{Eigenentropy-aware densification}
                \For{each Gaussian}{
                    compute local Eigenentropy $E$ from kNN covariance\;
                    \If{$E \leq \tau_{\text{low}}$}{
                        split\;
                    }
                    \ElseIf{$E > \tau_{\text{high}}$}{
                        prune\;
                    }
                    \Else{  \tcp{$ \tau_{\text{low}} < E \leq \tau_{\text{high}} $}
                        keep/ continue\;
                    }
                }
            }
        }
    }
}
\end{algorithm}

\section{Experiments}
\label{sec:Experiments}

In this section, we present the experimental setup by introducing the used datasets (Section \ref{sec:dataset}), the evaluation metrics (Section \ref{sec:metrics}), and the implementation details (Section \ref{sec:implementation}).

\subsection{Data} \label{sec:dataset}

EntON is evaluated on two benchmarks, a small-scale and a large-scale dataset. 
\paragraph{Small-Scale dataset}
The small-scale dataset DTU \cite{dtu} consists of scenes featuring real objects, including either 49 or 64 RGB images, corresponding camera poses, and reference point clouds obtained from a structured-light scanner (SLS). We specifically focus on the same 12 scenes as as previous approaches \cite{2DGS, PGSR, Surfels, MVG_Splatting}.
\paragraph{Large-Scale dataset}

The large-scale dataset TUM2TWIN \cite{wysocki2026TUM2TWIN} contains large-scale ourdoor scenes of the Technical University of Munich, including different building types and sizes.
Motivated by GS4BUILDINGS \cite{gs4buildings}, we focus on two distinct building clusters, each containing approximately 20–30 images. For geometric 3D accuracy comparison, we use each a subset of the reference point clouds from UAV laser scanning (ULS).

\subsection{Metrics}\label{sec:metrics}

To evaluate our method quantitatively and qualitatively. For 3D geometric surface accuracy Chamfer cloud-to-cloud distance by following the DTU evaluation procedure \cite{dtu}, which masks out points above 10\,mm since the reference point clouds are incomplete. Low Chamfer distance indicates high geometric accuracy. 2D rendering quality of the images is evaluated with the Peak Signal-to-Noise Ratio (PSNR) in dB, whereby a high PSNR is targeted. Considering efficiency of our method, we report the training time and the number of Gaussians needed to represent the scene.

\subsection{Implementation Details and Experiments}\label{sec:implementation}

\paragraph{Implementation}
3D Gaussian Splatting\footnote{\url{https://github.com/graphdeco-inria/gaussian-splatting} \\(last access 07/21/2024)} for comparison purpose, and as fundament for our method, is processed according to the original implementation, using default densification strategies and the default parameters with a view-space position gradient of 0.0002, learning rates of 0.0025 for spherical harmonics features, 0.05 for opacity adjustments, 0.005 for scaling operations and 0.001 for rotation transformations, on a NVIDIA RTX3090 GPU. 
All experiments on the large-scale dataset were performed by using the automatic image resolution downscaling applied by 3DGS in its default configuration, in order to match the memory constraints of the used GPU.
For further comparison purpose, we use 2D Gaussian Splatting\footnote{\url{https://github.com/hbb1/2d-gaussian-splatting} \\(last access 04/29/2025)} and PGSR\footnote{\url{https://github.com/zju3dv/PGSR} \\(last access 05/11/2025)}, which are processed according to the original implementation by using default parameters.

\paragraph{Neighborhood Definition}
The size of the local neighborhood directly influences the covariance matrix and, consequently, the resulting eigenvalues and derived shape features, and consequently the Eigenentropy. Different neighborhood sizes capture geometric structure at varying spatial scales, affecting the sensitivity of Eigenentropy to local surface characteristics. 
We investigate multiple neighborhood sizes to analyze their impact on the resulting Eigenentropy-aware densification:
We test our approach on different neighborhood sizes of $k \in \{25, 50, 75, 100\}$ to analyze the impact of the number of nearest neighbors on the results. Since the Gaussian spatial density increases during training, we also experiment with adaptive neighborhood sizing, where the kNN is incremented every 2500 iterations (from 25 to 50, to 75, and finally 100).
Computing k-nearest neighbors for each Gaussian can be time-consuming, especially for large-scale scenes. To efficiently handle neighborhood queries, we leverage the optimized kNN implementations in PyTorch3D\footnote{\url{https://github.com/facebookresearch/pytorch3d} (last accessed: 08/02/2024)}, which provide GPU-accelerated neighborhood searches.

\paragraph{Densification Thresholds}

For 3DGS densifcation we use the default view-space gradient threshold \cite{3DGS} of 3DGS setted to $\tau_{\mathrm{pos}} = 0.0002$. The gradient threshold for EntON is set to $\tau_{\mathrm{pos}} = 0.0001$, to allow a more sensitive splitting of the Gaussians.
For the Eigenentropy-aware densification in EntON, the Eigenentropy thresholds $\tau_{\text{high}}$  and $\tau_{\text{low}}$ are derived from the characteristic behavior of the Eigenentropy feature for locally structured neighborhoods, in particular linear and planar configurations with low sphericity, as described in Section~\ref{sec:geometry_feature_characteristics}. i)
For an ideal planar neighborhood, the Eigenentropy equals $\ln(2)$ is therefore used as the lower threshold $\tau_{\text{low}} = \ln(2)$, enabling splitting and explicitly favoring Gaussians embedded in low-entropy, flat neighborhood structures.
ii) The upper threshold $\tau_{\text{high}}$, which triggers pruning, is determined empirically based on the observed Eigenentropy distribution of outlier Gaussians and set to $\tau_{\text{high}} = 0.95$ (Ablation study, Section \ref{sec:ablation}) As illustrated in Figure~\ref{fig:Eigenentropy_behavior} in Section~\ref{sec:geometry_feature_characteristics}, neighborhoods with Eigenentropy values approx. $E \geq 0.868$ predominantly correspond to unstructured, spherical distributions. iii) Gaussians with values within the interval $\tau_{\text{low}} < E \leq \tau_{\text{high}}$ remain unchanged.

\section{Experimental Results}\label{sec:Results}

The following sections present both qualitative and quantitative results of EntON in comparison to 3DGS, 2DGS, and PGSR. First, we demonstrate that EntON effectively influences the Eigenentropy of local Gaussian neighborhoods in general (Section \ref{sec:Eigenentropy_distribution}), report the results on the small-scale DTU dataset (Section \ref{sec:Results_small}), followed by the results on the large-scale TUM2TWIN dataset (Section \ref{sec:Results_large}).

\subsection{Eigenentropy Distribution}\label{sec:Eigenentropy_distribution}

\begin{figure}[H]
    \centering
        \includegraphics[width=1\linewidth]{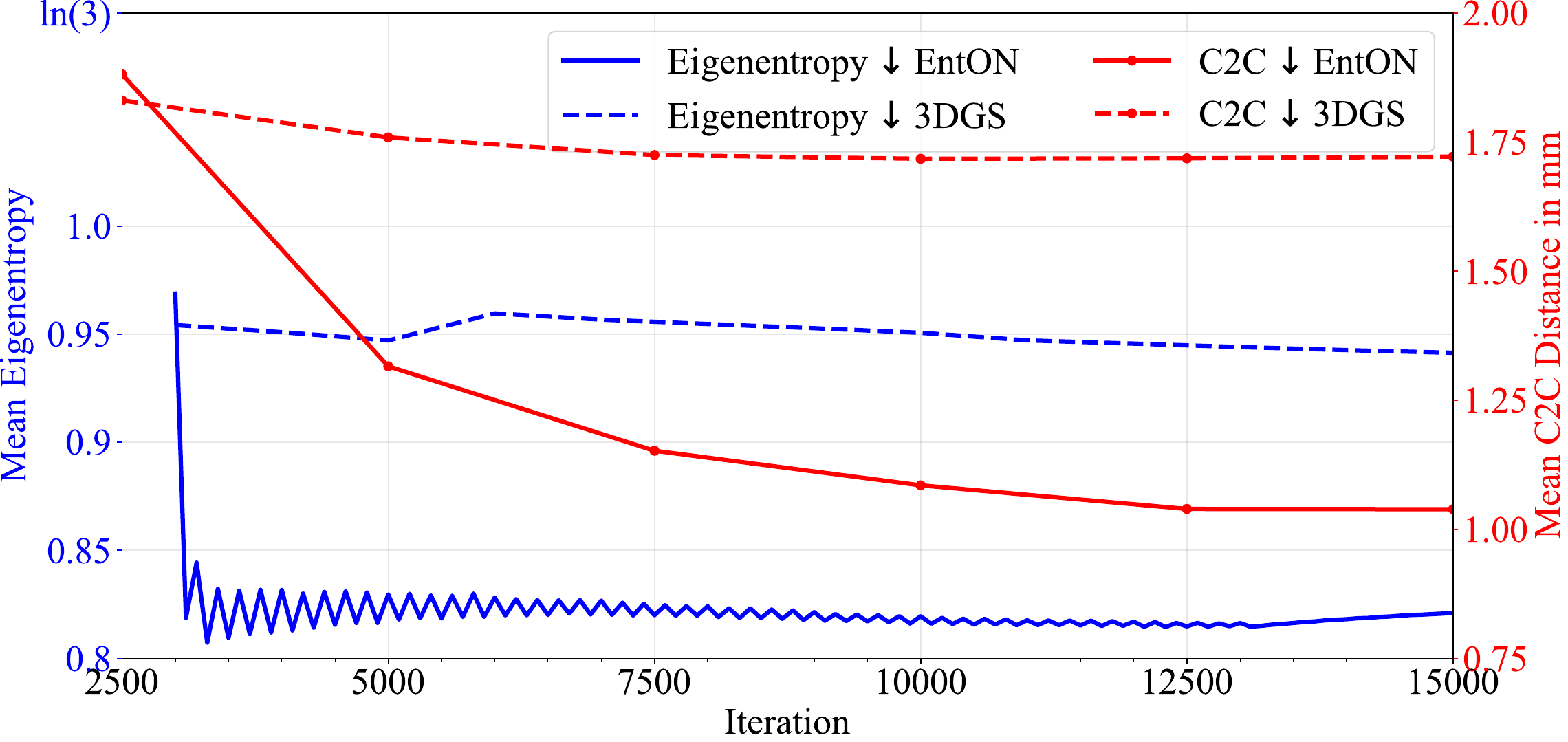}
 \caption{Trend of the mean Eigenentropy $\downarrow$ and mean cloud to cloud (C2C) distance $\downarrow$ of all DTU scenes during the training process. Comparison of 3DGS and EntON. As EntON starts at iteration 3000, C2C is first reported at iteration 2500.}
    \label{fig:Eigenentropy_vs_iteration}
\end{figure}

\begin{figure*}[h!]
    \centering
    \vspace{-2mm}
    \begin{tabular}{c c c c c c }
        \textbf{} 
        & \scriptsize\textbf{3DGS} 
        & \scriptsize\textbf{EntON ($knn_{100}$)} 
        & \scriptsize\textbf{EntON ($knn_{75}$)} 
        & \scriptsize\textbf{EntON ($knn_{50}$)} 
        & \scriptsize\textbf{EntON ($knn_{25}$)} \\[1ex]
    \vspace{-2mm}
        \rotatebox{90}{\scriptsize\textbf{Eigenentropy$\downarrow$}} &
        \includegraphics[width=0.15\textwidth]{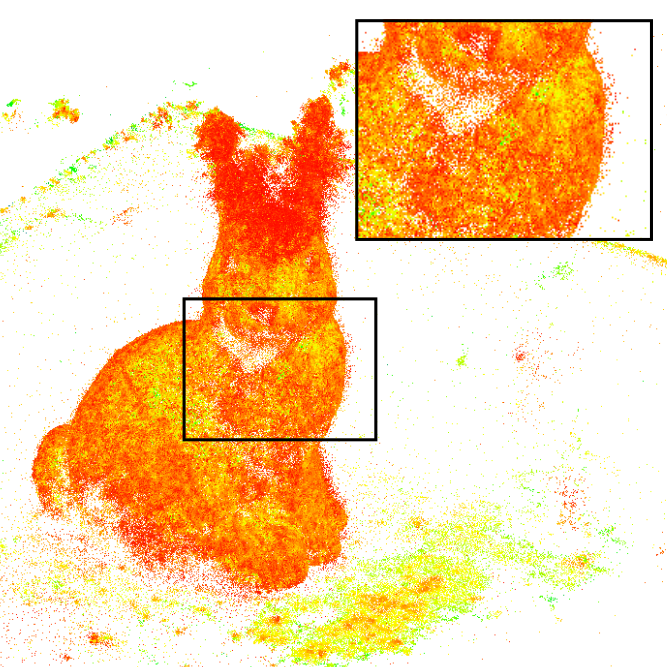} &
        \includegraphics[width=0.15\textwidth]{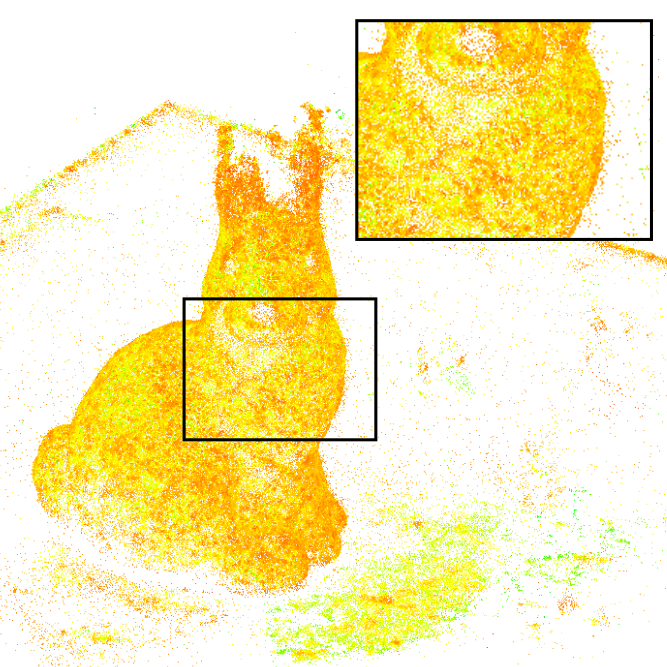} &
        \includegraphics[width=0.15\textwidth]{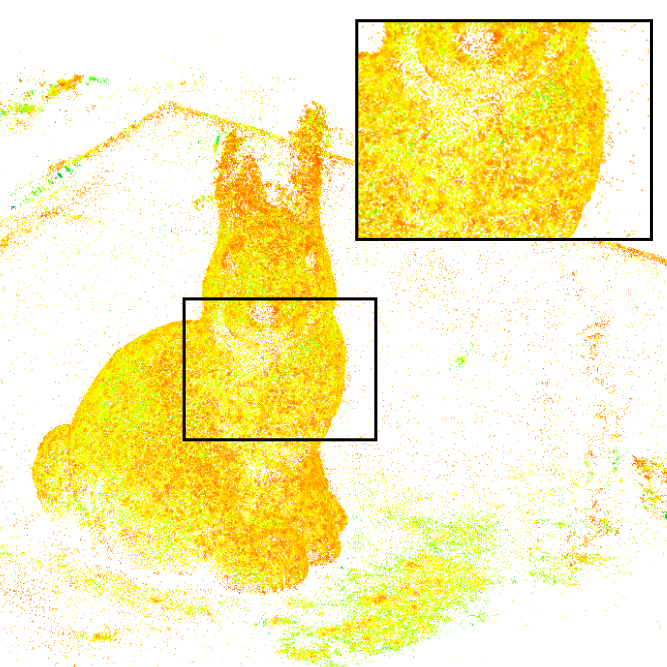} &
        \includegraphics[width=0.15\textwidth]{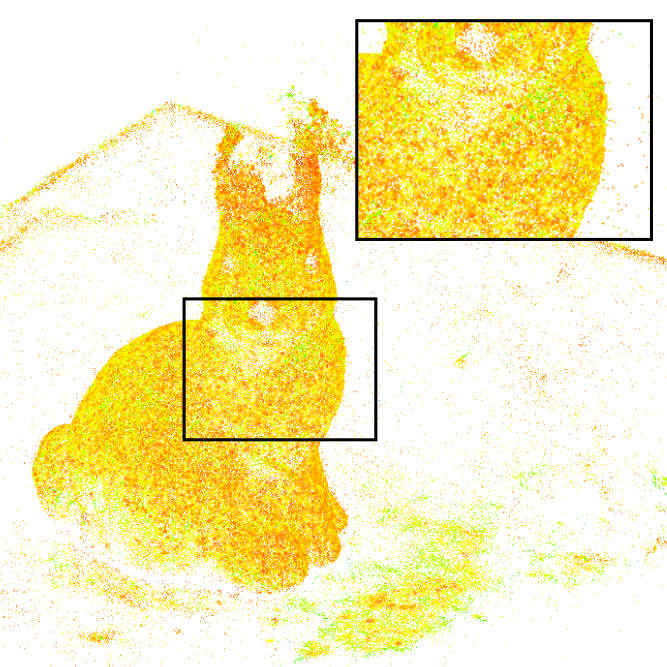} &
        \includegraphics[width=0.15\textwidth]{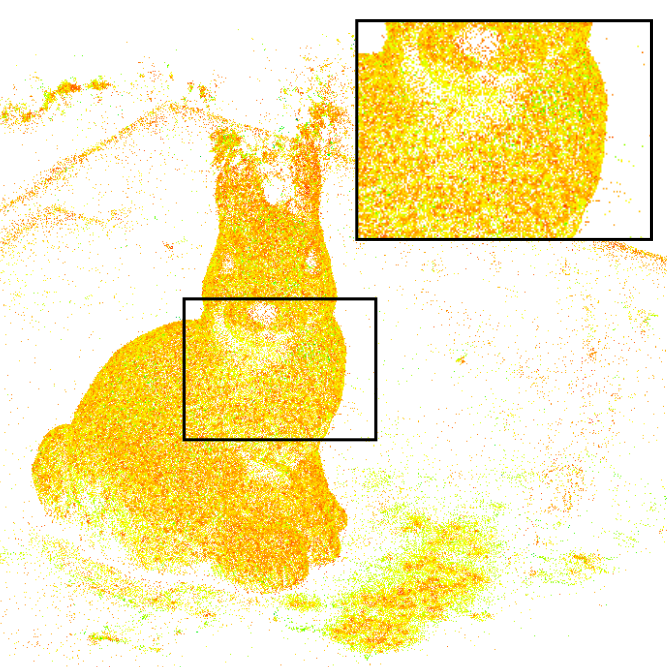} \\[2ex]
    \vspace{-2mm}
        \rotatebox{90}{\scriptsize\textbf{C2C $\downarrow$}} &
        \includegraphics[width=0.15\textwidth]{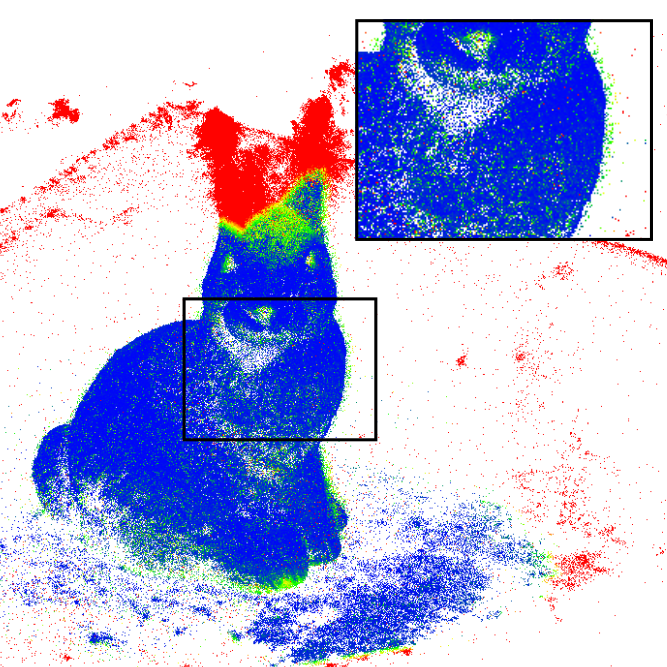} &
        \includegraphics[width=0.15\textwidth]{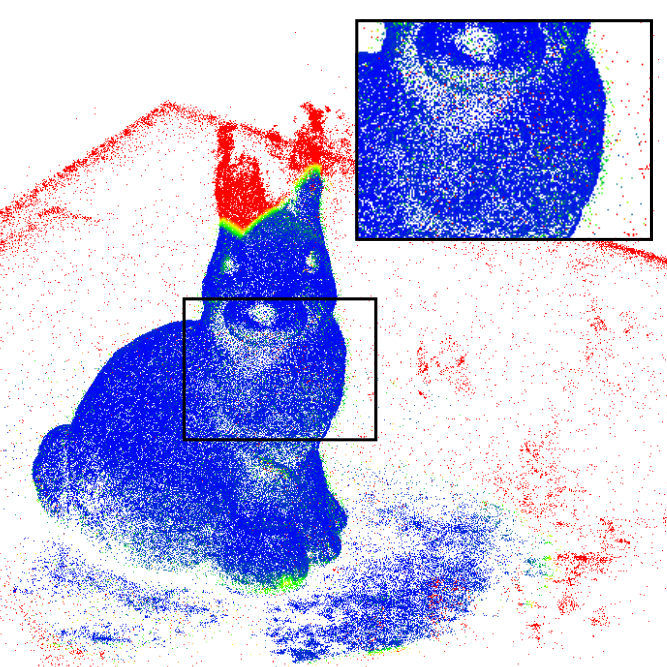} &
        \includegraphics[width=0.15\textwidth]{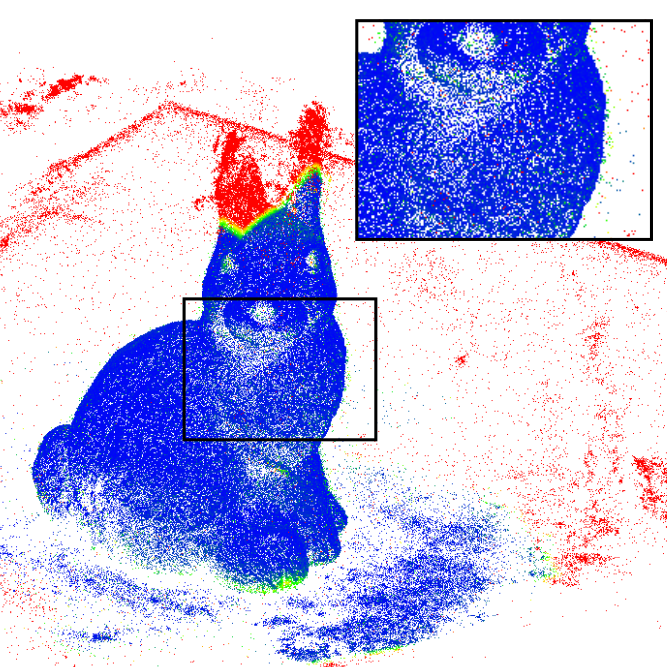} &
        \includegraphics[width=0.15\textwidth]{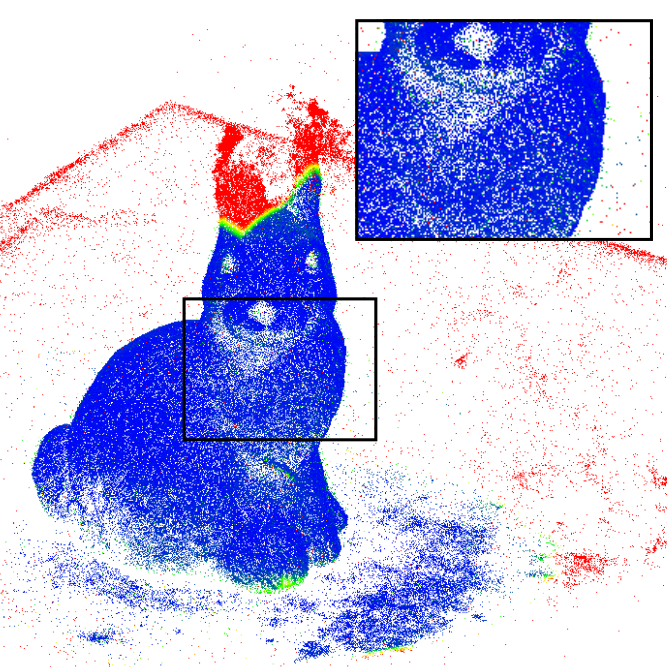} &
        \includegraphics[width=0.15\textwidth]{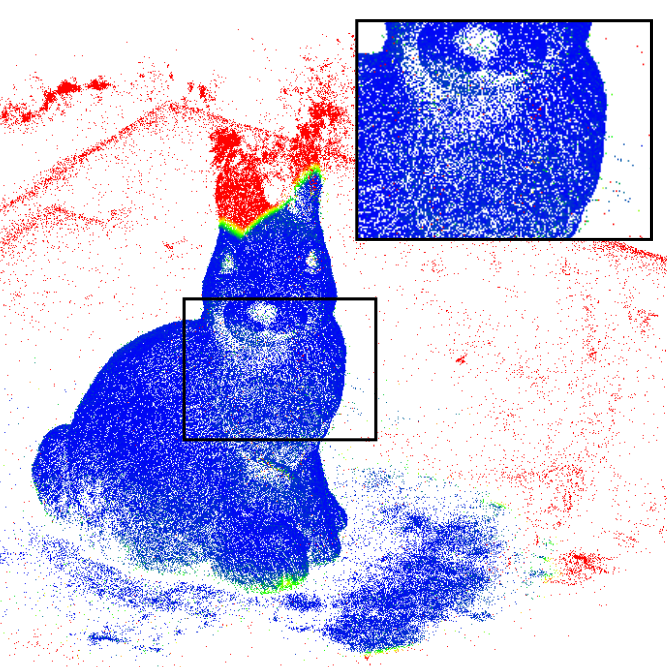} \\[2ex]
    \end{tabular}
    \hspace{15mm}
\includegraphics[width=0.2\textwidth]{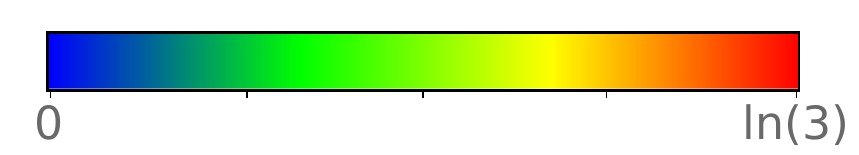}
\hspace{25mm}
\includegraphics[width=0.2\textwidth]{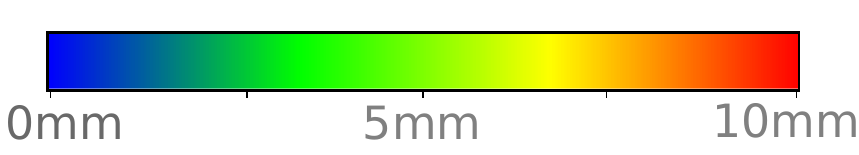}

    \caption{
        Comparison of Eigenentropy and geometric accuracy for DTU scene scan55. 
        Point cloud of 3DGS and our method for mean Eigenentropy with different neighborhood sizes and cloud-to-cloud distance (C2C).
    }
    \label{fig:Eigenentropy_vs_C2C}
\end{figure*}

First, we demonstrate that EntON is effective and that the proposed Eigenentropy-aware densification and pruning strategy actively influences both the Eigenentropy of local Gaussian neighborhoods and the resulting geometric reconstruction accuracy during training process.
The final mean Eigenentropy across all Gaussian neighborhoods is significantly lower when applying our strategy compared to using the 3DGS densification strategy. Table~\ref{tab:Eigenentropy_end} summarizes the per-scene and mean Eigenentropy values after 15\,000 training iterations on the DTU dataset.
The training progression further highlights this effect (Figure~\ref{fig:Eigenentropy_vs_iteration}).  Across all scenes, higher Eigenentropy correlates with increased geometric error, whereas low to medium Eigenentropy aligns with superior surface reconstruction accuracy. In 3DGS, the mean Eigenentropy across all DTU scenes remains persistently high (typically above 0.95, approaching 1.0) and even shows a slight upward trend over time. Correspondingly, geometric accuracy remains limited, with cloud-to-cloud (C2C) distances only marginally improving from approximately 1.75\,mm to around 1.60\,mm. In contrast, EntON causes a rapid decrease in mean Eigenentropy starting from the onset of alternating densification (at iteration 3000). After approximately 5\,000 iterations (i.e., 2\,000 iterations of Eigenentropy-aware densification), the value stabilizes at a consistently lower level between roughly 0.78 and 0.82. The geometric accuracy follows the same trend: upon the start of EntON, the mean C2C distance drops sharply from about 1.75\,mm to approximately 1.20\,mm and continues to decrease steadily, converging to a stable value of around 1.05\,mm after roughly 12\,500 iterations.
This inverse relationship between high-Eigenentropy and high C2C distance is further visualized in Figure~\ref{fig:Eigenentropy_vs_C2C}, by comparing the Eigenentropy point clouds and cloud-to-cloud (C2C) distance point clouds. Our approach consistently achieves lower Eigenentropy in structured regions and reduced geometric error compared to 3DGS, confirming that enforcing of using Gaussians in planar, low-Eigenentropy regions as splitting candidates to enhance low local Eigenentropy promotes both geometric regularity and surface accuracy.

\begin{table}[h!]
\centering
\resizebox{\textwidth}{!}{
\begin{tabular}{lccccccccccccccccc}
\hline
Method & 24 & 37 & 40 & 55 & 63 & 65 & 69 & 83 & 97 & 105 & 106 & 110 & 114 & 118 & 122 & mean \\
\hline
3DGS &  0.95 & 0.96 & 0.99 & 0.97 & 0.86 & 0.90 & 0.92 & 0.95 & 0.91 & 0.96 & 0.96 & 0.95 &  0.96 & 0.95 & 0.95 & 0.96 \\
EntON & 
0.82 & 0.82 & 0.82 & 0.82 & 0.81 & 0.81 & 0.82 & 0.82 & 0.82 & 0.81 & 0.83 & 0.83 & 0.82 & 0.82 & 0.82 & 0.82\\
\hline
\hline
\end{tabular}
}
\caption{Eigenentropy comparison on the DTU dataset: Mean Eigenentropy of all Gaussian centers within a local neighborhoods after the training process using 15\,000 iterations.}
\label{tab:Eigenentropy_end}
\end{table}

\subsection{Small-Scale Data}\label{sec:Results_small}

The following sections present quantitative (Section~\ref{sec:quantitative}) and qualitative (Section~\ref{sec:qualitative}) results of our method EntON in comparison to 3DGS, 2DGS, and PGSR. 
We evaluate the approaches after a fixed number of training iterations and distinguish between four key aspects: geometric surface accuracy, photometric rendering photometric quality, and efficiency of memory and training time.

\subsubsection{Quantitative Results}\label{sec:quantitative}

The following quantitative results, obtained after a fixed training duration of 15\,000 iterations, the geometric surface accuracy via the Chamfer cloud-to-cloud distance, rendering quality in terms of PSNR, the final number of Gaussians, and the total training time. Overall, the performance of the evaluated methods is analyzed with respect to different scene characteristics, including textured, reflective, and rough surfaces or materials.

\paragraph{Geometric Accuracy}\label{sec:geometric_accuracy}

We evaluate the geometric accuracy of the reconstructed surface points using Chamfer cloud-to-cloud (C2C) distance measured against the reference point cloud, considering only points within 10\,mm of the reference (following the standard DTU evaluation). The reported values represent the accuracy of the Gaussian centers, which serve as the geometric backbone for subsequent meshing pipelines.

Table \ref{tab:iterations_c2c_surface_Eigenentropy_knn} summarizes C2C distances (in mm, $\downarrow$) per scene and in average across all evaluated methods after 15\,000 training iterations. EntON achieves the best overall performance when using small neighborhood sizes. In particular, EntON ($k_{\text{nn}}=25$) obtains the lowest mean distance of 0.97\,mm, slightly outperforming PGSR with 1.00\,mm. Larger neighborhood sizes lead to progressively worse accuracy in average: $k_{\text{nn}}=50$ reaches 1.04\,mm, $k_{\text{nn}}=75$ yields 1.13\,mm, the adaptive variant 1.14\,mm, and $k_{\text{nn}}=100$ 1.19\,mm. Both 2DGS with 1.33\,mm and the 3DGS with 1.61\,mm in average are clearly outperformed by our approach and PGSR.

However, the performance remains highly scene dependent. On well-textured surfaces (e.g., scans 40, 55, 106, 118, 122), EntON with small neighborhoods ($k_{\text{nn}}=25$ or $k_{\text{nn}}=50$) consistently produces the most accurate geometry, while its also frequently surpassing PGSR. On reflective or specular surfaces (e.g., scans 63, 97, 110), accuracy degrades noticeably for very small neighborhoods. Likely because too aggressive pruning removes relevant Gaussians in regions with low spatial density of the Gaussians. Larger neighborhoods ($k_{\text{nn}}\geq75$ or adaptive) provide greater robustness in such cases. Although PGSR generally remains more stable under strong reflections and gloss surfaces. Both 3DGS and 2DGS suffer substantially from material properties such as specularity and fine surface structure. On rough-textured surfaces (e.g., scans 83, 105), EntON and PGSR perform comparably, with smaller neighborhoods again tending to yield the best local accuracy.

\definecolor{yellow}{RGB}{255, 255, 204}
\definecolor{orange}{RGB}{255, 204, 153}
\definecolor{red}{RGB}{255, 153, 153}

\begin{table}[h!]
\centering
\resizebox{\textwidth}{!}{
\begin{tabular}{lccccccccccccccccc}
\hline
Method & 24 & 37 & 40 & 55 & 63 & 65 & 69 & 83 & 97 & 105 & 106 & 110 & 114 & 118 & 122 & mean \\
\hline
3DGS                    & 1.82 & 1.84 & 1.72 & 1.55 & 2.17 & 1.70 & 1.66 & 2.37 & 1.90 & 1.71 & 1.50 & 1.51 & 1.47 & 1.30 & 1.34 & 1.61 \\
2DGS                    & 1.25 & 1.51 & 0.97 & 0.73 & 2.03 & 1.46 & 1.26 & 1.97 & 1.76 & 1.51 & 0.78 & 1.37 & 0.94 & 0.69 & 0.75 & 1.33 \\
PGSR                   & 0.92 & 1.15 & 0.77 & 0.67 & 1.43 & 0.99 & 0.97 & 1.28 & 1.79 & 1.13 & 0.60 & 0.96 & 0.66 & 0.68 & 0.79 & \cellcolor{orange}1.00 \\
\midrule
EntON($knn_{adaptive}$) & 1.13 & 1.22 & 0.79  & 0.57 & 1.73 & 1.13 & 1.06 & 1.82 & 1.79 & 1.27 & 0.72 & 1.09 & 0.73 & 0.66 & 0.71 & 1.14\\
EntON($knn_{100}$)      & 1.17 & 1.32 & 0.88  & 0.62 & 1.77 & 1.23 & 1.12 & 1.85 & 1.80 & 1.32 & 0.82 & 1.11 & 0.80 & 0.73 & 0.77 & 1.19\\
EntON($knn_{75}$)       & 1.13 & 1.29 & 0.82  & 0.58 & 1.68 & 1.15 & 1.05 & 1.71 & 1.79 & 1.24 & 0.76 & 1.14 & 0.75 & 0.68 & 0.72 & 1.13\\
EntON($knn_{50}$)       & 1.08 & 1.21 & 0.76  & 0.55 & 1.60 & 1.06 & 0.98 & 1.59 & 1.79 & 1.21 & 0.69 & 1.12 & 0.68 & 0.62 & 0.67 & \cellcolor{yellow}1.04\\
EntON($knn_{25}$)       & 0.90 & 1.12 & 0.70  & 0.51 & 1.62 & 0.95 & 0.90 & 1.41 & 1.81 & 1.09 & 0.56 & 1.21 & 0.62 & 0.55 & 0.61 & \cellcolor{red}0.97 \\

\hline
\hline
\end{tabular}
}
\caption{Surface accuracy. \textbf{Geometric accuracy} comparison on the DTU dataset with Chamfer cloud-to-cloud distances $\downarrow$ in mm for points $\leq$10\,mm from the reference, according to the DTU evaluation script. Best results are highlighted as \colorbox{red}{1st}, \colorbox{orange}{2nd}, and \colorbox{yellow}{3rd}. Mean scores are listed at the bottom. The training incorporates 15\,000 iterations. }
\label{tab:iterations_c2c_surface_Eigenentropy_knn}
\end{table}

\paragraph{Rendering Quality}

We assess rendering quality using peak signal-to-noise ratio (PSNR), while higher PSNR values indicate better photometric fidelity. Table \ref{tab:iterations_psnr_Eigenentropy_knn} reports the PSNR values (in dB, $\uparrow$) per scene and the overall mean. Our method consistently achieves competitive or superior rendering quality compared to all baselines. In particular, EntON ($k_{\text{nn}}=100$) and \textbf{EntON ($k_{\text{nn}}=75$)} deliver the highest average PSNR of 34.71\,dB and 34.75\,dB, respectively, slightly surpassing 3DGS (34.84\,dB) in several configurations.
On well-textured scenes (e.g., scans 24, 40, 55, 65, 69, 106, 114, 118, 122), EntON matches or exceeds the rendering quality of 3DGS across most neighborhood sizes, often producing the highest per-scan PSNR values. Larger neighborhoods ($k_{\text{nn}}=75$ or $k_{\text{nn}}=100$) tend to provide the best overall photometric fidelity. Smaller neighborhoods ($k_{\text{nn}}=25$ or $k_{\text{nn}}=50$) still yield very strong results but show slightly lower average PSNR, particularly on scenes with challenging reflective surfaces or low textured surfaces. In contrast, 2DGS (32.54\,dB), and PGSR (32.32\,dB) exhibit noticeably lower rendering quality, often suffering from detail loss. 

\begin{table}[h!]
\centering
\small
\setlength{\tabcolsep}{4pt}
\resizebox{\textwidth}{!}{
\begin{tabular}{lcccccccccccccccc}
\hline
Method     & 24    & 37     & 40    & 55   & 63    & 65 & 69 & 83 & 97 & 105 & 106 & 110 & 114 & 118 & 122 & mean \\
\hline
3DGS                & 35.86 & 30.28 & 34.59 & 34.06 & 37.22 & 34.02 & 33.28 & 34.51 & 33.53 & 37.59 & 37.97 & 31.55 & 35.39 & 36.47 & 36.33 &  \cellcolor{red}34.84 \\
2DGS                & 32.69 & 28.51 & 31.74 & 32.13 & 35.10 & 31.92 & 31.01 & 31.51 & 31.27 & 34.15 & 35.92 & 30.17 & 32.60 & 35.06 & 34.38 & 32.54 \\
PGSR                & 31.77 & 27.67 & 30.92 & 31.99 & 34.20 & 31.82 & 30.15 & 30.71 & 30.83 & 34.63 & 36.00 & 30.90 & 32.55 & 35.84 & 34.80 & 32.32 \\
\midrule
EntON($knn_{adaptive}$) & 35.38 & 28.70 & 33.51 & 35.32 & 37.01 & 35.66 & 32.05 & 35.05 & 33.02 & 37.49 & 37.74 & 30.46 & 34.46 & 36.24 & 35.99  & 34.60\\
EntON($knn_{100}$)      & 34.36 & 29.44 & 34.51 & 34.50 & 36.87 & 35.93 & 31.47 & 35.36 & 32.95 & 37.24 & 37.86 & 30.52 & 34.06 & 36.32 & 36.24  & \cellcolor{yellow}34.71\\
EntON($knn_{75}$)       & 34.68 & 28.72 & 34.90 & 35.13 & 36.45 & 34.94 & 31.60 & 34.94 & 32.93 & 37.42 & 37.87 & 30.25 & 34.47 & 35.94 & 36.47  & \cellcolor{orange}34.75\\
EntON($knn_{50}$)       & 34.64 & 29.14 & 34.34 & 34.71 & 36.23 & 35.79 & 31.88 & 35.39 & 32.64 & 37.47 & 37.40 & 29.98 & 34.01 & 36.19 & 36.03  & 34.39\\
EntON($knn_{25}$)       & 34.36 & 29.09 & 34.25 & 34.96 & 36.56 & 35.33 & 31.64 & 35.39 & 31.47 & 37.39 & 37.19 & 29.65 & 33.63 & 35.76 & 36.14  & 34.19 \\

\hline
\end{tabular}
}
\caption{\textbf{Rendering quality} comparison on the DTU dataset. We report the PSNR $\uparrow$ in dB. Mean scores are listed. The training incorporates 15\,000 iterations. Best results are highlighted as \colorbox{red}{1st}, \colorbox{orange}{2nd}, and \colorbox{yellow}{3rd}}
\label{tab:iterations_psnr_Eigenentropy_knn}
\end{table}

\paragraph{Efficiency}

We evaluate the efficiency of memory and training time, by measuring the final number of Gaussians and training time. A lower number of Gaussians directly correlates with reduced memory consumption (and faster rendering). Table \ref{tab:iterations_gaussians} reports the number of Gaussians ($\downarrow$) per scene and in average. 3DGS produces by far the largest number of Gaussians on average (392\,129), indicating a highly redundant and uncompressed representation. In comparison, 2DGS (208\,572), PGSR (232\,004), and EntON achieve substantially more compact scene representations.

Among our variants, smaller neighborhood sizes lead to significantly fewer Gaussians while preserving or even improving geometric and rendering quality (as shown in previous sections). In particular, EntON ($k_{\text{nn}}=25$) achieves the lowest average number of Gaussians at 157\,391, approximately 60\% fewer than 3DGS and clearly outperforming all baselines in compactness. Larger neighborhoods result in higher Gaussian counts: $k_{\text{nn}}=50$ yields 187\,759, $k_{\text{nn}}=75$ 211\,925, $k_{\text{nn}}=100$ 233\,116, and the adaptive variant 228\,578 Gaussians.
All our neighborhood sizes remain more compact than 3DGS and are competitive with or better than 2DGS and PGSR, especially at smaller $k_{\text{nn}}$ values. Notably, even our most compact variant ($k_{\text{nn}}=25$) starts from the same sparse SfM initialization (average 22\,771 points) but efficiently densifies only where necessary, avoiding the excessive densification typical of 3DGS.

\begin{table}[h!]
\centering
\resizebox{\textwidth}{!}{
\begin{tabular}{lcccccccccccccccc}
\hline
Method & 24 & 37 & 40 & 55 & 63 & 65 & 69 & 83 & 97 & 105 & 106 & 110 & 114 & 118 & 122 & Mean \\
\hline
3DGS      
& 611\,749 & 695\,233 & 778\,308 & 622\,019 & 223\,472 & 233\,292 & 249\,407 & 200\,975 & 556\,486 & 238\,905 & 215\,615 & 200\,024 & 276\,819 & 291\,762 & 266\,299 & 392\,129 \\

2DGS      
& 317\,153 & 391\,404 & 354\,076 & 341\,317 & 151\,758 & 159\,384 & 169\,785 & 123\,622 & 240\,807 & 127\,348 & 105\,397 & 102\,635 & 132\,201 & 160\,033 & 144\,754 & \cellcolor{yellow}208\,572 \\

PGSR      
& 338\,583 & 368\,363 & 394\,813 & 376\,698 & 127\,514 & 221\,565 & 253\,433 & 122\,908 & 211\,997 & 205\,436 & 134\,614 & 175\,274 & 166\,209 & 211\,044 & 195\,750 & 232\,004 \\
\midrule
EntON($knn_{adaptive}$) & 328401 & 370351 & 414967 & 373220 & 156504 & 224221 & 179262 & 140561 & 260293 & 186702 & 125623 & 116045 & 155316 & 171382 & 164187 & 228578\\
EntON($knn_{100}$)      & 329222 & 403205 & 409569 & 365176 & 156361 & 207048 & 173000 & 151514 & 285136 & 185796 & 136097 & 114874 & 159414 & 175780 & 169528 & 233116\\
EntON($knn_{75}$)       & 316775 & 346756 & 409775 & 366452 & 150623 & 184157 & 153795 & 148258 & 287247 & 173638 & 130814 & 104196 & 163856 & 165875 & 158206 & 211925\\
EntON($knn_{50}$)       & 283176 & 331361 & 377030 & 331516 & 134875 & 177036 & 155980 & 129713 & 242596 & 164645 & 116202 & 88650  & 140847 & 153449 & 148484 & \cellcolor{orange}187759\\
EntON($knn_{25}$)       & 257560 & 309748 & 352715 & 313075 & 116733 & 162947 & 152141 & 115340 & 175660 & 147020 & 104522 & 72692  & 121997 & 136422 & 132683 & \cellcolor{red}157391\\

Initial SfM
& 15\,479 & 24\,857 & 39\,158 & 33\,506 & 10\,869 & 13\,203 & 15\,264 & 10\,652 & 20\,467 & 25\,291 & 33\,523 & 11\,382 & 25\,761 & 27\,650 & 20\,975 & 22\,771 \\
\hline
\end{tabular}
}
\caption{\textbf{Number of Gaussians} on the DTU dataset. We report the total number of Gaussians $\downarrow$ resulting from the EntON, compared to 3DGS, 2DGS, PGSR, and the number of SfM points used for initialization. The training incorporates 15\,000 iterations. Mean scores are listed. Best results (low total number is better) are highlighted as \colorbox{red}{1st}, \colorbox{orange}{2nd}, and \colorbox{yellow}{3rd}.}
\label{tab:iterations_gaussians}
\end{table}

In addition to memory efficiency, we compare the training time required by each method to complete 15\,000 iterations, serving as a indicator for computational efficiency. Table \ref{tab:dtu_time} presents training time per scene and in average. Our method consistently achieves the fastest training time across all neighborhood sizes, with smaller neighborhoods yielding the most significant speedups. In particular, EntON ($k_{\text{nn}}=25$) requires only 9.14 minutes on average, which is approximately 29\% faster than 3DGS with 12.88 minutes and substantially faster than 2DGS with 18.11 and PGSR with 32.89minutes .
Larger neighborhoods incur modestly higher training time due to increased per-iteration costs from processing more neighest neighbors: $k_{\text{nn}}=50$ averages 9.91\,min, $k_{\text{nn}}=75$ reaches 10.21\,min, $k_{\text{nn}}=100$ and the adaptive variant both require 10.52\,min. Nevertheless, even the slowest of our configurations remains markedly faster than the baselines.

\begin{table}[h!]
\centering
\resizebox{\textwidth}{!}{
\begin{tabular}{lcccccccccccccccc}
\hline
Method & 24 & 37 & 40 & 55 & 63 & 65 & 69 & 83 & 97 & 105 & 106 & 110 & 114 & 118 & 122 & Mean \\
\hline
3DGS      
& 18.48 & 15.61 & 15.95 & 13.05 & 11.86 & 11.97 & 11.85 & 11.22 & 14.20 & 11.88 & 11.03 & 10.38 & 11.84 & 11.02 & 10.93 & 12.88 \\

2DGS      
& 21.87 & 20.24 & 19.53 & 18.22 & 17.22 & 18.27 & 17.44 & 18.00 & 19.37 & 18.12 & 17.32 & 17.16 & 18.17 & 17.19 & 17.51 & 18.11 \\

PGSR      
& 33.45 & 30.04 & 32.58 & 28.86 & 27.55 & 27.04 & 31.04 & 28.08 & 28.38 & 40.17 & 38.07 & 38.00 & 40.03 & 22.20 & 37.34 & 32.89 \\
\midrule
EntON($knn_{adaptive}$) & 12.35 & 10.97 & 11.91 & 10.76 & 10.06 & 9.60 & 9.57 & 9.07 & 9.77 & 9.46 & 9.28 & 9.09 & 9.36 & 9.23 & 9.28 & 10.52\\
EntON($knn_{100}$)      & 13.02 & 11.77 & 12.74 & 11.18 & 10.33 & 9.75 & 9.80 & 9.33 & 10.26 & 9.71 & 9.80 & 8.80 & 9.66 & 9.67 & 9.95 & 10.52\\
EntON($knn_{75}$)       & 12.49 & 11.27 & 12.34 & 10.68 & 10.17 & 9.54 & 9.65 & 9.39 & 9.93 & 9.41 & 9.78 & 8.73 & 9.80 & 9.18 & 9.34 & \cellcolor{yellow}10.21\\
EntON($knn_{50}$)       & 11.78 & 10.62 & 11.42 & 10.37 & 9.87 & 9.31 & 9.01 & 9.07 & 9.43 & 9.25 & 9.27 & 8.51 & 9.14 & 9.09 & 8.91 &  \cellcolor{orange}9.91\\
EntON($knn_{25}$)       & 10.97 & 10.15 & 10.95 & 10.09 & 9.45 & 9.02 & 9.13 & 9.08 & 9.10 & 9.18 & 8.64 & 8.09 & 8.76 & 8.75 & 8.83 & \cellcolor{red}9.14\\

\hline
\end{tabular}
}
\caption{\textbf{Training time} comparison on the DTU dataset. We report the minutes for 15\,000 iterations in min. Mean scores are listed. Best results (low training time is better) are highlighted as \colorbox{red}{1st}, \colorbox{orange}{2nd}, and \colorbox{yellow}{3rd}. The training incorporates 15\,000 iterations.}
\label{tab:dtu_time}
\end{table}

Our Eigenentropy-aware Gaussian densification approach, guided local neighborhood of Gaussians, consistently delivers a compelling balance across the four key metrics: rendering quality, geometric accuracy, memory efficiency, and training time.

\begin{figure}[h!]
    \centering
        \includegraphics[width=0.8\linewidth]{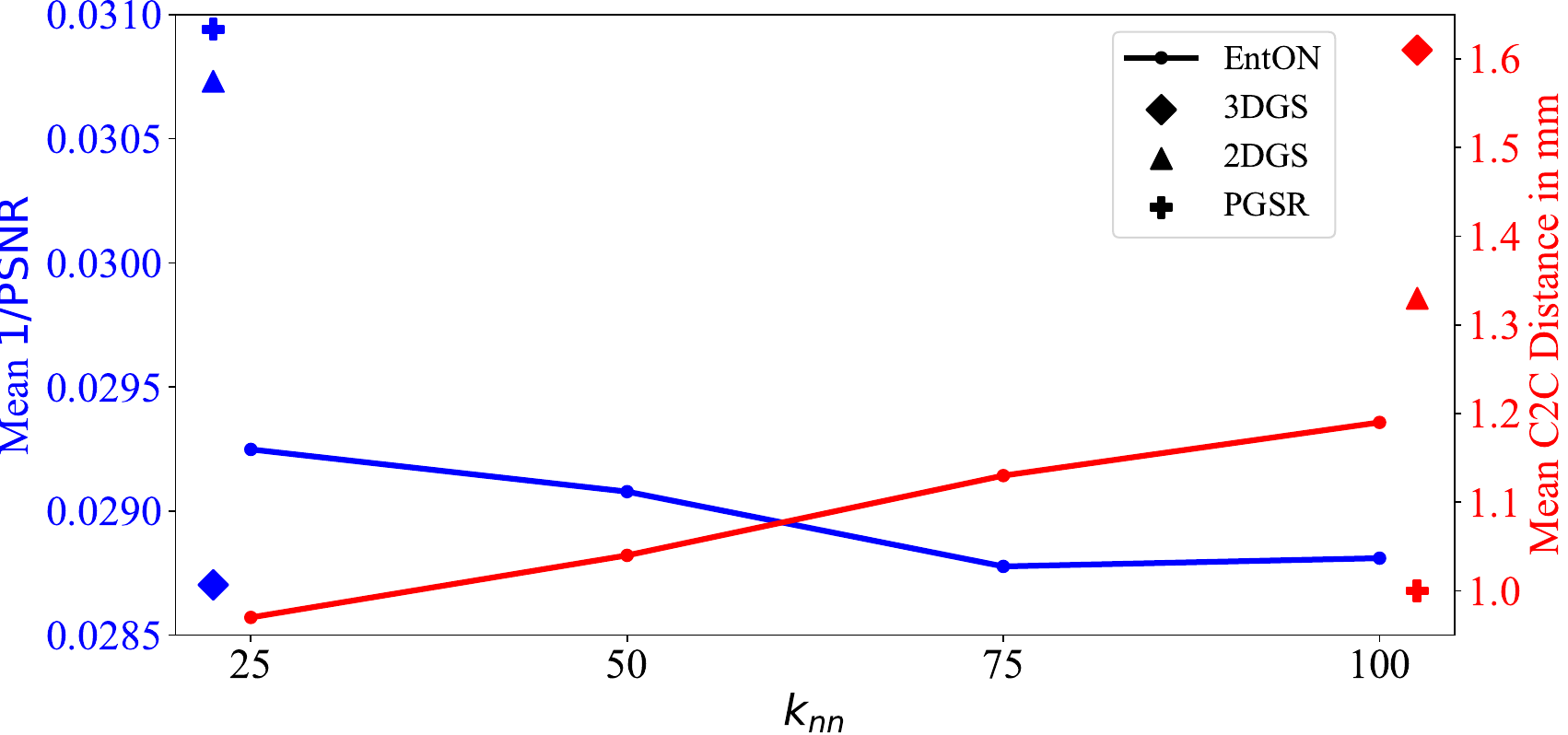}
 \caption{Trend of the mean PSNR (plotted as its inverse to illustrate the trade-off) and mean cloud to cloud (C2C) distance in EntON of all DTU scenes over different neighborhood sizes $k_{\text{nn}}$, in comparison to 3DGS, 2DGS and PGSR.}
    \label{fig:knn_tradeoff}
\end{figure}

The highest average rendering quality is achieved with larger neighborhood sizes: $k_{\text{nn}}=75$ reaches 34.75\,dB and $k_{\text{nn}}=100$ yields 34.71\,dB, approaching or matching 3DGS (34.84\,dB) while using 40 to 46\,\% fewer Gaussians. Smaller neighborhoods cause a slight PSNR reduction ($k_{\text{nn}}=25$: 34.19\,dB; $k_{\text{nn}}=50$: 34.39\,dB) but remain substantially superior to 2DGS (32.54\,dB) and PGSR (32.32\,dB). In contrast, geometric accuracy strongly favors smaller neighborhoods. The best mean C2C distance is obtained with $k_{\text{nn}}=25$ at 0.97\,mm, outperforming PGSR (1.00\,mm) and significantly surpassing 2DGS (1.33\,mm) and 3DGS (1.61\,mm). Larger neighborhoods show progressively higher errors ($k_{\text{nn}}=50$: 1.04\,mm; $k_{\text{nn}}=75$: 1.13\,mm; $k_{\text{nn}}=100$: 1.19\,mm). Consumption of memory and training time follow a similar trend. The most compact representation is achieved at $k_{\text{nn}}=25$ with only 157\,k Gaussians on average, a reduction of approximately 60\,\% compared to 3DGS (392\,k), and the shortest training time of 9.14\,min (29\,\% faster than 3DGS at 12.88\,min). Larger neighborhoods scale toward 212 to 233\,k Gaussians and 10.2 to 10.5\,min, remaining markedly more efficient than both 2DGS and PGSR across all configurations.
Figure \ref{fig:knn_tradeoff} illustrates the small trade-off between geometric accuracy and rendering quality. For EntON, geometric accuracy improves with smaller neighborhood sizes, whereas rendering quality benefits from larger neighborhoods. The most balanced trade-off is observed approximately at the intersection of the curves, around $k_{\text{nn}}=50$ (or even $k_{\text{nn}}=75$). It is apparent that 2DGS, and particularly PGSR, achieve good geometric accuracy but exhibit notable weaknesses in rendering quality. In contrast, 3DGS displays the opposite behavior, providing high rendering quality at the expense of geometric precision.
Using $k_{\text{nn}}=50$ retains very strong rendering quality (34.39\,dB), delivers excellent geometric accuracy (1.04\,mm, corresponding to a 35.4\,\% improvement over 3DGS), achieves substantial compression (188\,k Gaussians, 52\,\% fewer than 3DGS), and maintains fast training (9.91\,min, 23\,\% faster than 3DGS). It therefore provides near-state-of-the-art performance in all four aspects simultaneously.

\subsubsection{Qualitative Results}\label{sec:qualitative}
Similar to the quantitative results, EntON yields promising qualitative results in terms of geometric accuracy of the 3D point clouds (Section \ref{sec:quali_c2c}) and rendering quality (Section \ref{sec:quali_rendering}). EntON, consistently accurate and photometric qualitative results are generated across all 15 scenes, compared to 3DGS, 2DGS and PGSR.

\paragraph{Geometric Accuracy}\label{sec:quali_c2c}

\begin{figure*}[htbp]
\vspace{-3cm}
    \centering
    \begin{tabular}{c c c c c}
        \textbf{} & \scriptsize\textbf{3DGS} & \scriptsize\textbf{2DGS} & \scriptsize\textbf{PGSR} & \scriptsize\textbf{EntON} \\[1ex]
        
        \rotatebox{90}{\scriptsize\textbf{scan24}} &
         \vspace{-2mm}
        \includegraphics[width=0.225\linewidth]{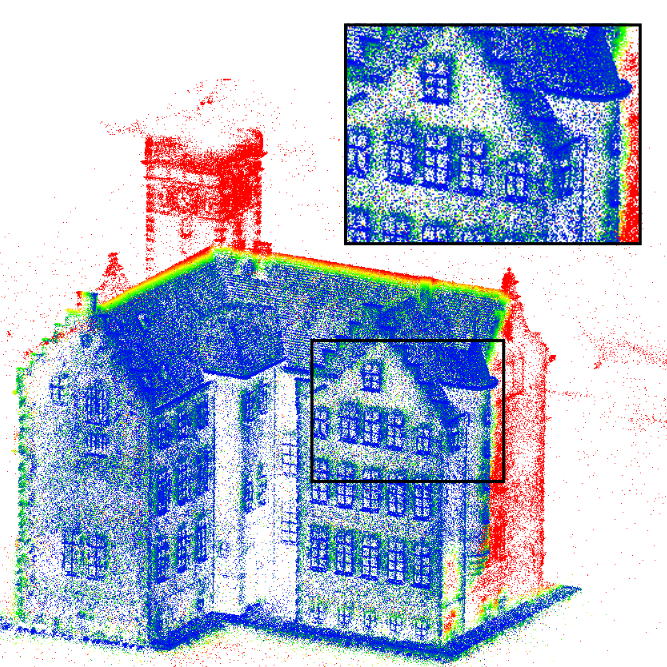} &
         \hspace{-4mm}
        \includegraphics[width=0.225\linewidth]{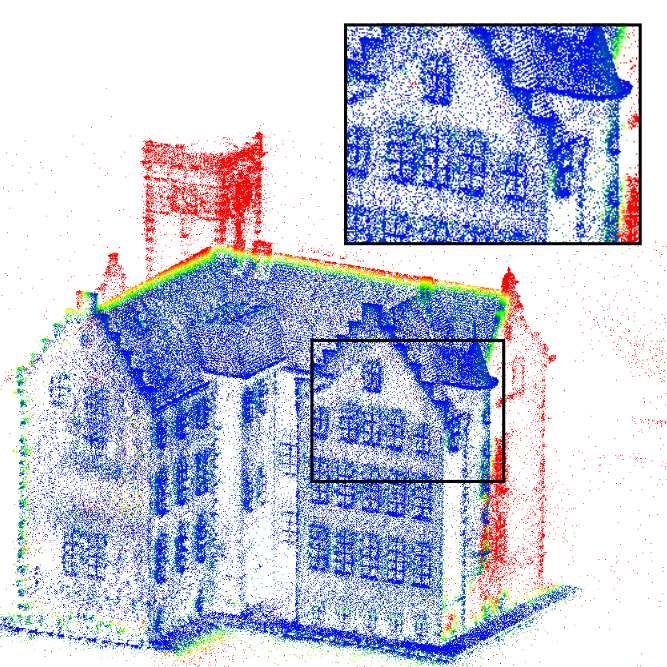} &
        \hspace{-4mm}
        \includegraphics[width=0.225\linewidth]{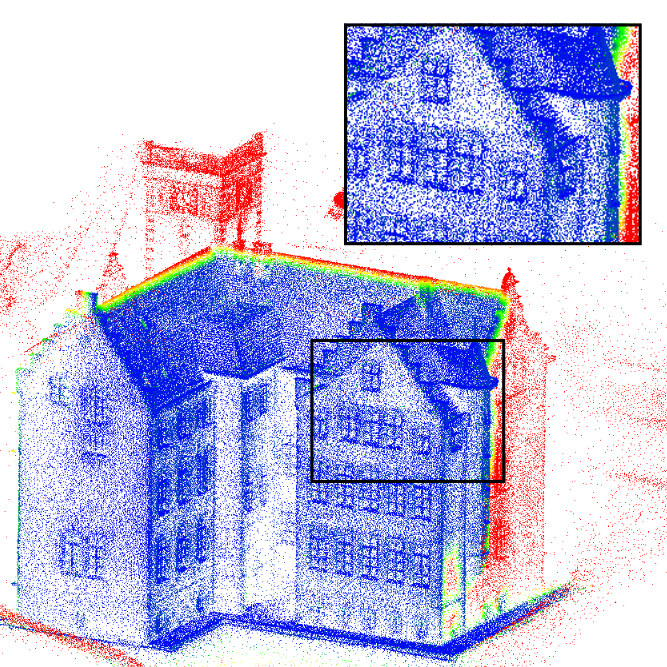} &
         \hspace{-4mm}
        \includegraphics[width=0.225\linewidth]{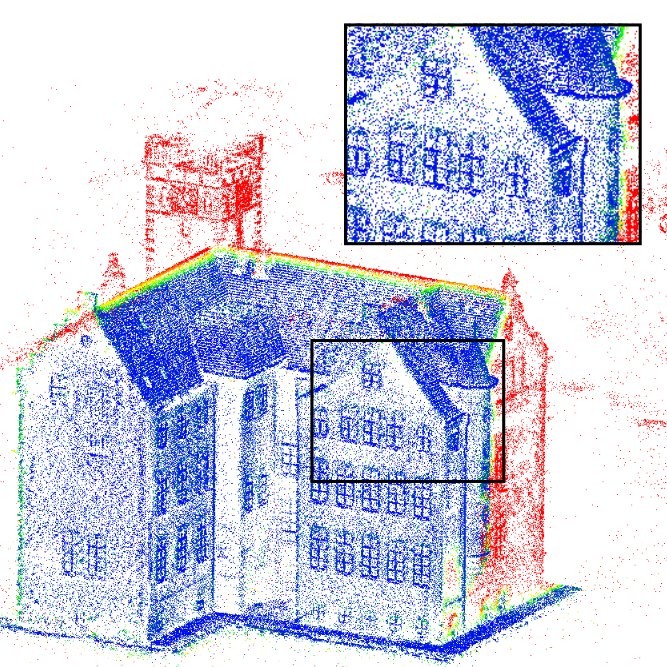} 
  \vspace{-2mm}\\[2ex]
       
        \rotatebox{90}{\scriptsize\textbf{scan37}} &
         \vspace{-2mm}
        \includegraphics[width=0.225\linewidth]{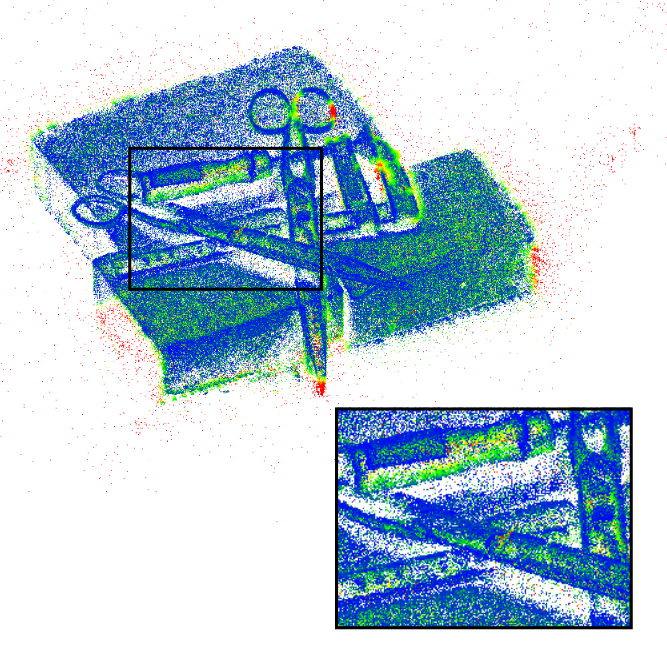} &
         \hspace{-4mm}
        \includegraphics[width=0.225\linewidth]{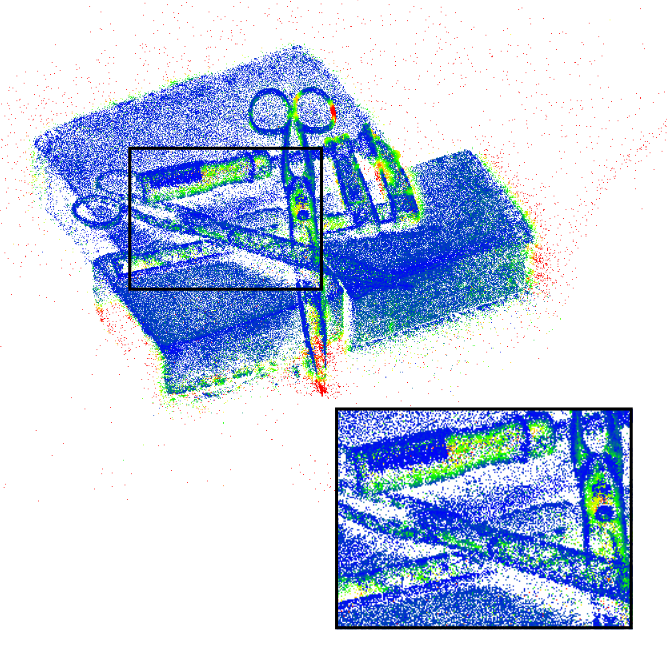} &
        \hspace{-4mm}
        \includegraphics[width=0.225\linewidth]{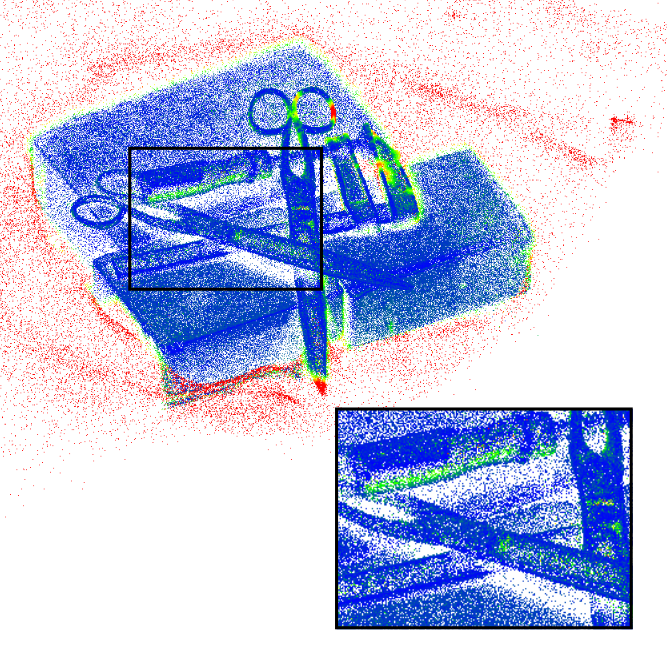} &
        \hspace{-4mm}
        \includegraphics[width=0.225\linewidth]{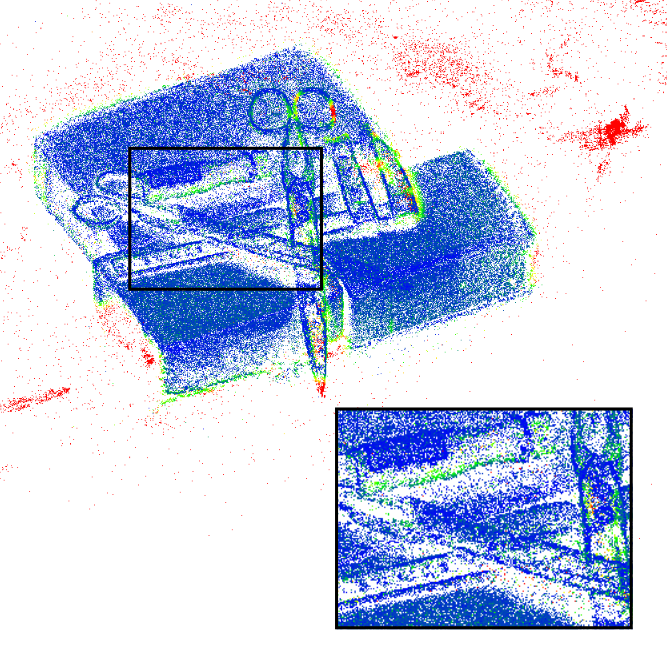} 

        \vspace{-2mm}\\[2ex]
    
        \rotatebox{90}{\scriptsize\textbf{scan40}} &
         \vspace{-2mm}
        \includegraphics[width=0.225\linewidth]{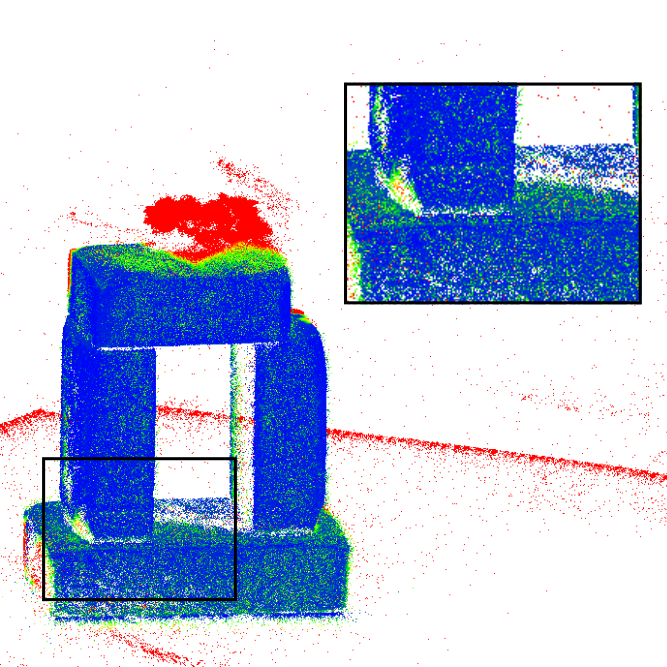} &
         \hspace{-4mm}
        \includegraphics[width=0.225\linewidth]{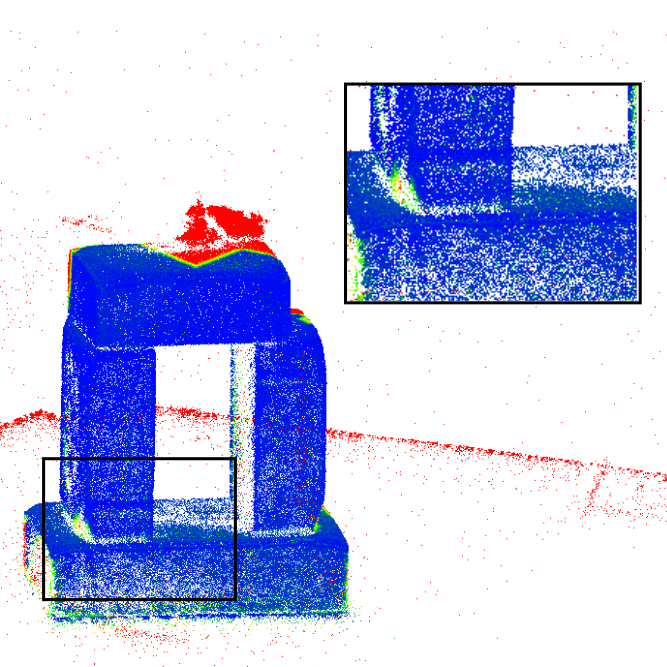} &
        \hspace{-4mm}
        \includegraphics[width=0.225\linewidth]{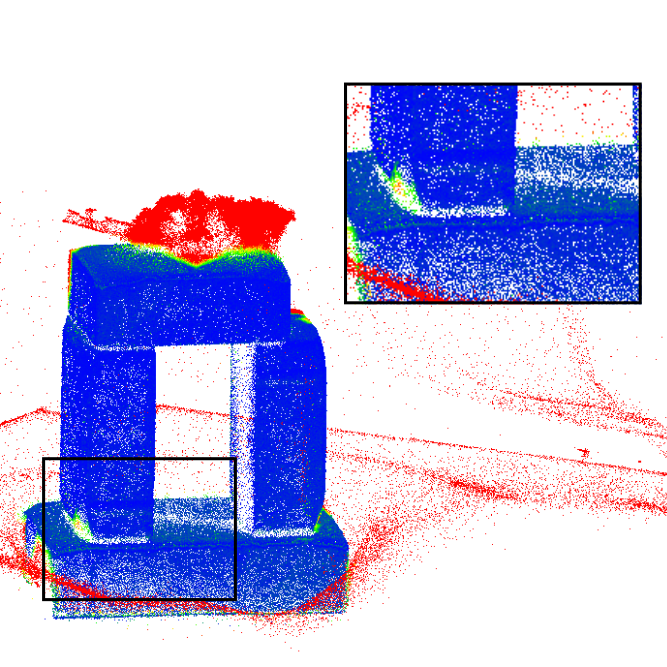} &
          \hspace{-4mm}
        \includegraphics[width=0.225\linewidth]{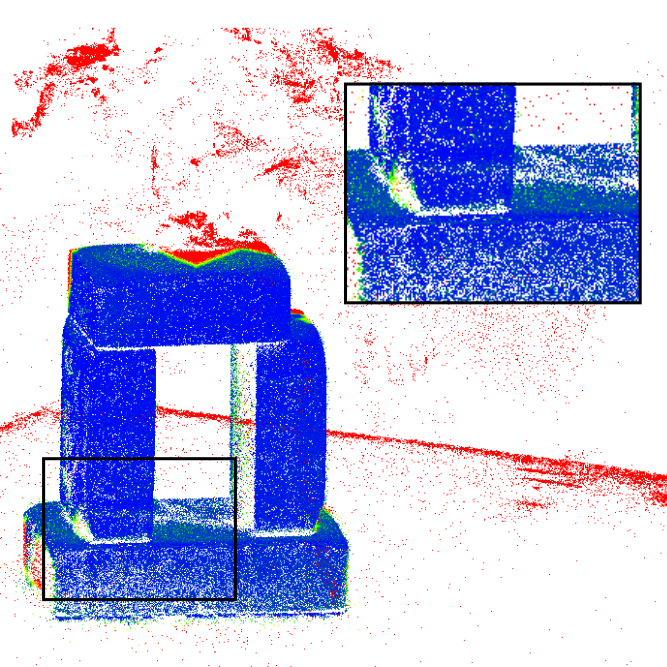} 
      \vspace{-2mm}
        \\[2ex]
       
        \rotatebox{90}{\scriptsize\textbf{scan55}} &
         \vspace{-2mm}
        \includegraphics[width=0.225\linewidth]{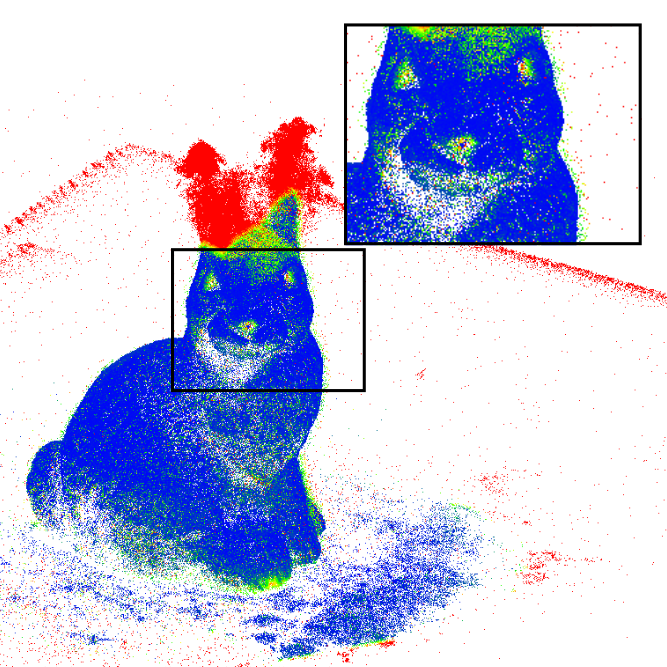} &
         \hspace{-4mm}
        \includegraphics[width=0.225\linewidth]{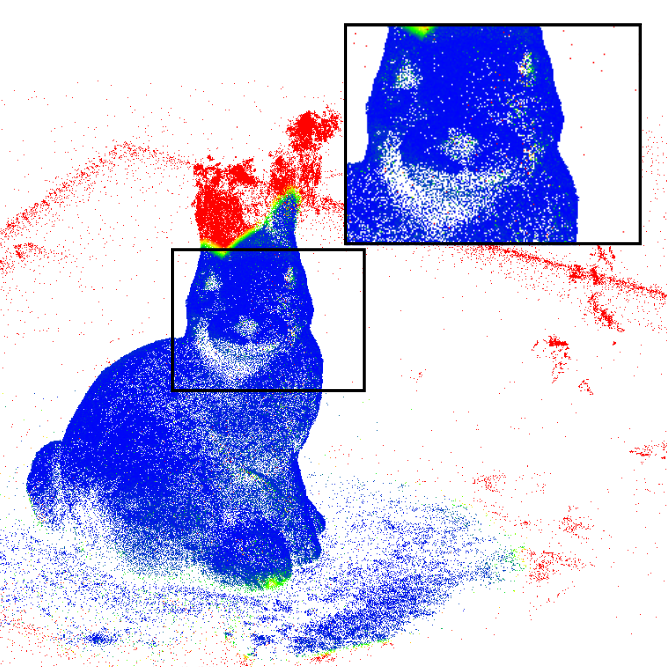} &
        \hspace{-4mm}
        \includegraphics[width=0.225\linewidth]{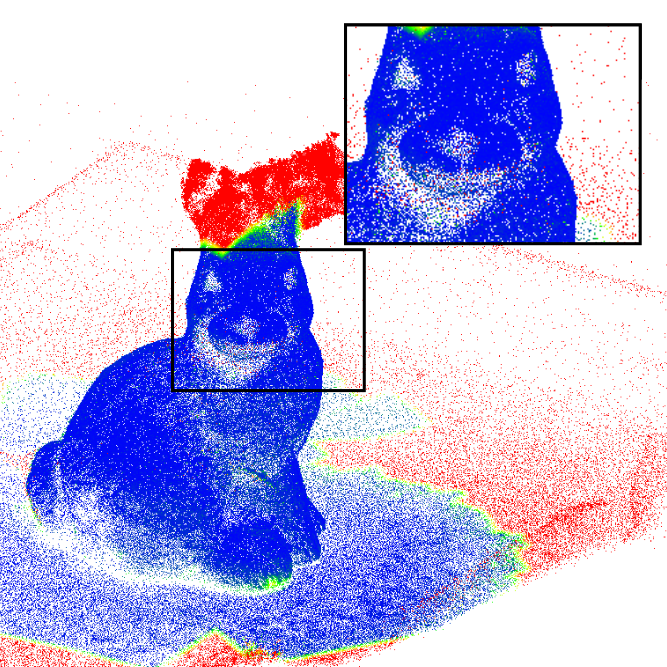} &
        \hspace{-4mm}
        \includegraphics[width=0.225\linewidth]{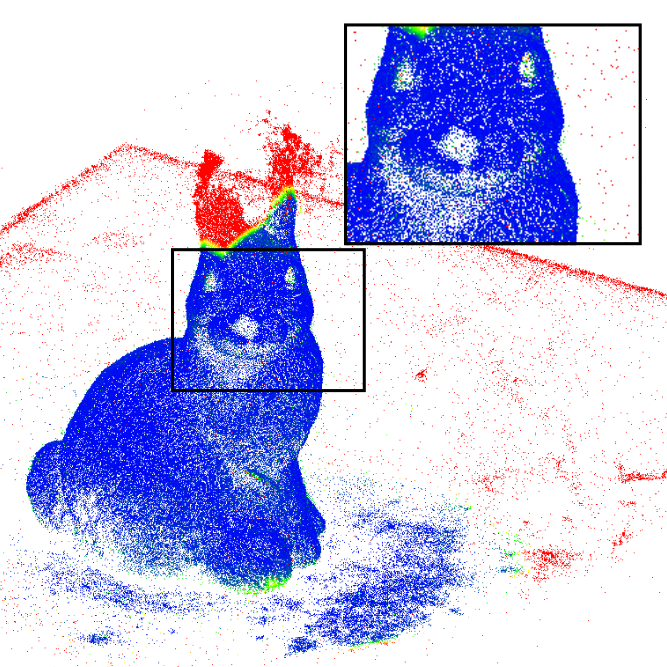} 
      \vspace{-2mm}\\[2ex]
       
         \rotatebox{90}{\scriptsize\textbf{scan114}} &
         \vspace{-2mm}
        \includegraphics[width=0.225\linewidth]{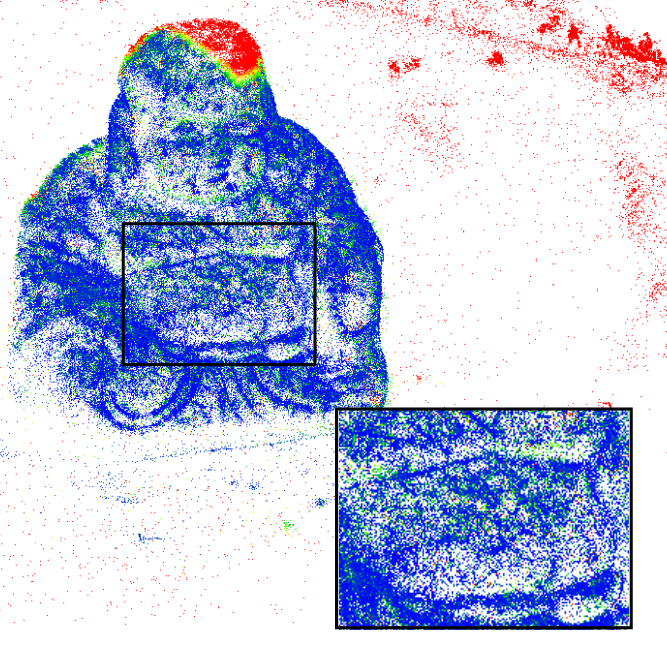} &
         \hspace{-4mm}
        \includegraphics[width=0.225\linewidth]{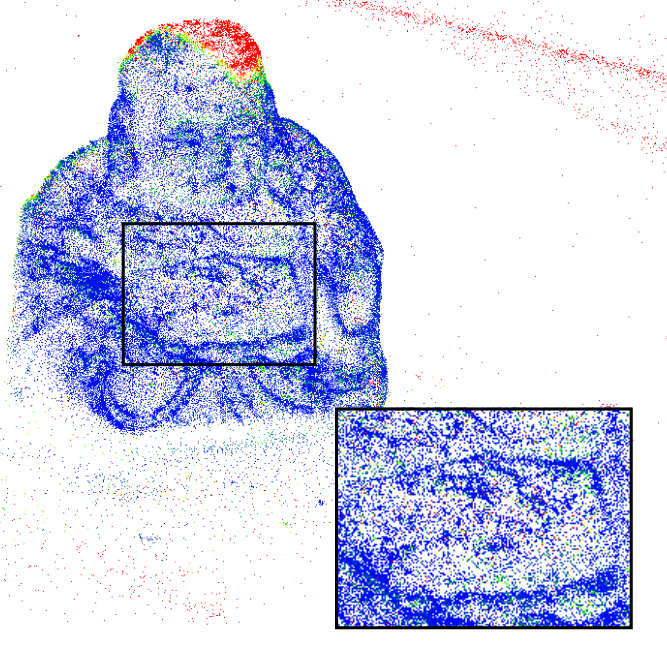} &
        \hspace{-4mm}
        \includegraphics[width=0.225\linewidth]{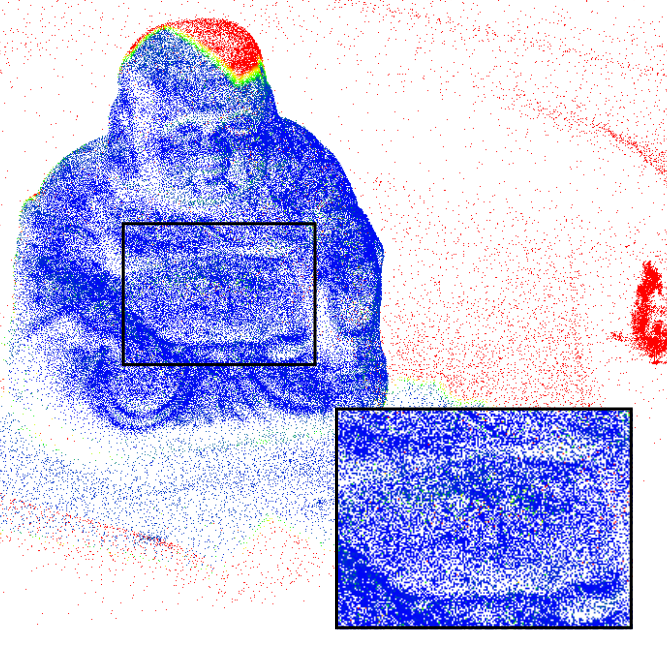} &
        \hspace{-4mm}
        \includegraphics[width=0.225\linewidth]{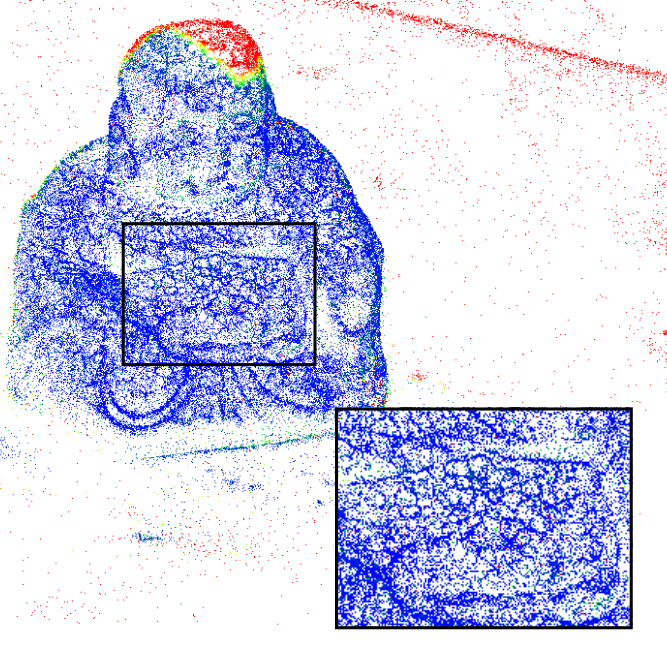} 
        \vspace{-2mm}\\[2ex]

         \rotatebox{90}{\scriptsize\textbf{scan122}} &
         \vspace{-2mm}
        \includegraphics[width=0.225\linewidth]{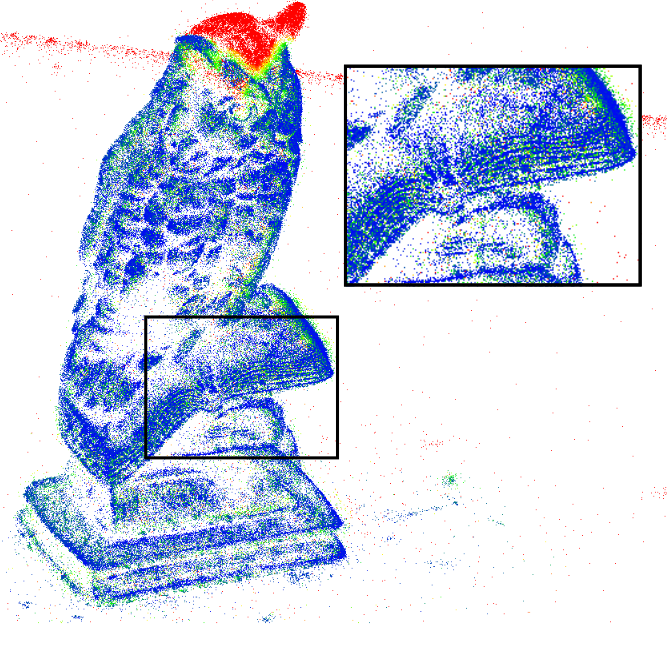} &
         \hspace{-4mm}
        \includegraphics[width=0.225\linewidth]{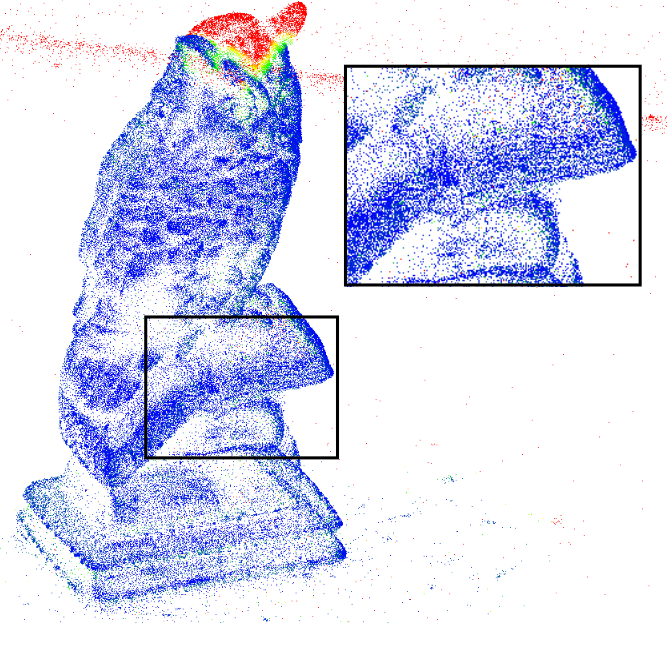} &
        \hspace{-4mm}
        \includegraphics[width=0.225\linewidth]{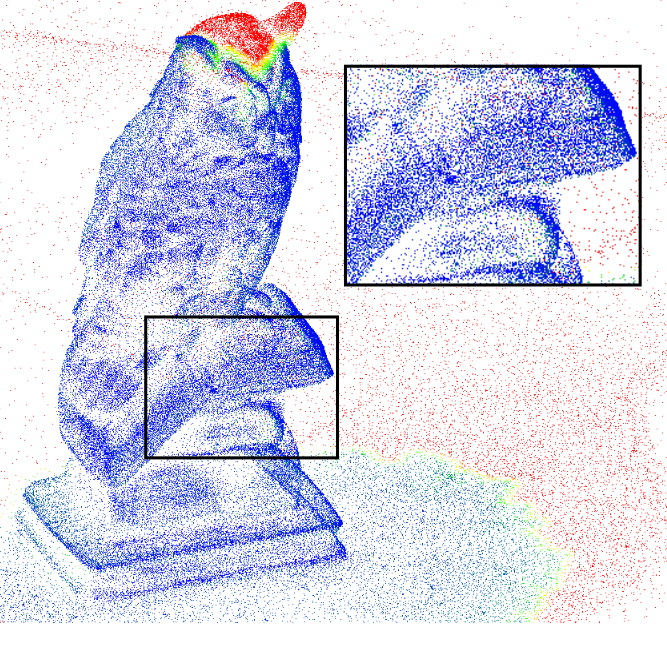} &
        \hspace{-4mm}
        \includegraphics[width=0.225\linewidth]{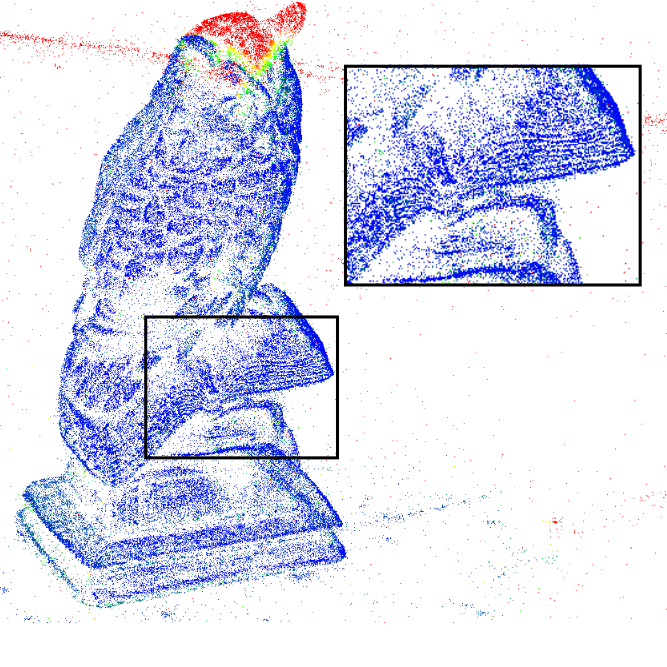} 
      \hspace{-2mm}
      \\[2ex]

    \end{tabular}

    \includegraphics[width=0.3\textwidth]{Figures/colorbar_c2c.pdf} 
    \caption{\textbf{Geometric accuracy} comparison of 3DGS, 2DGS, PGSR and EntON on the DTU dataset with Chamfer cloud-to-cloud distances $\downarrow$ for the same PSNR. Color values are cropped at 10mm distance.}
     \label{fig:Qualitative_c2c}
\end{figure*}

The geometric accuracy of Gaussian centers on DTU dataset, evaluated using the Chamfer cloud-to-cloud distance (Figure \ref{fig:Qualitative_c2c}), highlights the accurate geometric performance of EntON. Note that the reference point clouds are incomplete, which leads to high values on the object edges.
The results demonstrate high surface accuracy for EntON and confirm the quantitative findings. In addition, the Gaussian distributions produced by EntON exhibit sharp and well-defined structures, indicating efficient splitting along object surfaces. This allows splitting of Gaussians with lower view-space position gradients when they are located on the surface, which in turn enhances high-quality, sharp renderings with fine geometric details and pronounced object boundaries.
In contrast, the results obtained with 2DGS and PGSR achieve strong geometric accuracy with a more uniform point distribution. However, the resulting point clouds appear smoother, and object boundaries are partially blurred, leading to less sharply defined surface edges.

\paragraph{Rendering Quality}\label{sec:quali_rendering}

\begin{figure*}[htbp]
\hspace{-2cm}
    \centering
    \begin{tabular}{c c c c c c}
        \textbf{} & \scriptsize\textbf{3DGS} & \scriptsize\textbf{2DGS} & \scriptsize\textbf{PGSR} & \scriptsize\textbf{EntON} & \scriptsize\textbf{GT} \\[1ex]
        
        \rotatebox{90}{\scriptsize\textbf{scan24}} &
         \vspace{-2mm}
        \includegraphics[width=0.225\linewidth]{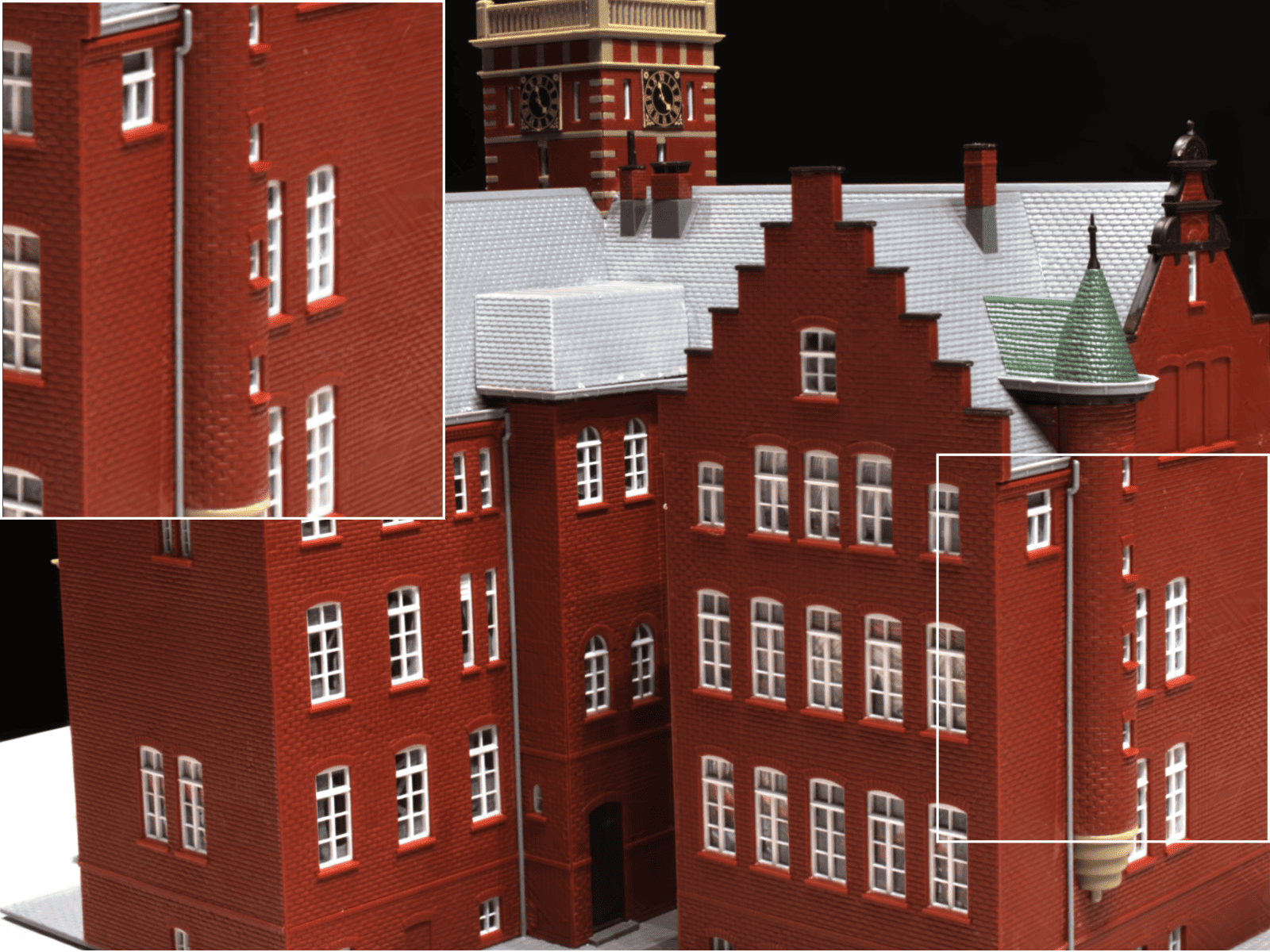} &
         \hspace{-4mm}
        \includegraphics[width=0.225\linewidth]{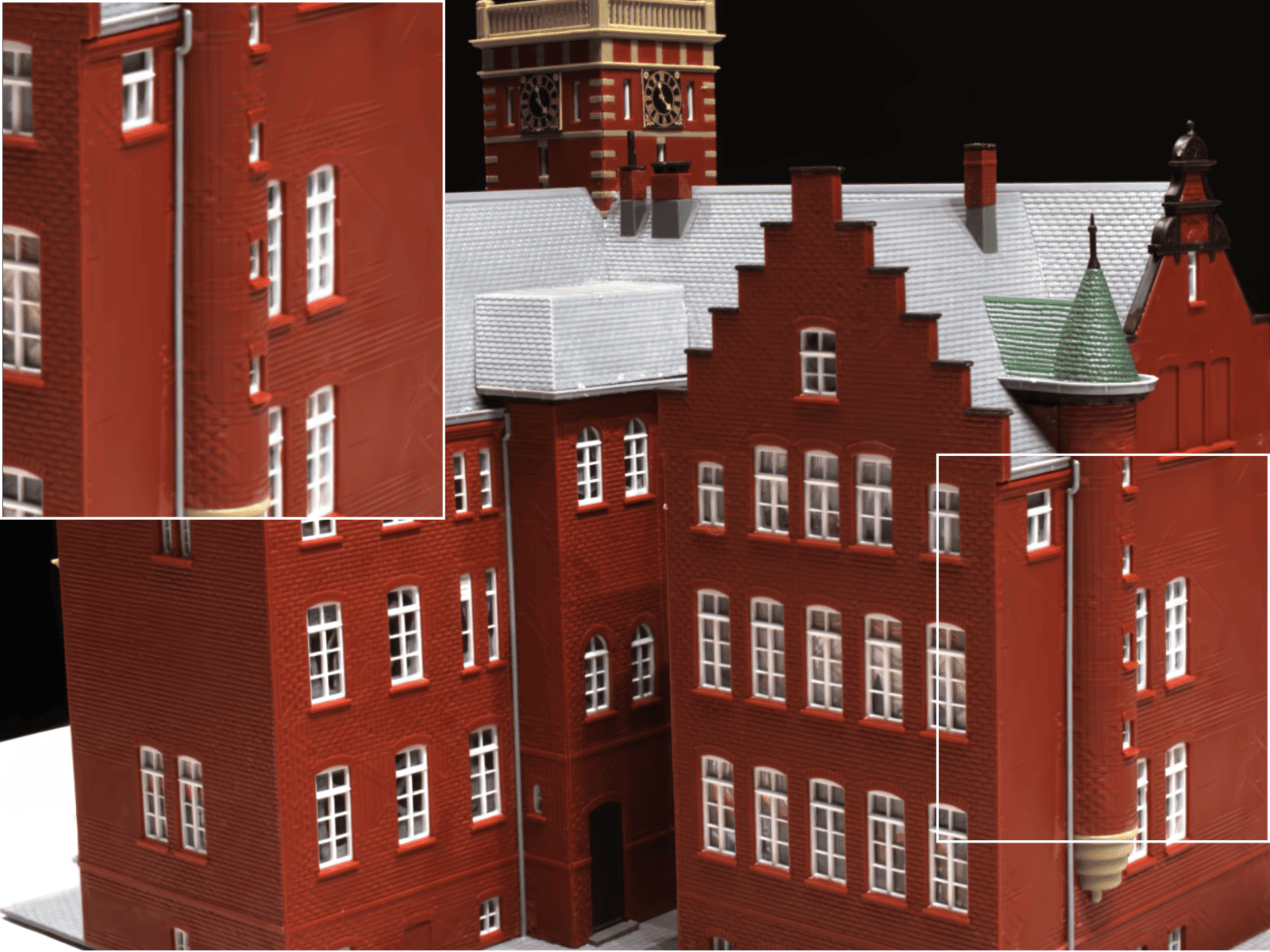} &
        \hspace{-4mm}
        \includegraphics[width=0.225\linewidth]{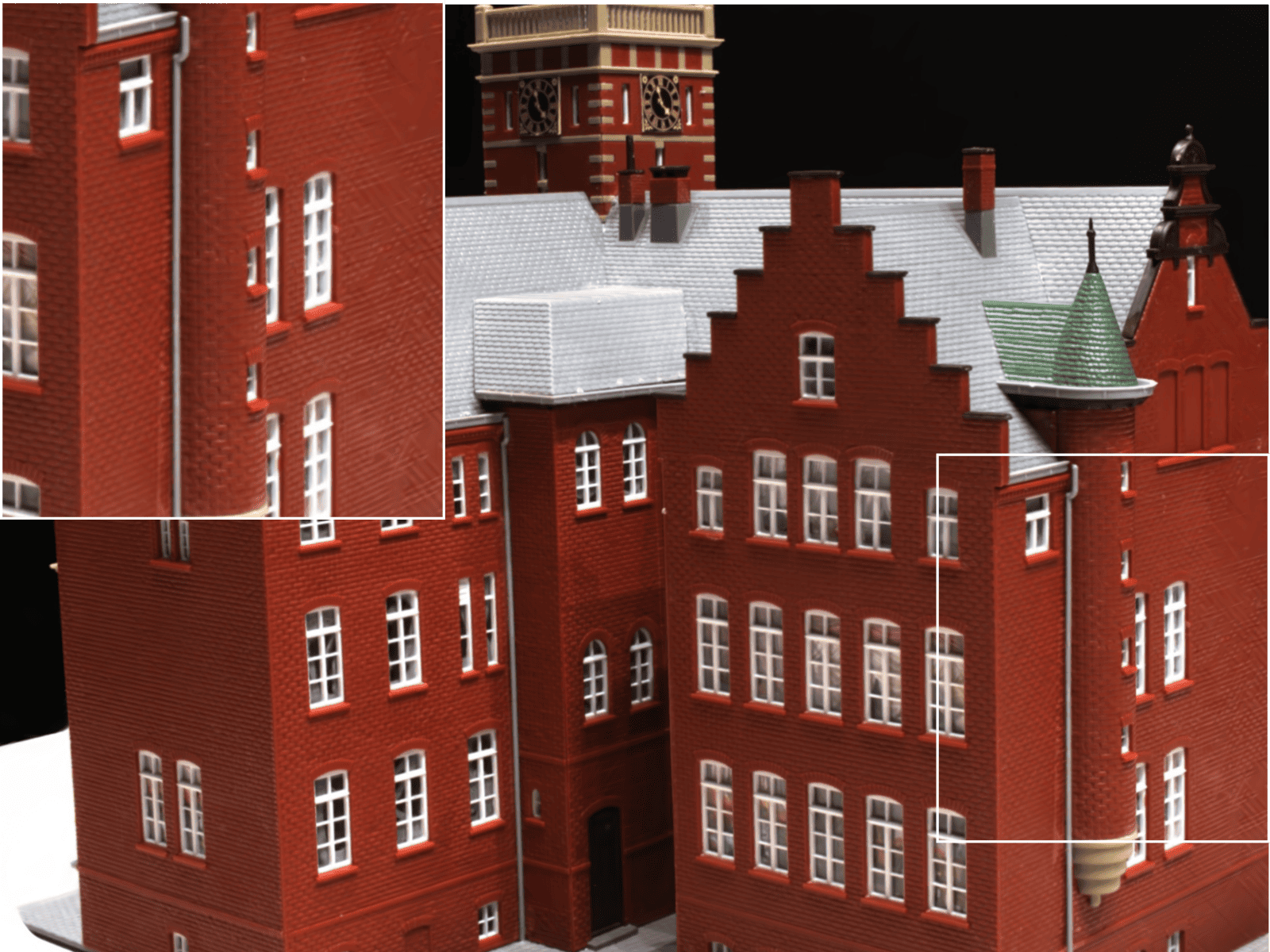} &
               \hspace{-4mm}
        \includegraphics[width=0.225\linewidth]{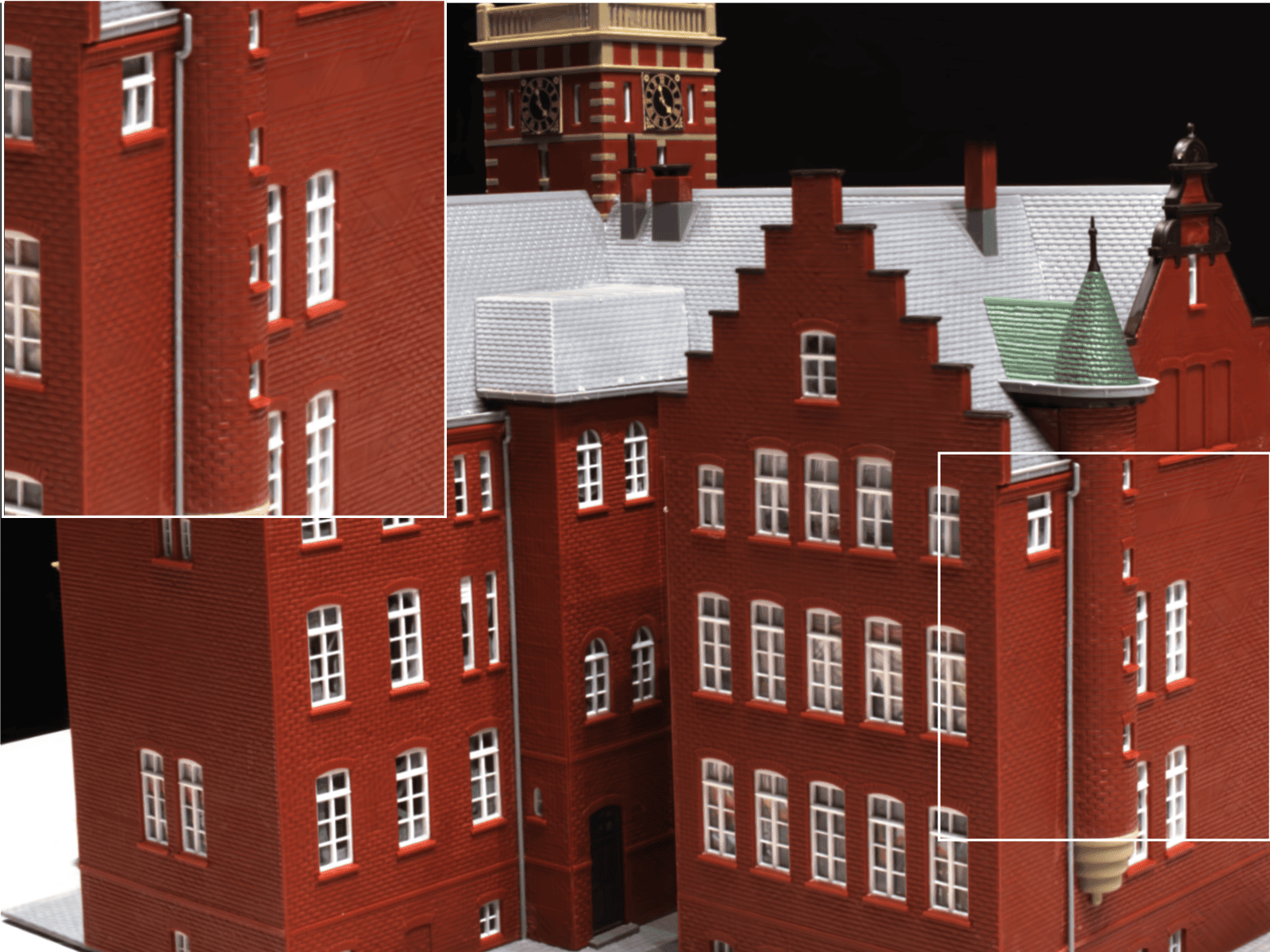} &
      \hspace{-4mm}
        \includegraphics[width=0.225\linewidth]{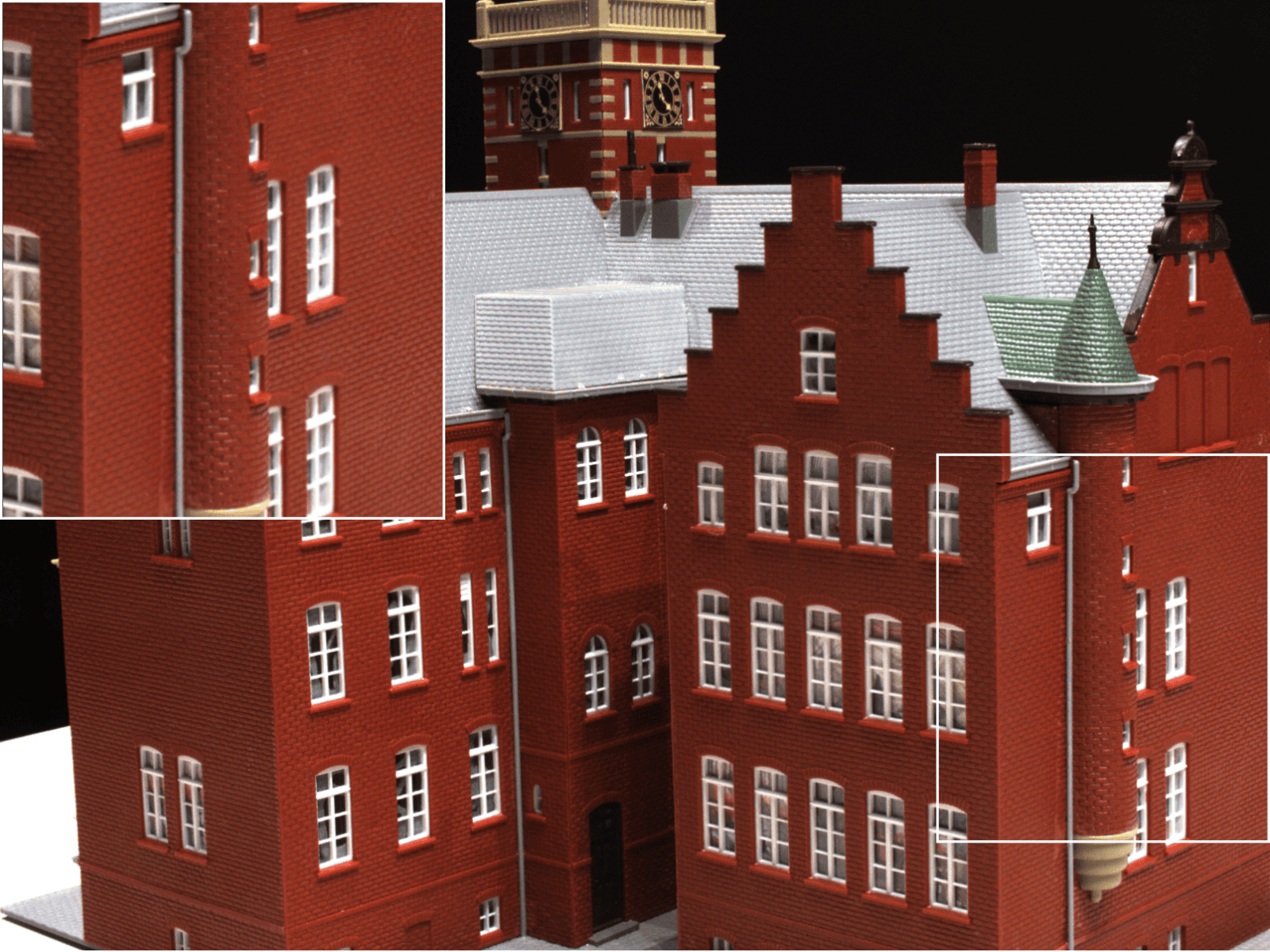}\vspace{-2mm}\\[2ex]
       
        \rotatebox{90}{\scriptsize\textbf{scan37}} &
         \vspace{-2mm}
        \includegraphics[width=0.225\linewidth]{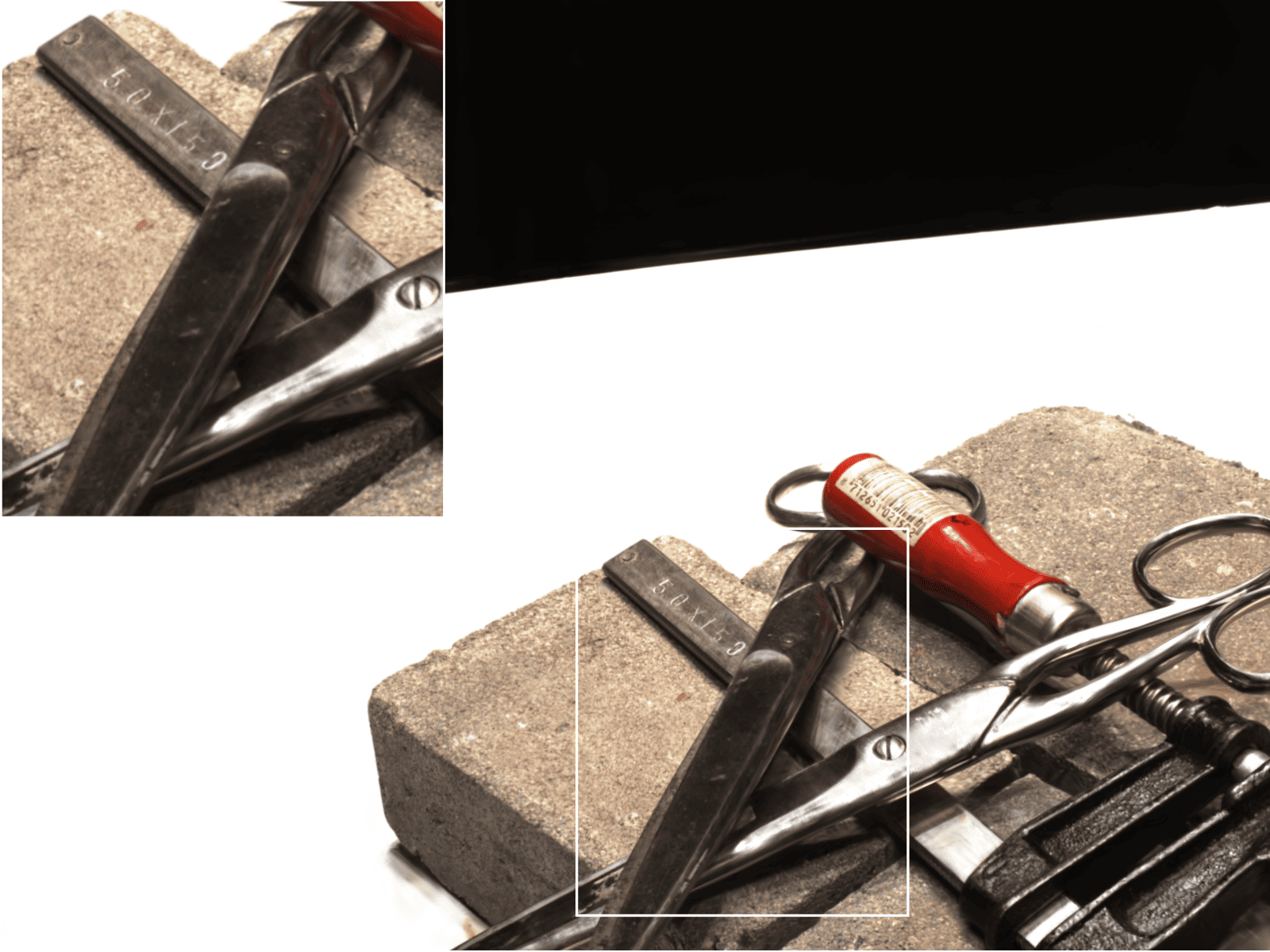} &
         \hspace{-4mm}
        \includegraphics[width=0.225\linewidth]{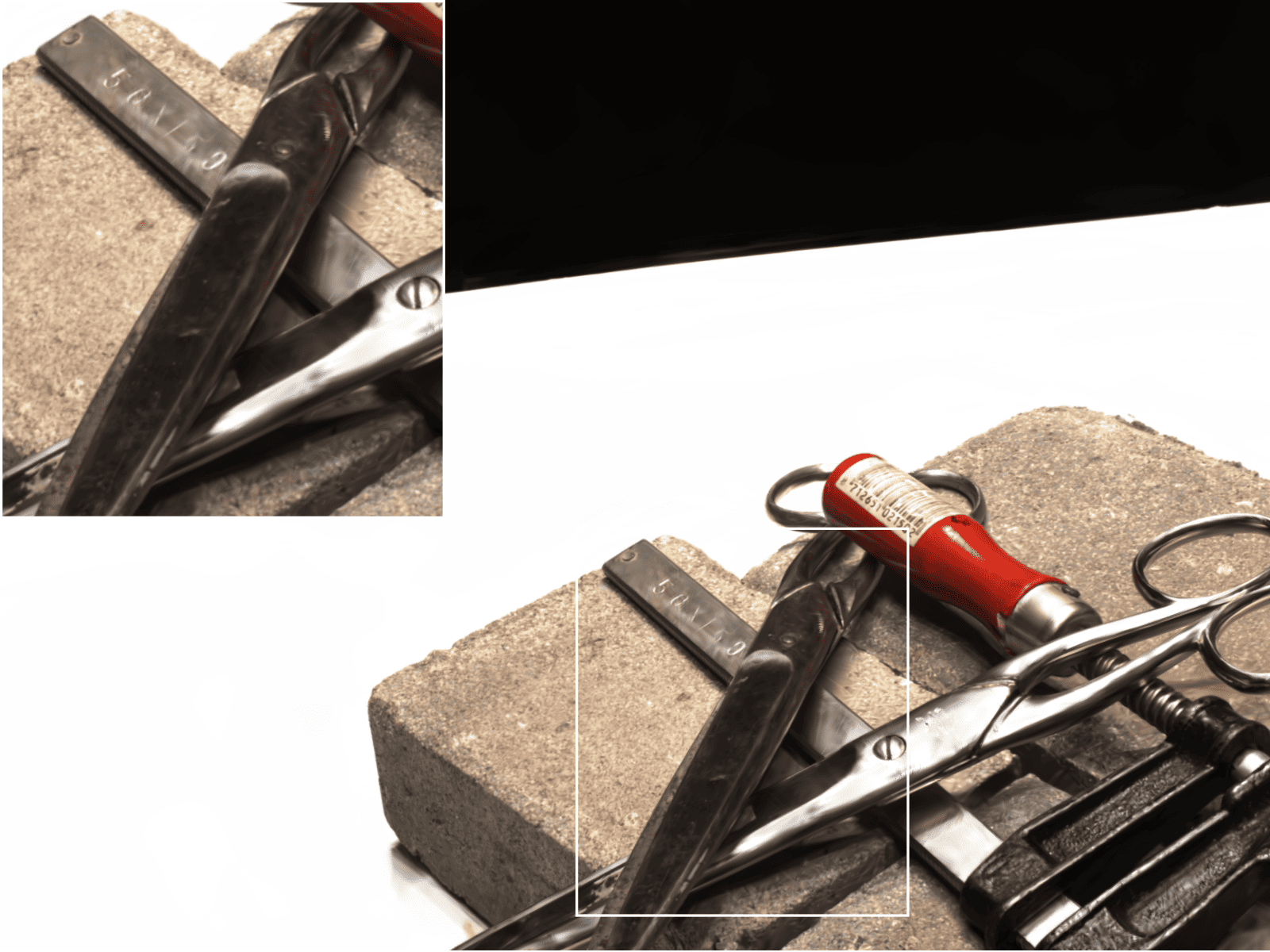} &
        \hspace{-4mm}
        \includegraphics[width=0.225\linewidth]{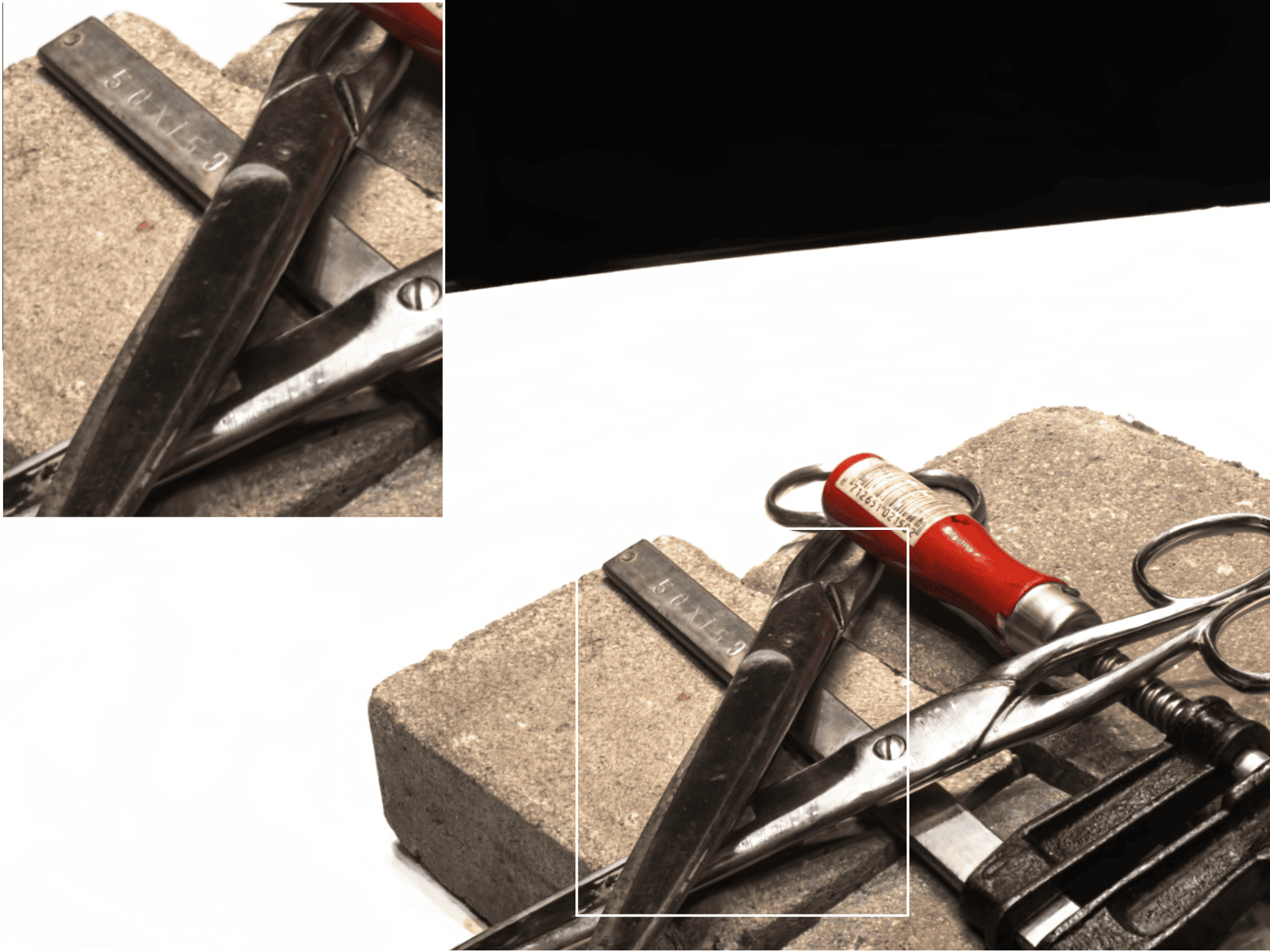} &
               \hspace{-4mm}
        \includegraphics[width=0.225\linewidth]{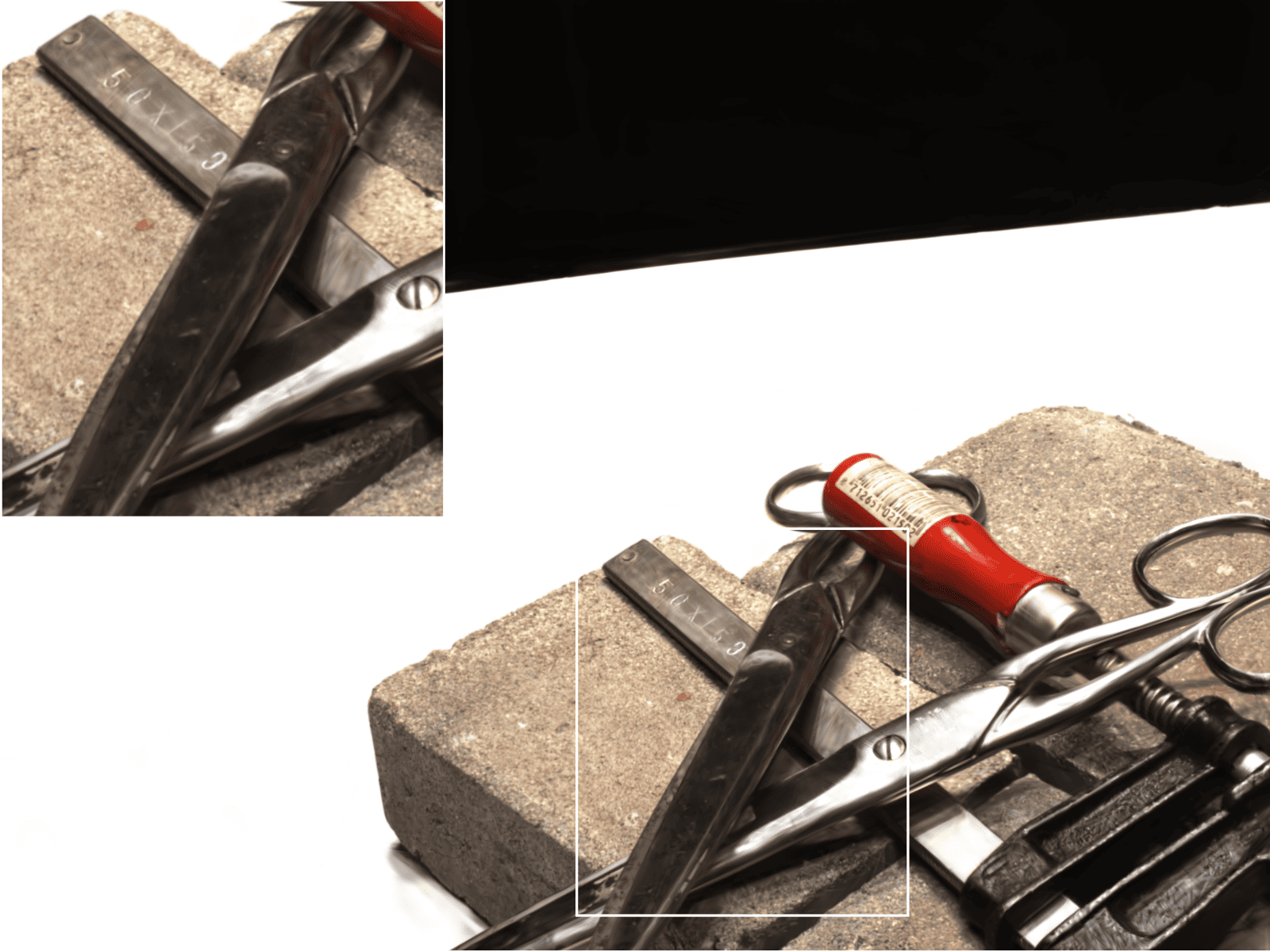} &
      \hspace{-4mm}
        \includegraphics[width=0.225\linewidth]{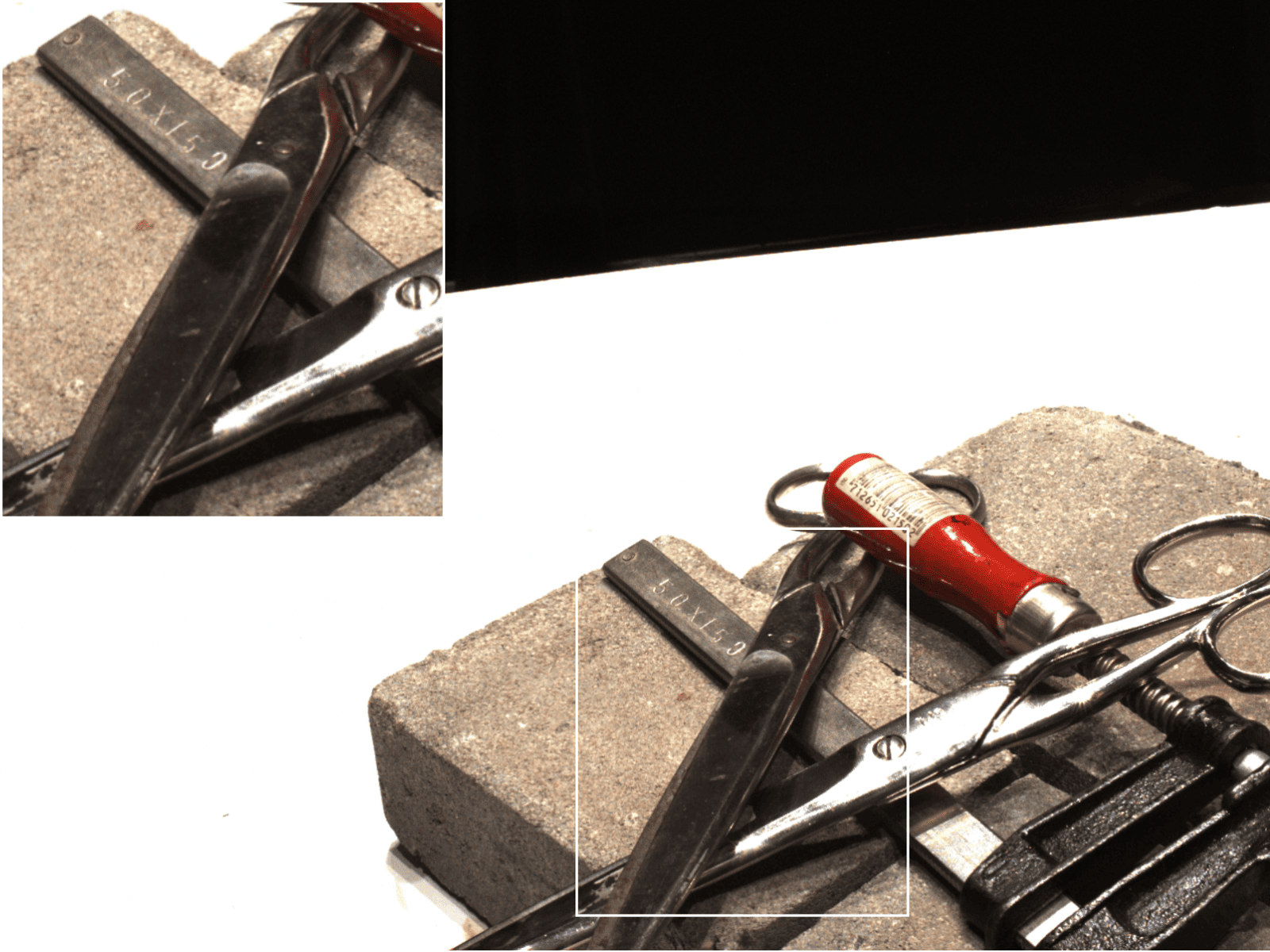}\vspace{-2mm}\\[2ex]
    
        \rotatebox{90}{\scriptsize\textbf{scan40}} &
         \vspace{-2mm}
        \includegraphics[width=0.225\linewidth]{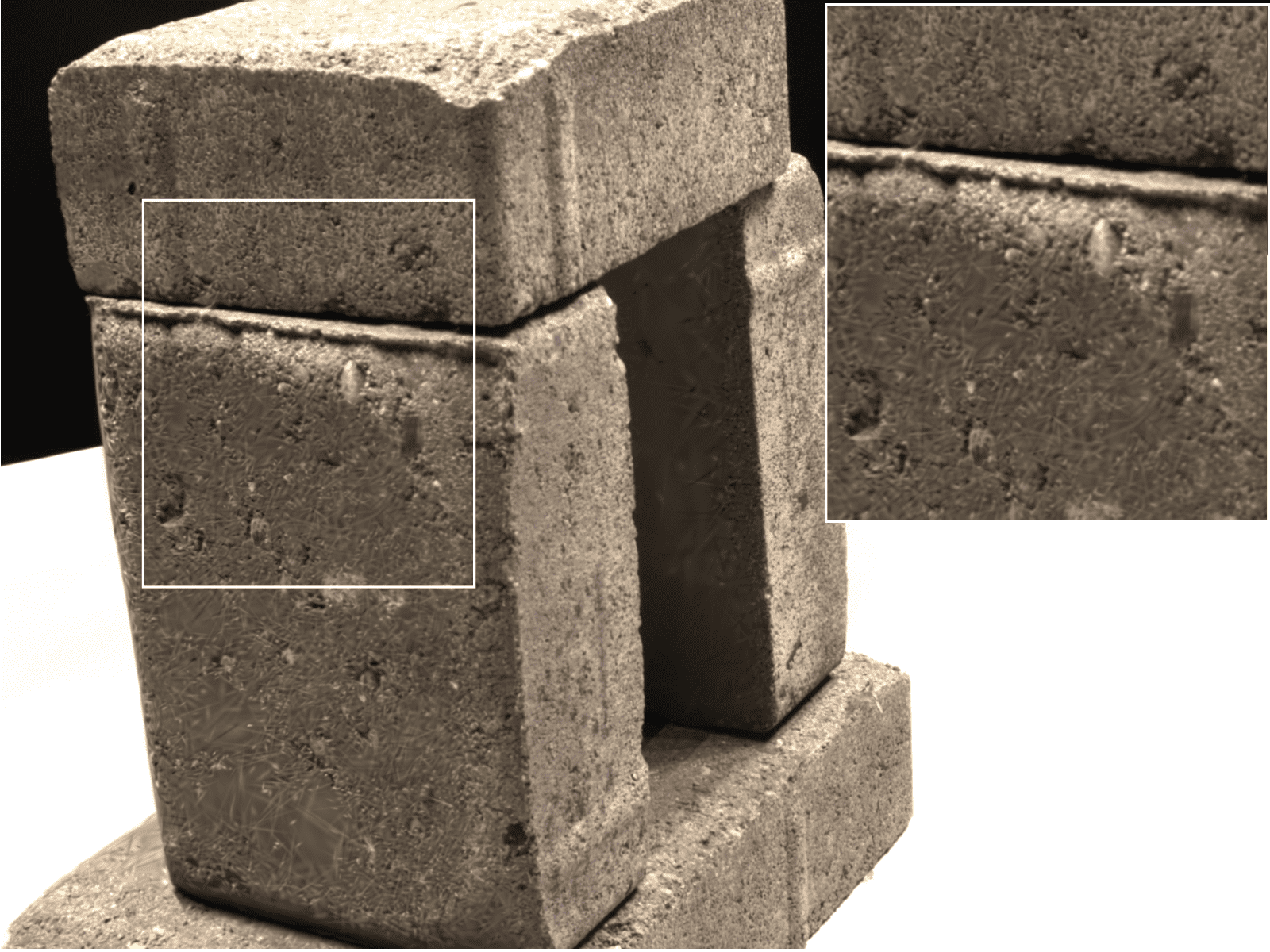} &
         \hspace{-4mm}
        \includegraphics[width=0.225\linewidth]{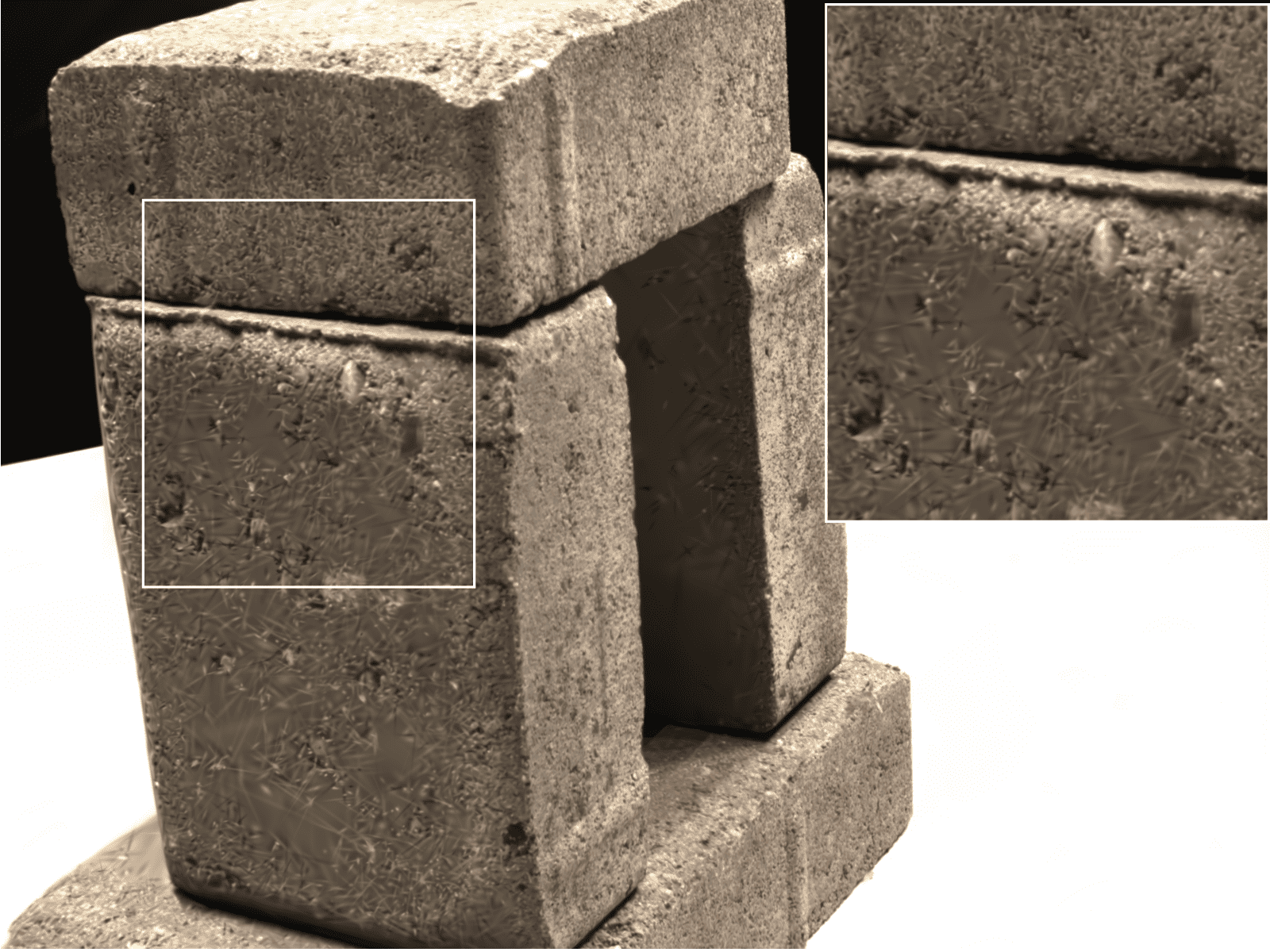} &
        \hspace{-4mm}
        \includegraphics[width=0.225\linewidth]{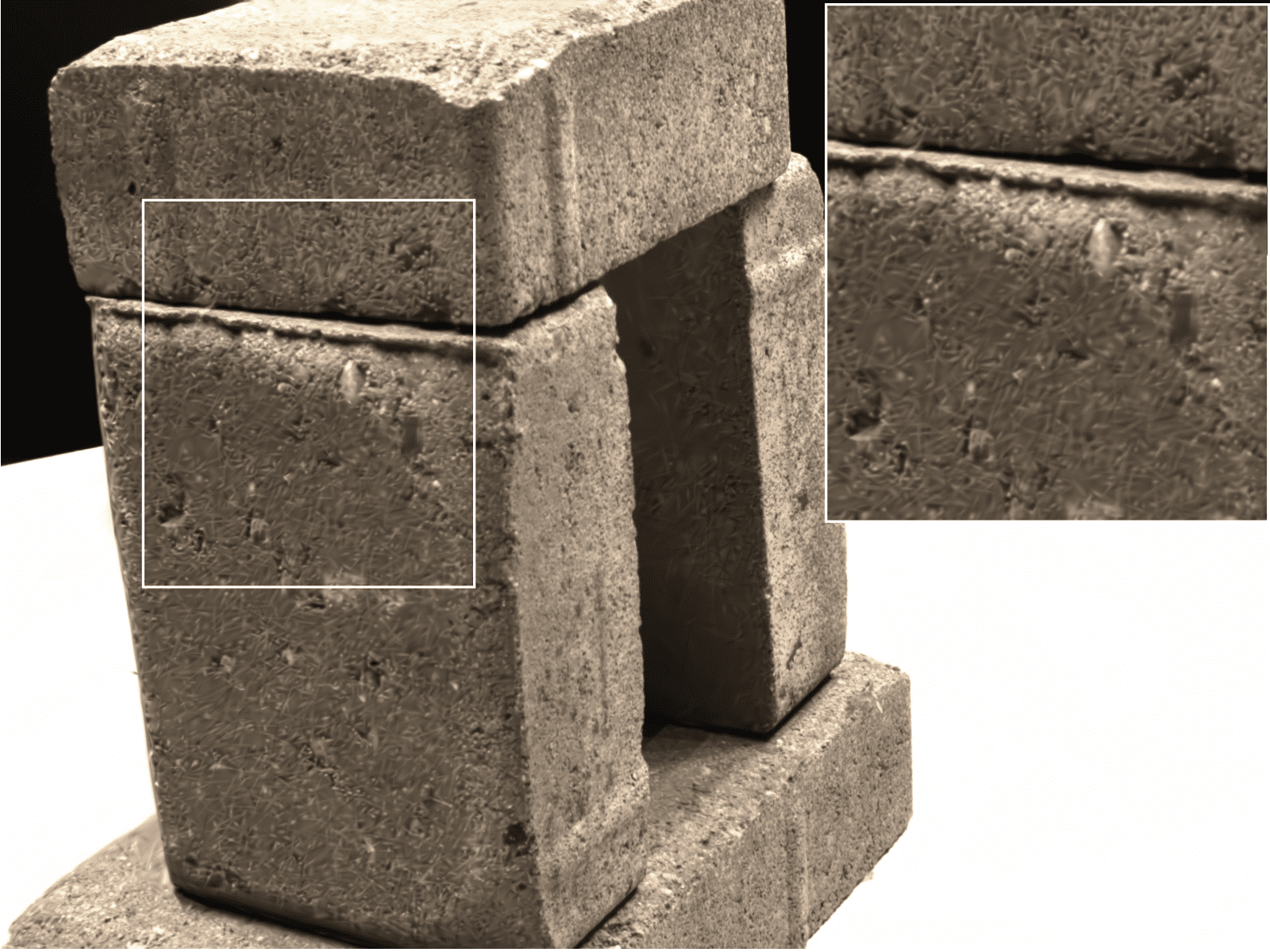} &
               \hspace{-4mm}
        \includegraphics[width=0.225\linewidth]{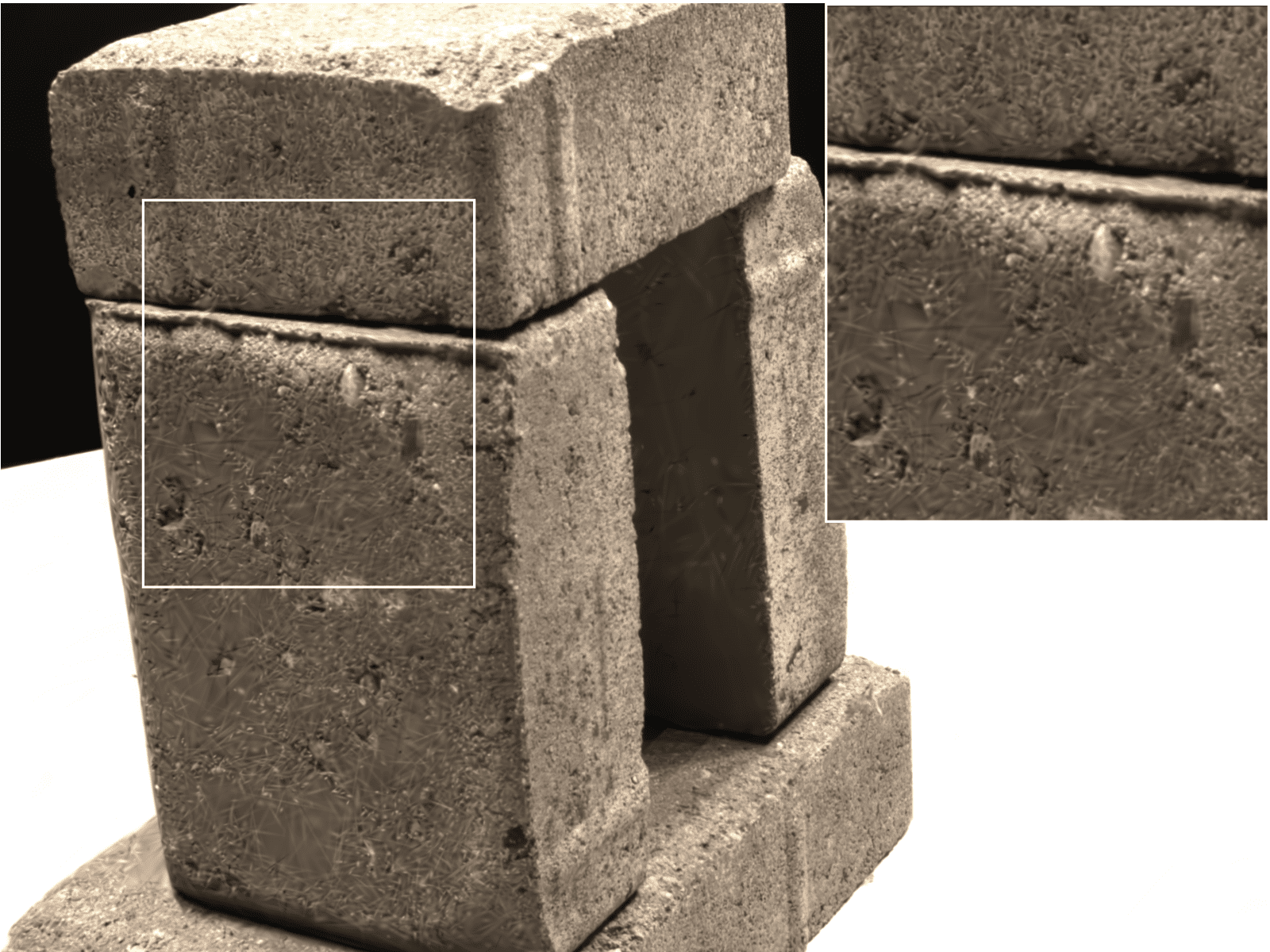} &
      \hspace{-4mm}
        \includegraphics[width=0.225\linewidth]{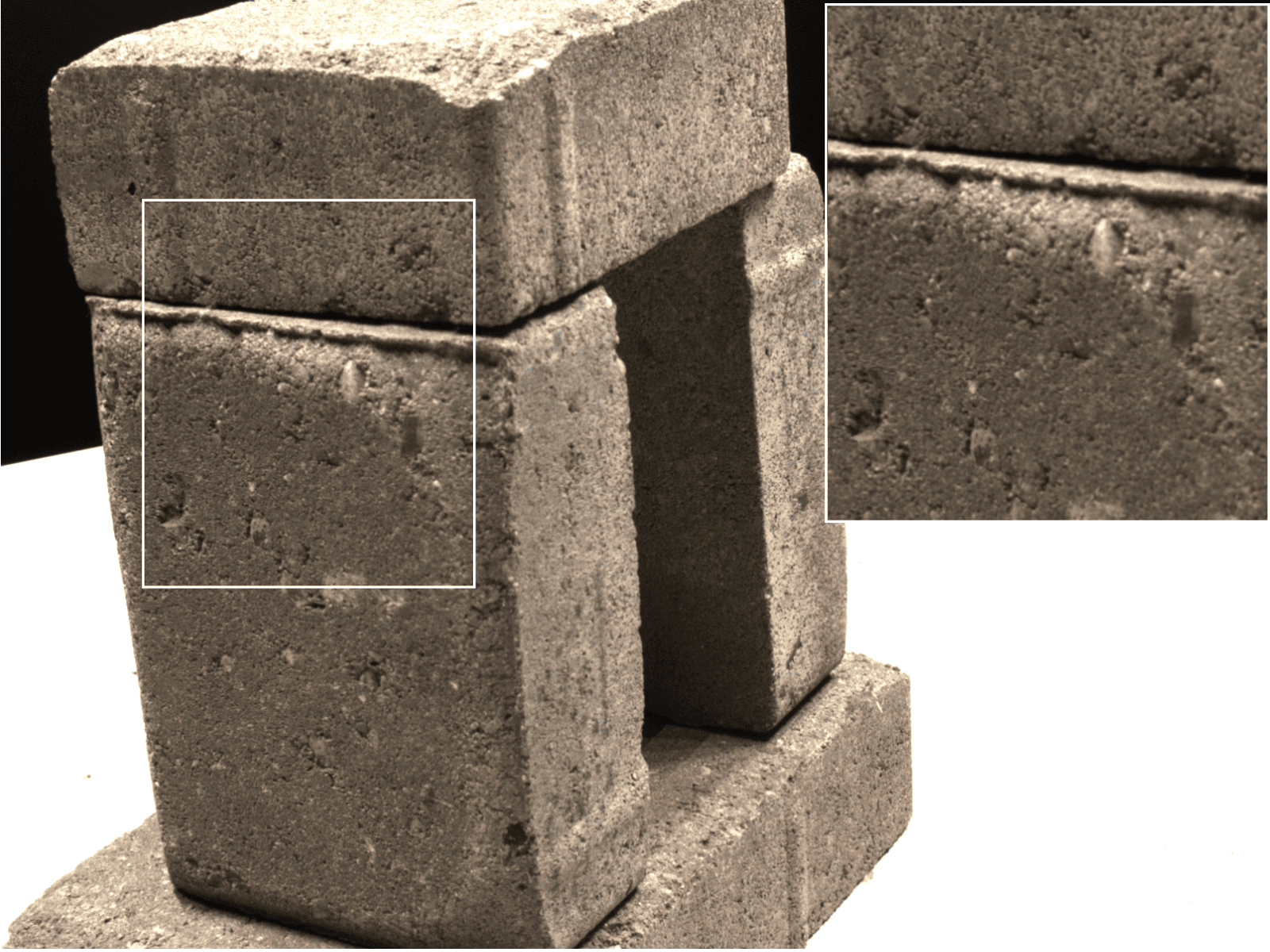}\vspace{-2mm}\\[2ex]

         \rotatebox{90}{\scriptsize\textbf{scan114}} &
                  \vspace{-2mm}
        \includegraphics[width=0.225\linewidth]{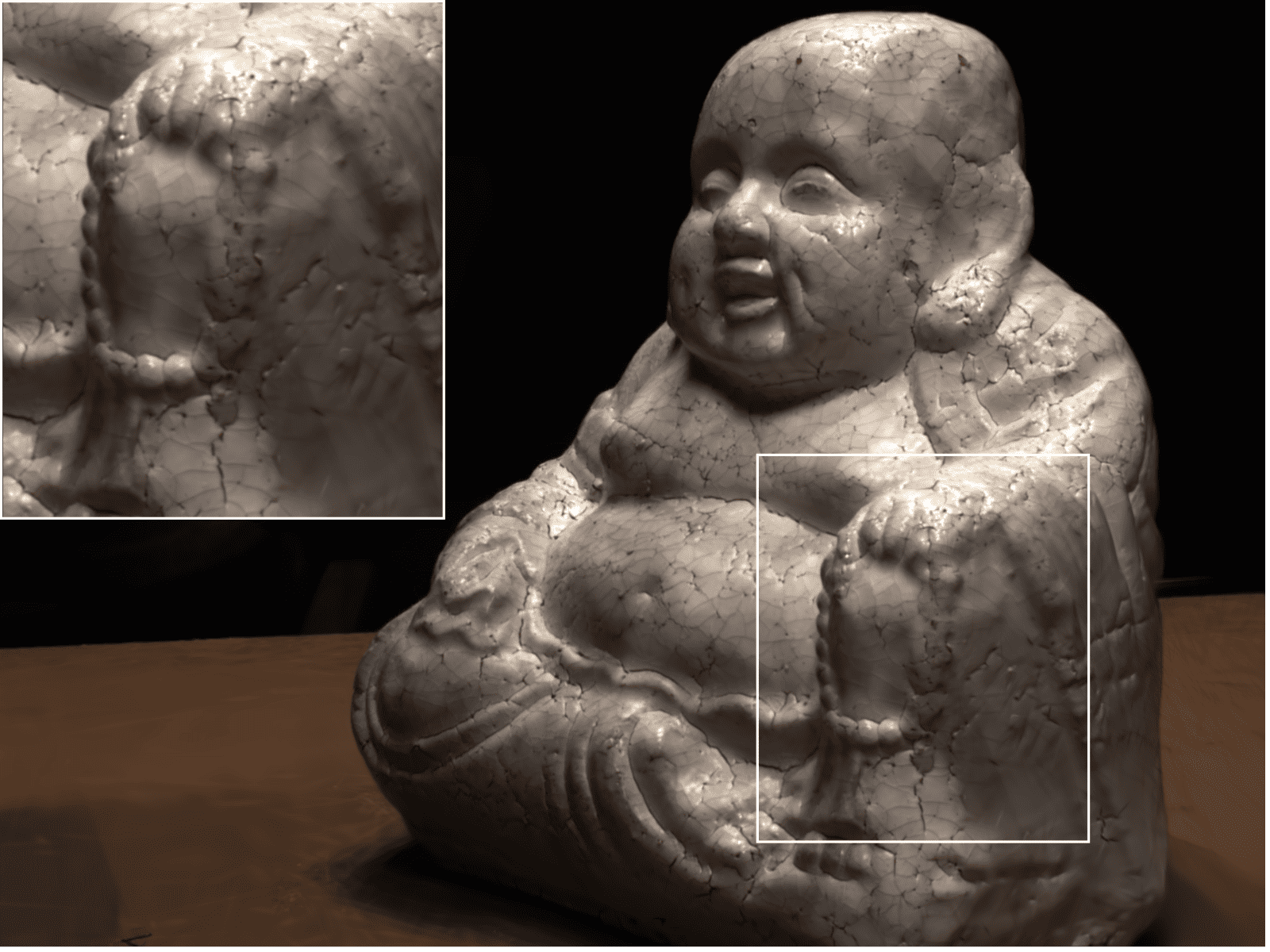} &
         \hspace{-4mm}
        \includegraphics[width=0.225\linewidth]{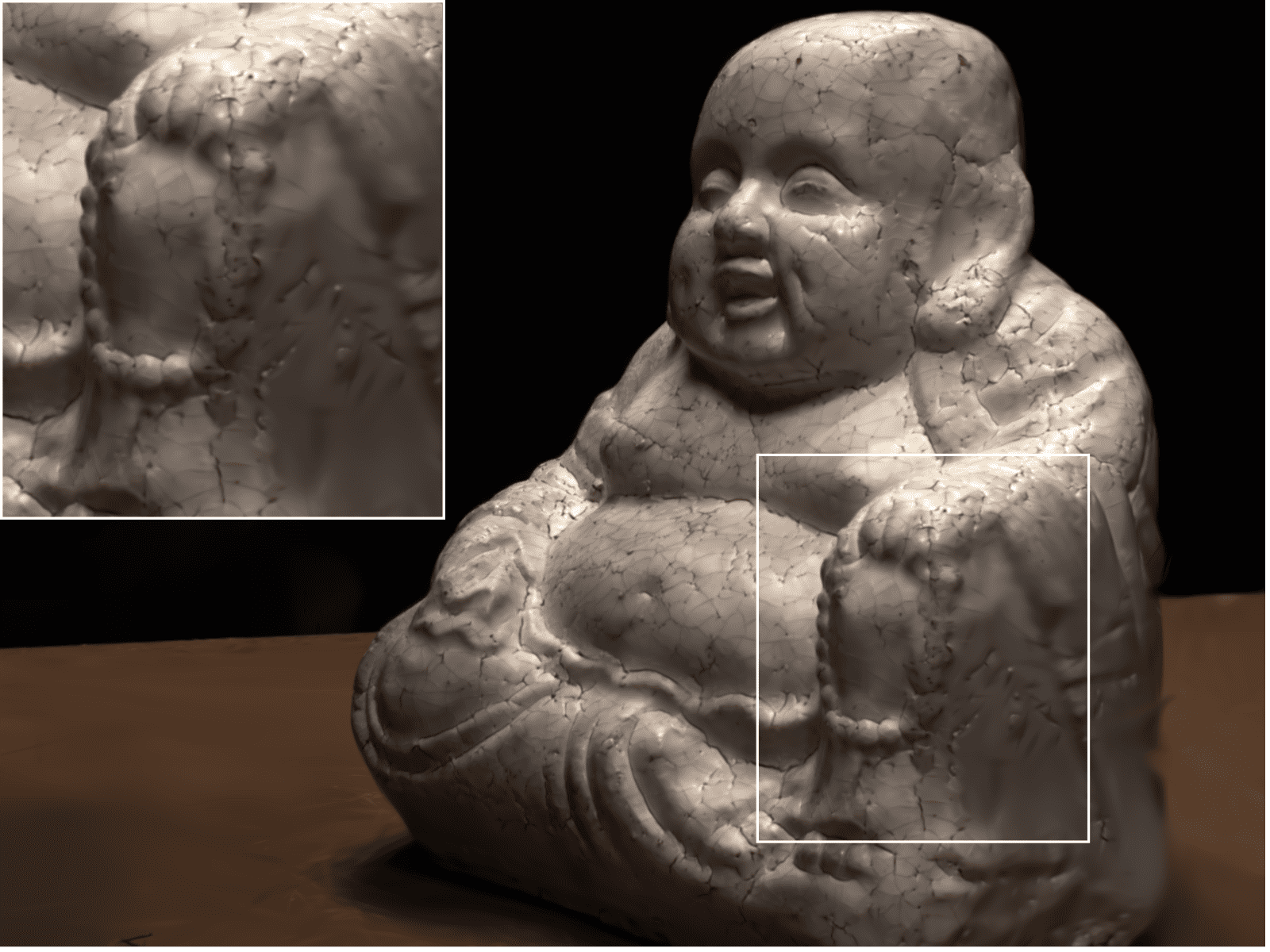} &
        \hspace{-4mm}
        \includegraphics[width=0.225\linewidth]{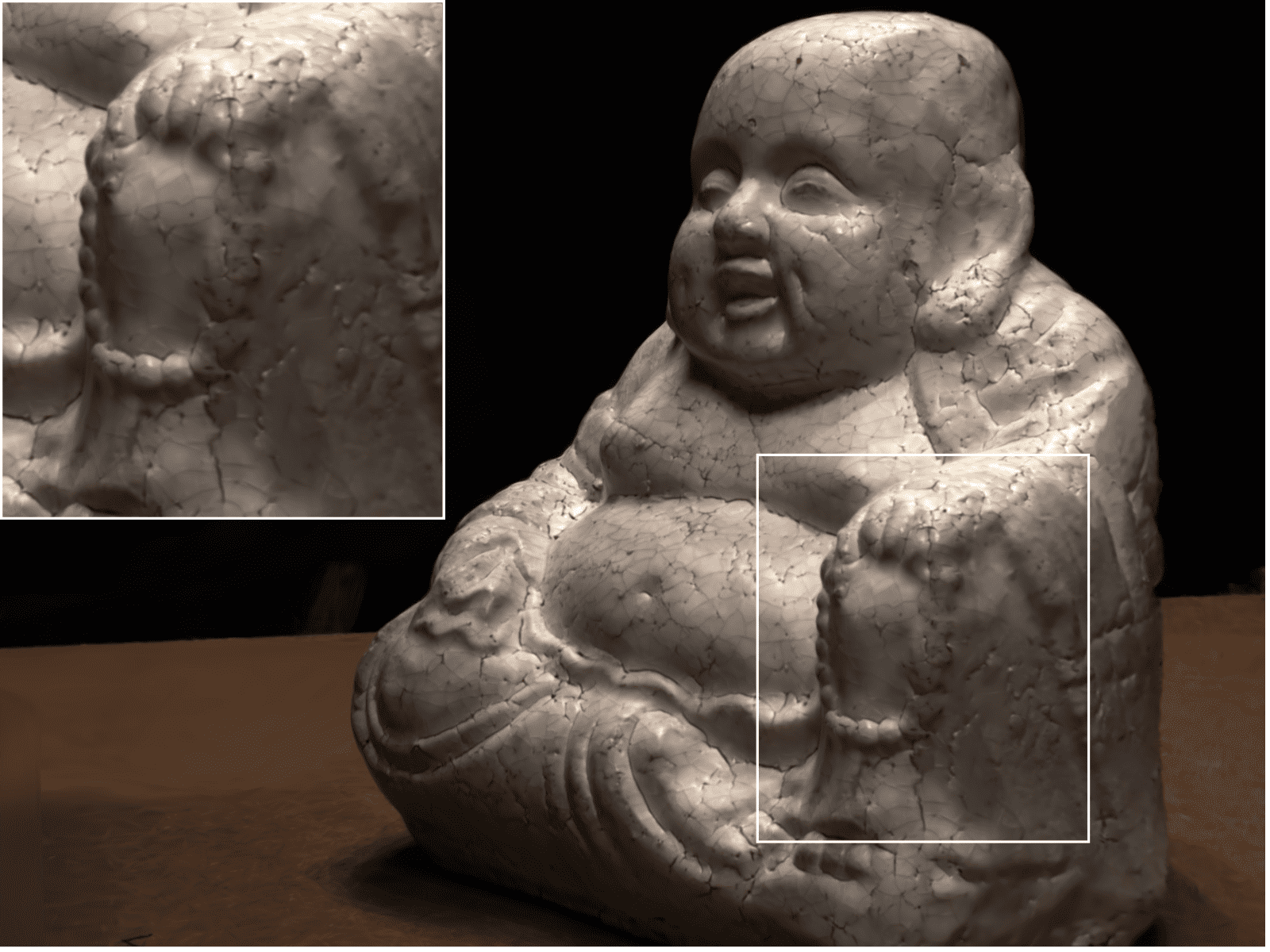} &
               \hspace{-4mm}
        \includegraphics[width=0.225\linewidth]{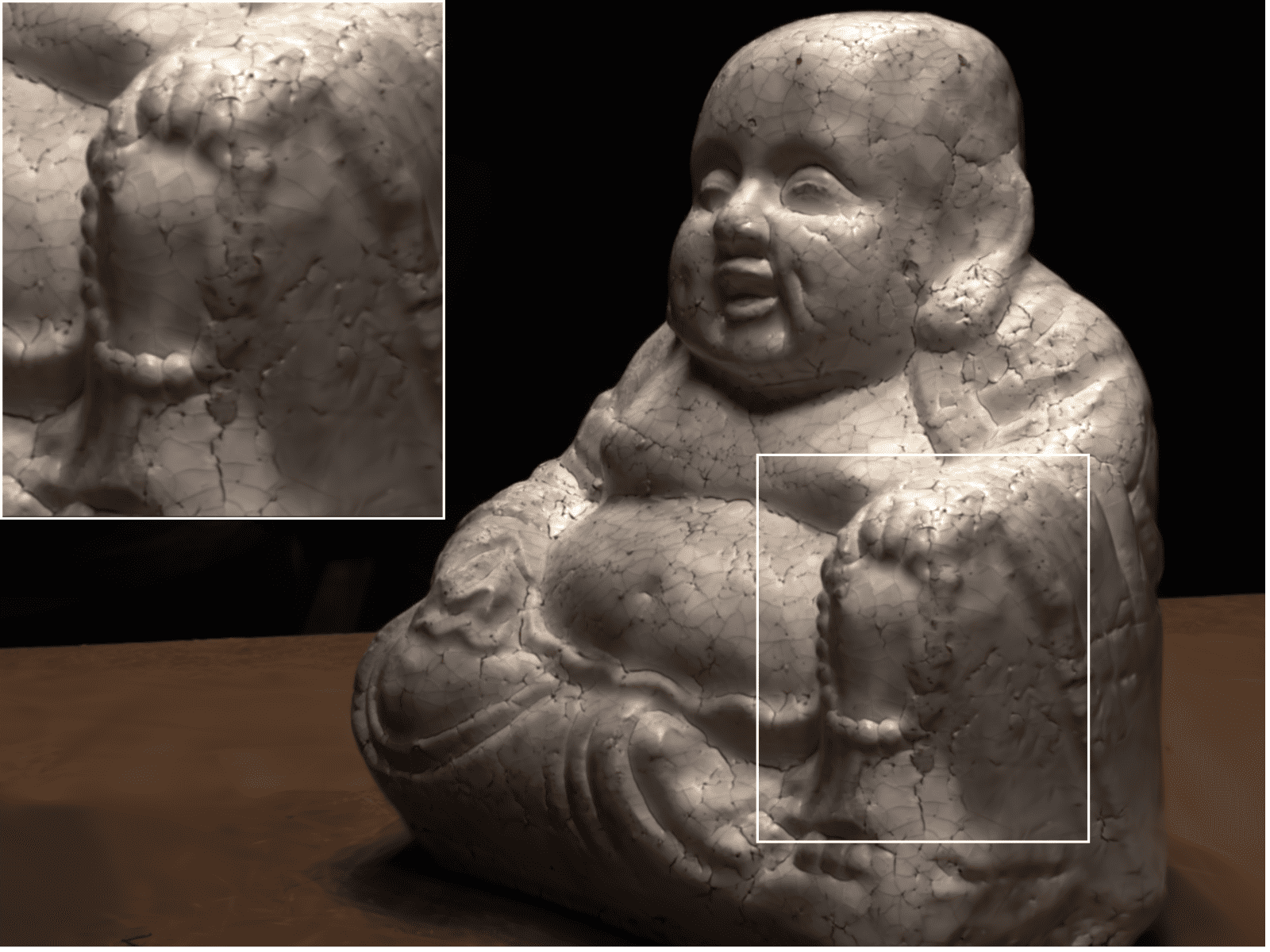} &
      \hspace{-4mm}
        \includegraphics[width=0.225\linewidth]{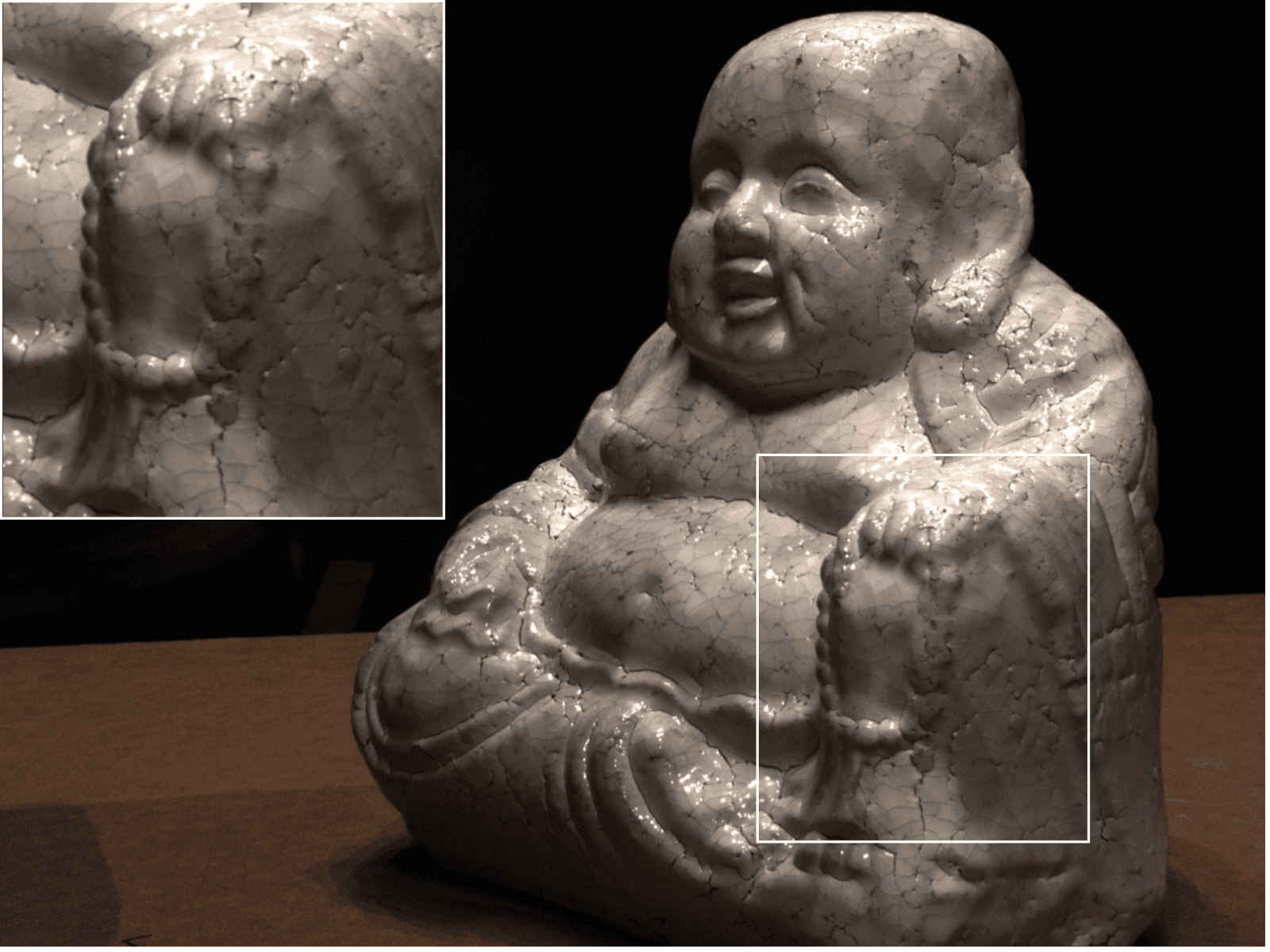}\vspace{-2mm}\\[2ex]
       
        \rotatebox{90}{\scriptsize\textbf{scan122}} &
         \vspace{-2mm}
        \includegraphics[width=0.225\linewidth]{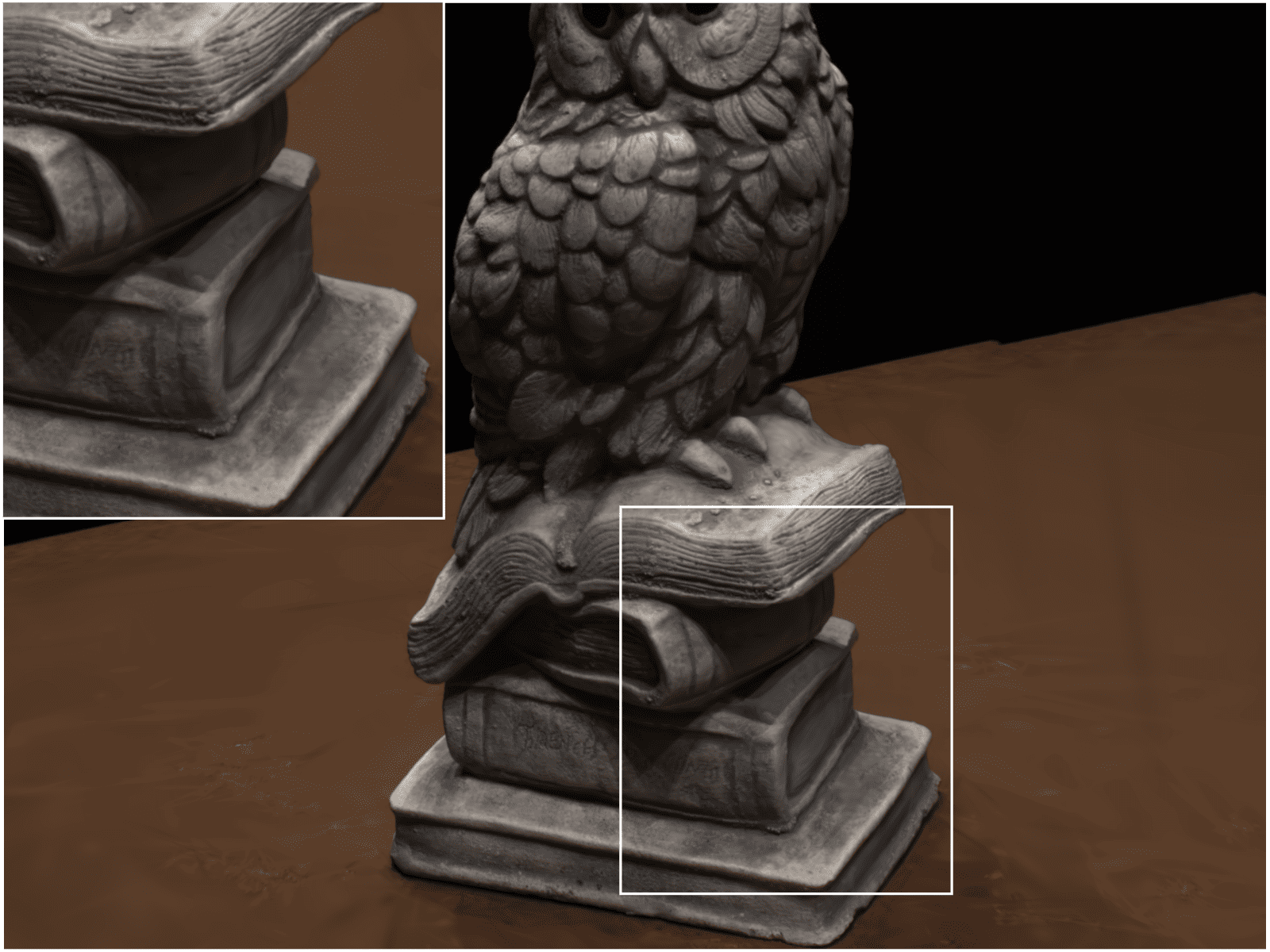} &
         \hspace{-4mm}
        \includegraphics[width=0.225\linewidth]{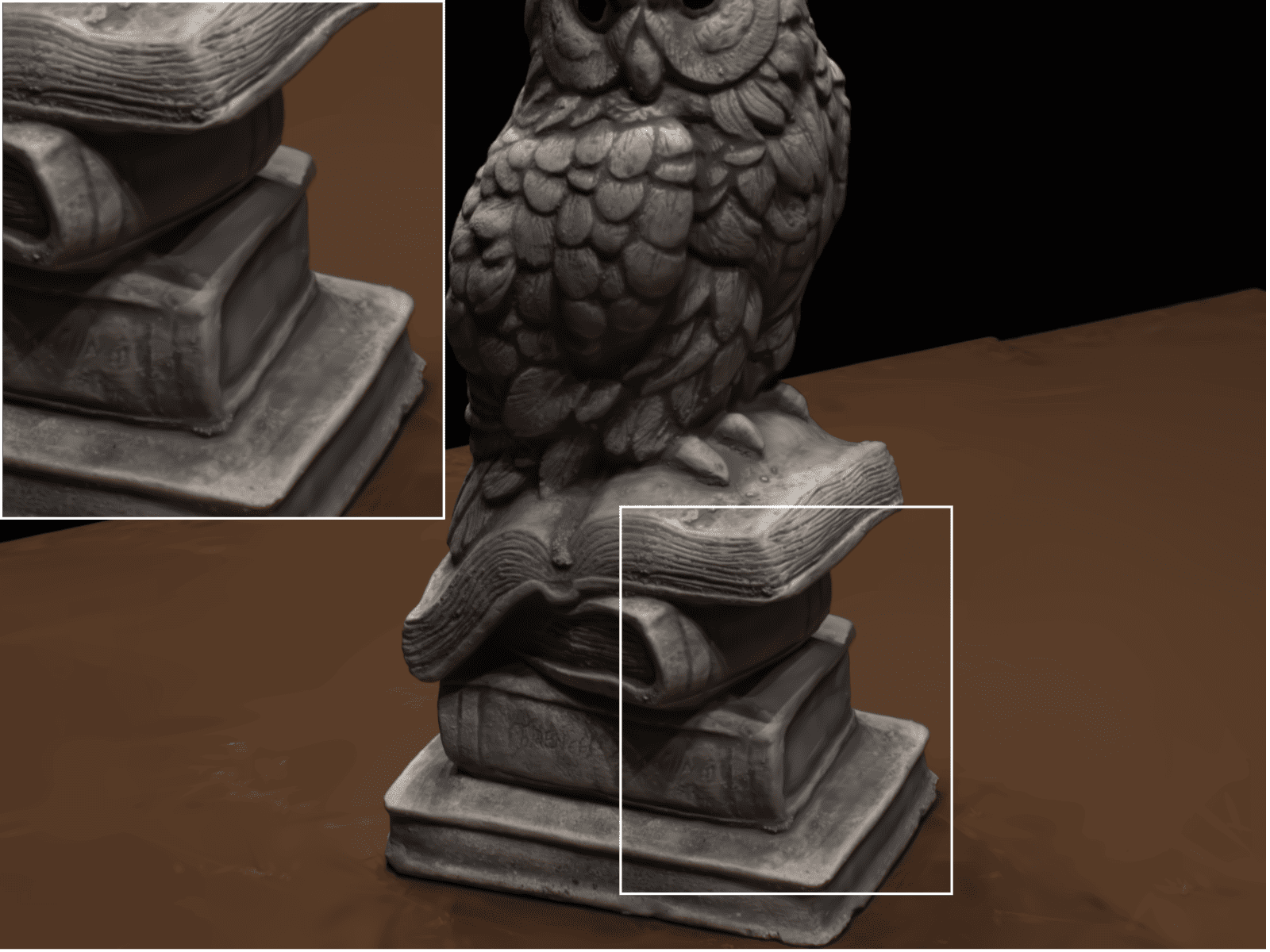} &
        \hspace{-4mm}
        \includegraphics[width=0.225\linewidth]{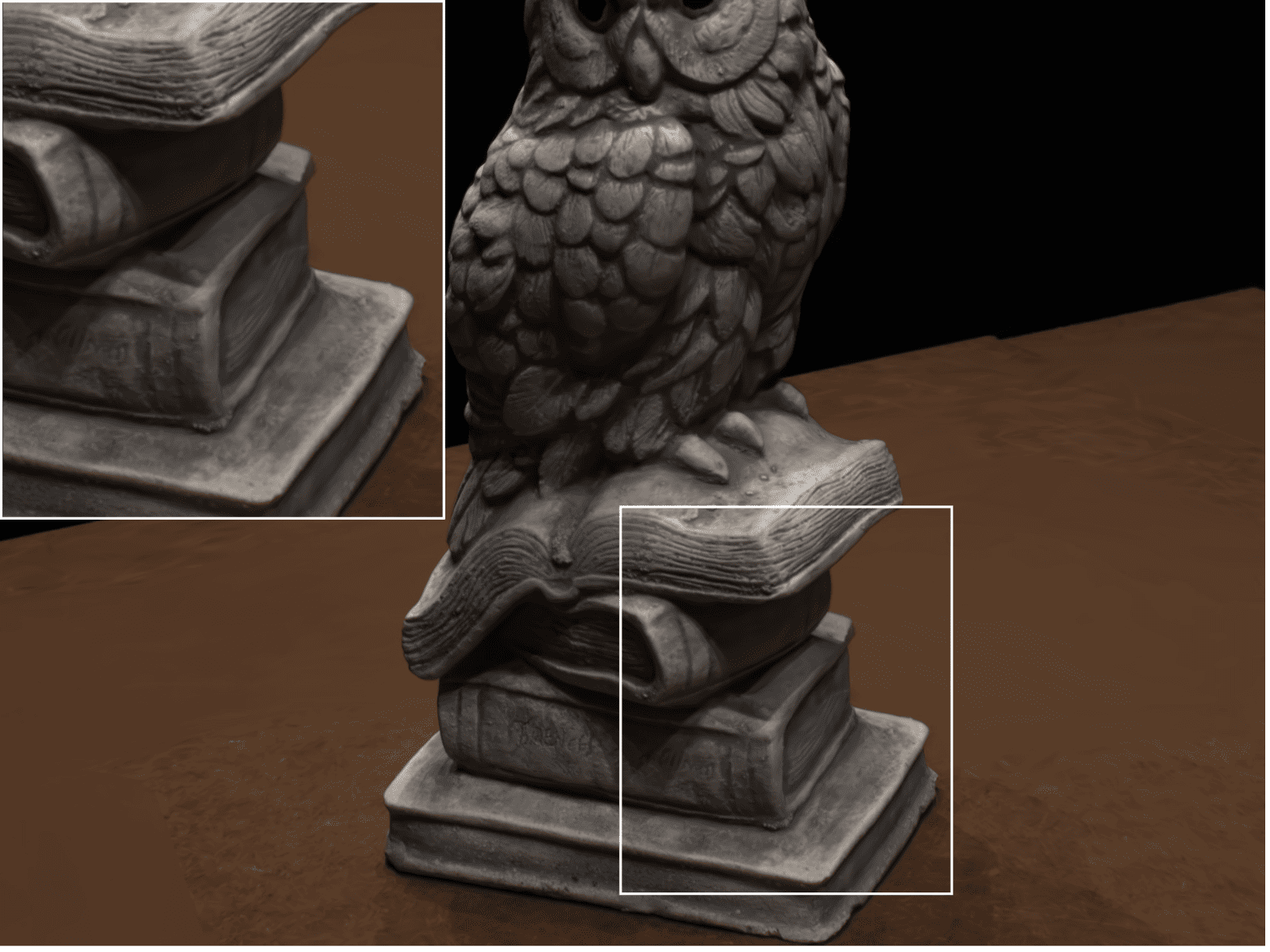} &
               \hspace{-4mm}

         \includegraphics[width=0.225\linewidth]{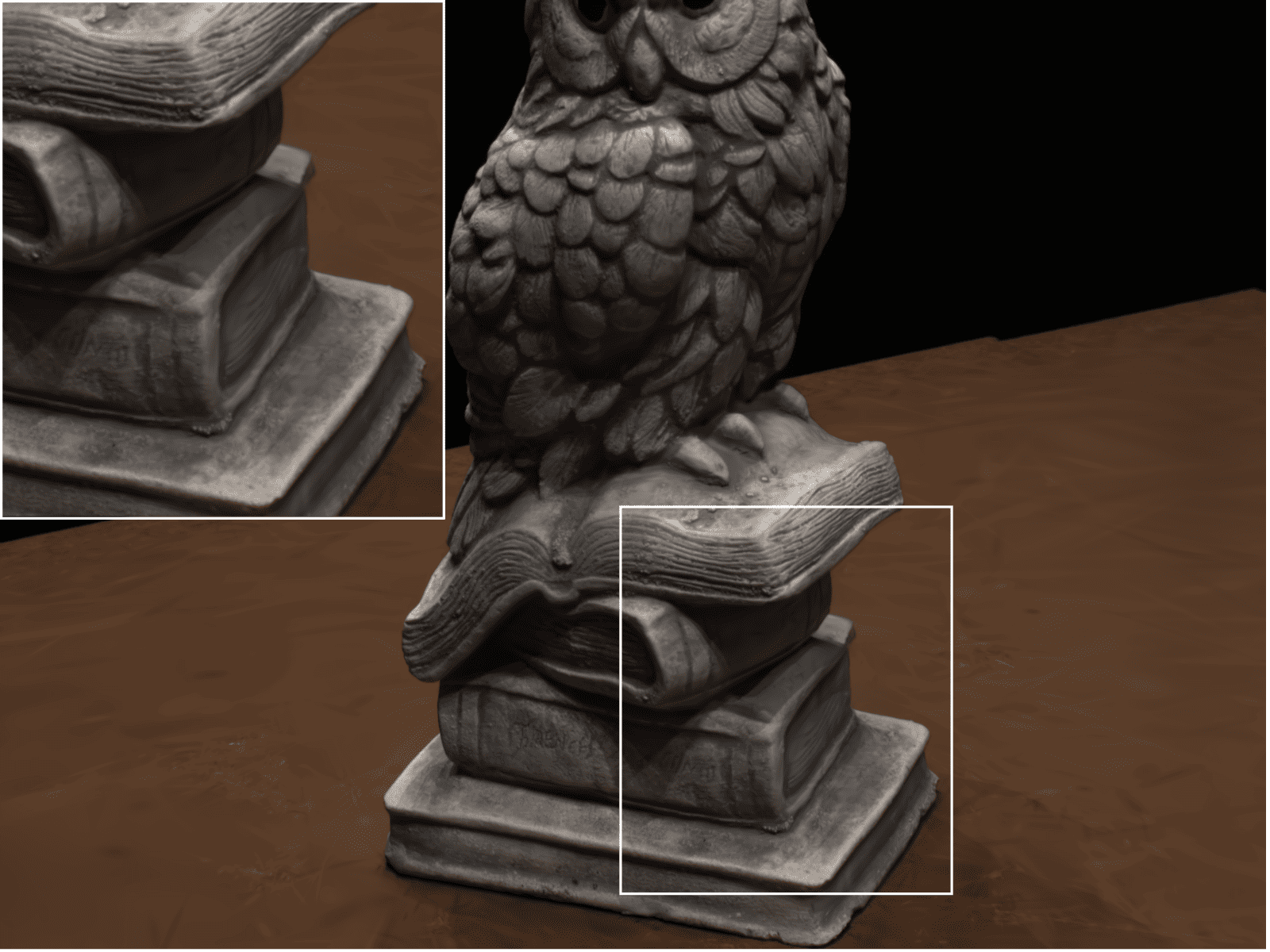} &
      \hspace{-4mm}
        \includegraphics[width=0.225\linewidth]{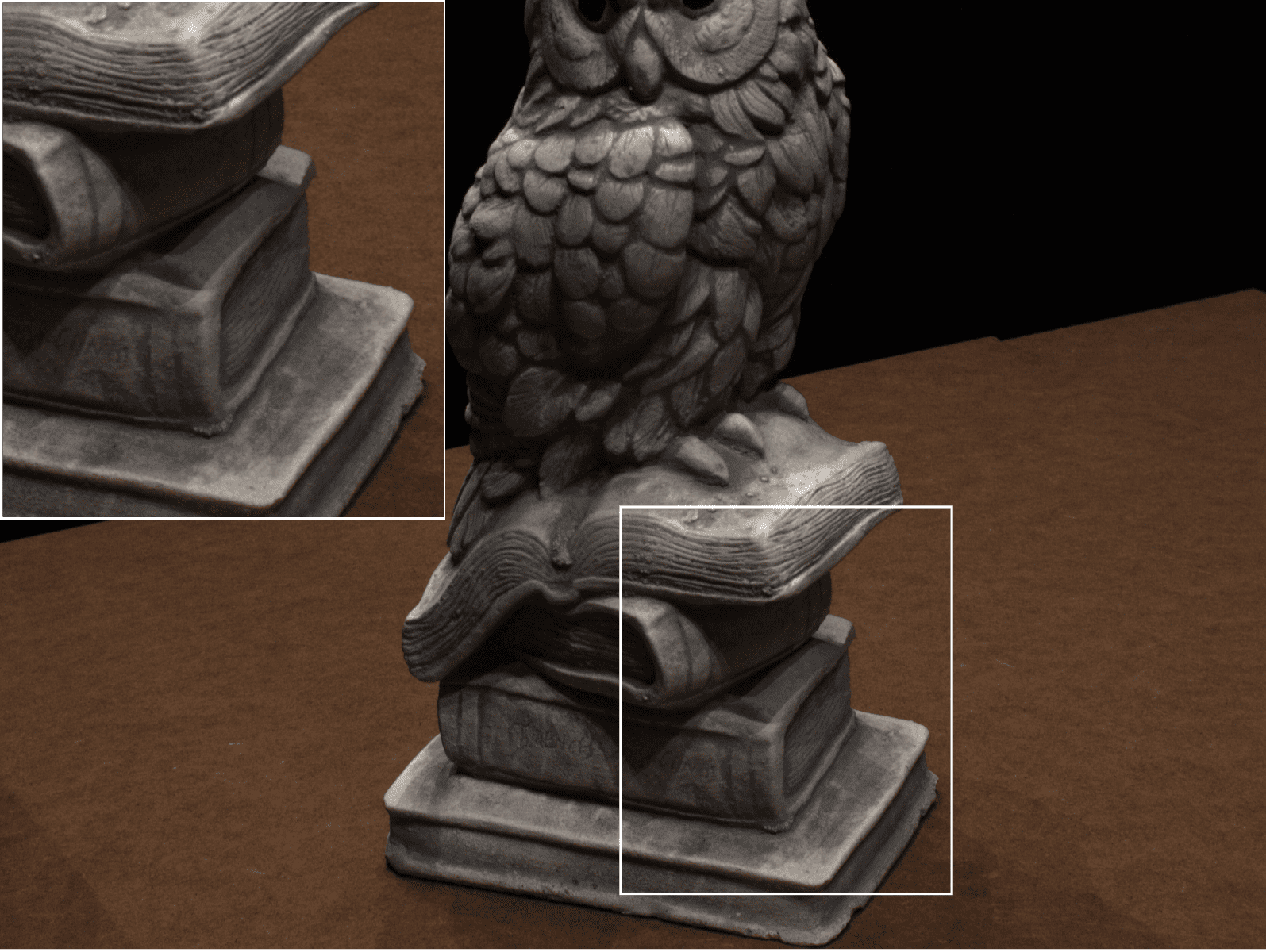}\vspace{-2mm}\\[2ex]

    \end{tabular}
\vspace{-2mm}
    \caption{\textbf{Rendering quality} comparison of 3DGS, 2DGS, PGSR and EntON on the small-scale DTU dataset, as well as ground truth (GT) images in original resolution.}
     \label{fig:Qualitative_rendering}
\end{figure*}

The rendering quality (Figure \ref{fig:Qualitative_rendering}) shown by the rendered test images also underlines the overall strong performance of EntON compared to 3DGS, 2DGS and PGSR. 
EntON is able to reconstruct fine details more accurately, avoiding the over-reconstructed regions that are often observed in 2DGS and, to some extent, in 3DGS and PGSR. This advantage is particularly pronounced on textured surfaces, as seen in scenes scan24, scan40, scan114, scan122. Some blurring and reduced sharpness are observed for EntON on reflective or homogeneous surfaces, such as in scene scan37. This is due to fewer Gaussians in these regions, rather no increased splitting or even pruning due to high Eigenentropy. Notably, EntON achieves high rendering quality despite using a relatively small number of Gaussians. While 3DGS also delivers good quality with few over-reconstructed areas, it relies on a substantially higher number of Gaussians. In contrast, 2DGS and PGSR use fewer Gaussians but often produce poorer or partially blurred reconstructions.

\subsection{Large-Scale Data}\label{sec:Results_large}

The following sections show qualitative (Section \ref{sec:results_2_quanti}) and quantitative (Section \ref{sec:results_2_quali}) results on the TUM2TWIN large-scale dataset of EntON in comparison to 3DGS, 2DGS and PGSR.

\subsubsection{Quantitative Results}\label{sec:results_2_quanti}
The results across the key metrics, geometric accuracy, rendering quality, number of Gaussians, as well as training time, are presented in Tables \ref{tab:tnt_surface} - \ref{tab:tnt_time}.

\paragraph{Geometric Accuracy}

We evaluate the geometric accuracy of the reconstructed surface points using Chamfer cloud-to-cloud (C2C) distance measured against the reference point cloud, considering only points within 0.5\,m of the reference. 
On building1, EntON consistently produces highly accurate surfaces across all neighborhood sizes, with the best performance achieved for $k_{\text{nn}}=75$ (0.179\,m) and an average over all neighborhood sizes of 0.183\,m. Compared to the baselines, EntON outperforms 3DGS (0.184\,m) and 2DGS (0.193\,m), while slightly underperforming PGSR (0.197\,m). Overall, EntON improves geometric accuracy by up to 2.7\% compared to 3DGS (best variant) and 7.3\% compared to 2DGS, while slightly trailing PGSR by 9.1\% for the best variant on building1.

\definecolor{yellow}{RGB}{255, 255, 204}
\definecolor{orange}{RGB}{255, 204, 153}
\definecolor{red}{RGB}{255, 153, 153}
\begin{table}[h!]
\centering
\resizebox{\textwidth}{!}{
\begin{tabular}{lccccccccc}
\hline
scene & 3DGS & 2DGS & PGSR & EntON($knn_{100}$) & EntON($knn_{75}$) & EntON($knn_{50}$) & EntON($knn_{25}$) \\
\hline
building1  & 0.184\cellcolor{yellow} & 0.193 & 0.197 & 0.184 & \cellcolor{red}0.179  &  \cellcolor{orange}0.182 & 0.187\\
building2  & 0.185 & \cellcolor{red}0.160 & 0.177 & 0.168 & 0.169  & \cellcolor{yellow}0.167  & \cellcolor{orange}0.163 \\
\hline
mean  & 0.185 & 0.177 & 0.187 & 0.176 & 0.174 & 0.175 & 0.175 \\
\hline
\end{tabular}
}
\caption{Surface accuracy. \textbf{Geometric accuracy} comparison on the TUM2TWIN dataset with Chamfer cloud-to-cloud distances $\downarrow$ in m for points $\leq$0.5\,m from the reference, according to the DTU evaluation script. Best results are highlighted as \colorbox{red}{1st}, \colorbox{orange}{2nd}, and \colorbox{yellow}{3rd}. Mean scores are listed at the bottom. The training incorporates 15\,000 iterations.}
\label{tab:tnt_surface}
\end{table}

\paragraph{Rendering Quality}

We assess rendering quality using peak signal-to-noise ratio (PSNR), where higher values indicate better photometric fidelity. Across all neighborhood sizes, EntON achieves consistently strong PSNR values, in average with the best variant ($k_{\text{nn}}=75$) reaching 31.54\,dB and the mean across all neighborhood sizes being 31.44\,dB. EntON surpasses the rendering quality of 2DGS (28.95\,dB) and PGSR (28.87\,dB), while remaining competitive with 3DGS (31.68\,dB). These results correspond to up to 8.9\% improvement compared to PGSR (9.2\% for the best variant) and 8.6\% compared to 2DGS (8.9\% best), while remaining comparable to 3DGS.

\definecolor{yellow}{RGB}{255, 255, 204}
\definecolor{orange}{RGB}{255, 204, 153}
\definecolor{red}{RGB}{255, 153, 153}
\begin{table}[h!]
\centering
\resizebox{\textwidth}{!}{
\begin{tabular}{lcccccccc}
\hline
scene & 3DGS & 2DGS & PGSR & EntON($knn_{100}$) & EntON($knn_{75}$) & EntON($knn_{50}$) & EntON($knn_{25}$) \\
building1  & \cellcolor{orange}32.36 & 29.90 & 30.05 & 31.55 & \cellcolor{red}32.57 & \cellcolor{yellow}32.09 & 31.94 \\
building2  & \cellcolor{red}31.00 & 28.00 & 27.69 & \cellcolor{orange}30.84 & 30.50 & 30.56 & \cellcolor{yellow}30.78 \\
\hline
mean  & 31.68 & 28.95 & 28.87 & 31.20 & 31.54 & 31.26 & 31.36 \\
\end{tabular}
}
\caption{\textbf{Rendering quality} comparison on the TUM2TWIN dataset. We report the PSNR $\uparrow$ in dB. Best results are highlighted as \colorbox{red}{1st}, \colorbox{orange}{2nd}, and \colorbox{yellow}{3rd}. Mean scores are listed at the bottom. The training incorporates 15\,000 iterations.}
\label{tab:tnt_psnr}
\end{table}

\paragraph{Efficiency}

Efficiency is evaluated in terms of the number of Gaussians and training time. EntON achieves a mean number of 2\,435\,371 Gaussians across neighborhood sizes, with the most compact variant ($k_{\text{nn}}=50$) using only 2\,362\,827, clearly more compact than all baselines. This corresponds to up to 29.3\% fewer Gaussians compared to 3DGS, 15.1\% fewer than 2DGS, and 8.1\% fewer than PGSR for the best variant. Training time for 15\,000 iterations show that EntON requires on average 22.27\,min, with the fastest variant completing in 19.30\,min. This corresponds to up to 51.0\% faster training compared to PGSR, while remaining competitive with 2DGS and slightly slower than 3DGS on average.

\definecolor{yellow}{RGB}{255, 255, 204}
\definecolor{orange}{RGB}{255, 204, 153}
\definecolor{red}{RGB}{255, 153, 153}
\begin{table}[h!]
\centering
\resizebox{\textwidth}{!}{
\begin{tabular}{lcccccccccc}
\hline
scene & 3DGS & 2DGS & PGSR & EntON($knn_{100}$) & EntON($knn_{75}$) & EntON($knn_{50}$) & EntON($knn_{25}$) \\
\hline
building1  & 3\,622\,623 & 3\,457\,686 & 3\,495\,252   & 2\,716\,037 & \cellcolor{yellow}2\,693\,175 & \cellcolor{orange}2\,500\,641 & \cellcolor{red}2\,475\,797 \\
building2  & 3\,037\,856 & \cellcolor{orange}2\,087\,933 & \cellcolor{red}1\,625\,555  & 2\,306\,556 & 2\,247\,377 & \cellcolor{yellow}2\,225\,013 & 2\,278\,891 \\
\hline
mean  & 3\,330\,240 & 2\,772\,810 & 2\,560\,404 & 2\,511\,297 & 2\,470\,276 & 2\,362\,827 & 2\,377\,344 \\

\hline
\end{tabular}
}
\caption{\textbf{Number of Gaussians} on the TUM2TWIN dataset. We report the total number of Gaussians $\downarrow$. Best results are highlighted as \colorbox{red}{1st}, \colorbox{orange}{2nd}, and \colorbox{yellow}{3rd}. Mean scores are listed at the bottom. The training incorporates 15\,000 iterations.}
\label{tab:tnt_gaussians}
\end{table}

\definecolor{yellow}{RGB}{255, 255, 204}
\definecolor{orange}{RGB}{255, 204, 153}
\definecolor{red}{RGB}{255, 153, 153}
\begin{table}[h!]
\centering
\resizebox{\textwidth}{!}{
\begin{tabular}{lcccccccc}
\hline
scene & 3DGS & 2DGS & PGSR & EntON($knn_{100}$) & EntON($knn_{75}$) & EntON($knn_{50}$) & EntON($knn_{25}$) \\
building1  & \cellcolor{red}18.35 & 25.32 & 40.48 & 27.75 & 26.24 & \cellcolor{yellow}22.82 & \cellcolor{orange}20.56 \\
building2  & \cellcolor{red}17.42 & 22.77 & 37.97 & 22.21 & 20.93 & \cellcolor{yellow}19.19 & \cellcolor{orange}18.04 \\
\hline
mean  & 17.89 & 24.05 & 39.23 & 24.98 & 23.59 & 21.00 & 19.30 \\

\hline
\end{tabular}
}
\caption{\textbf{Training time} comparison on the TUM2TWIN dataset. We report the minutes for 15\,000 iterations in min. Best results are highlighted as \colorbox{red}{1st}, \colorbox{orange}{2nd}, and \colorbox{yellow}{3rd}. Mean scores are listed at the bottom.}
\label{tab:tnt_time}
\end{table}

Overall, EntON demonstrates a compelling balance between rendering quality, geometric accuracy, memory efficiency, and computational cost on large-scale urban scenes. Smaller neighborhood sizes prioritize geometric accuracy and efficiency in terms of compactness, whereas larger neighborhoods slightly enhance photometric quality, by showing that the method scales effectively while outperforming 2DGS and PGSR, and remaining competitive with 3DGS.

\subsubsection{Qualitative Results}\label{sec:results_2_quali}

\paragraph{Geometric Accuracy}

\begin{figure*}[htbp]
    \vspace{-4cm}
    \centering
    \begin{tabular}{c c c}
        \textbf{} & \scriptsize\textbf{building1} & \scriptsize\textbf{building2} \\[1ex]
                \vspace{-5mm}
         \rotatebox{90}{\scriptsize\textbf{3DGS}} &
        \includegraphics[width=0.45\linewidth]{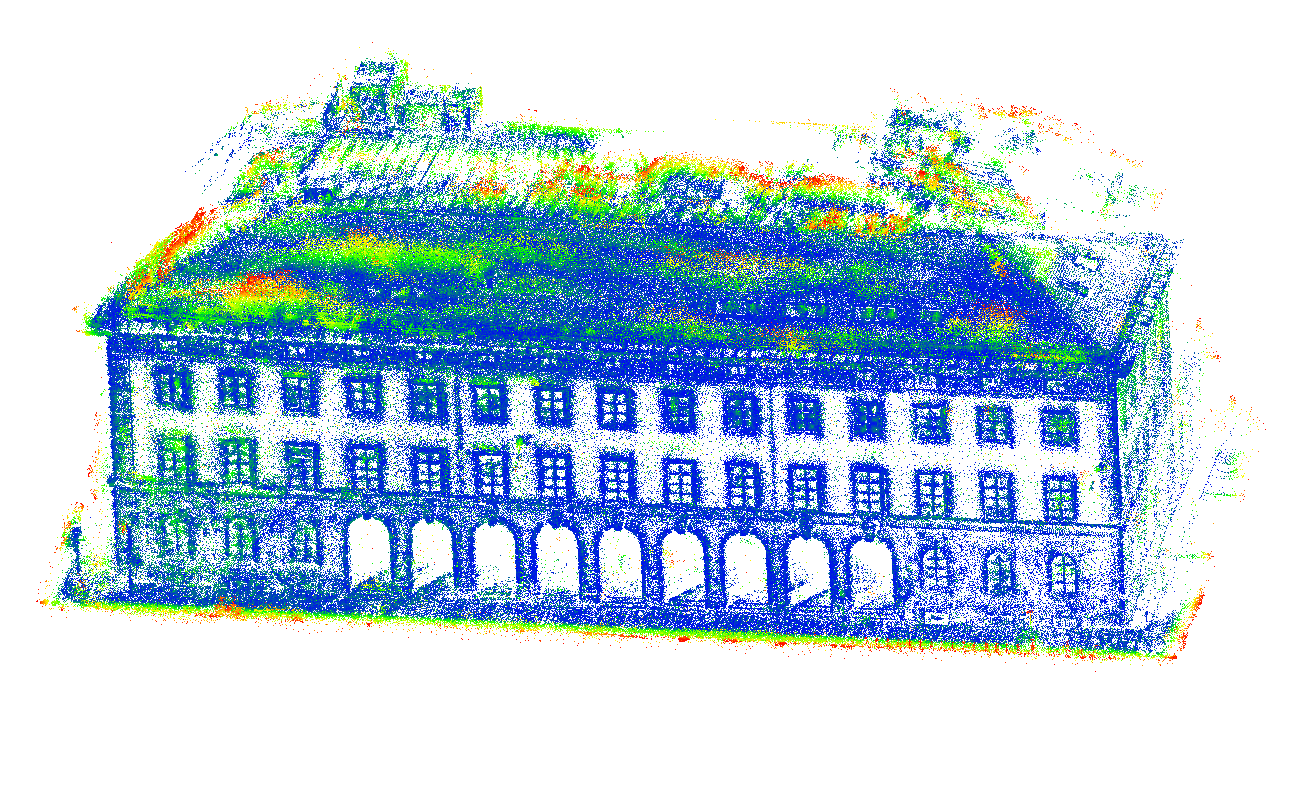} &
        \includegraphics[width=0.43\linewidth]{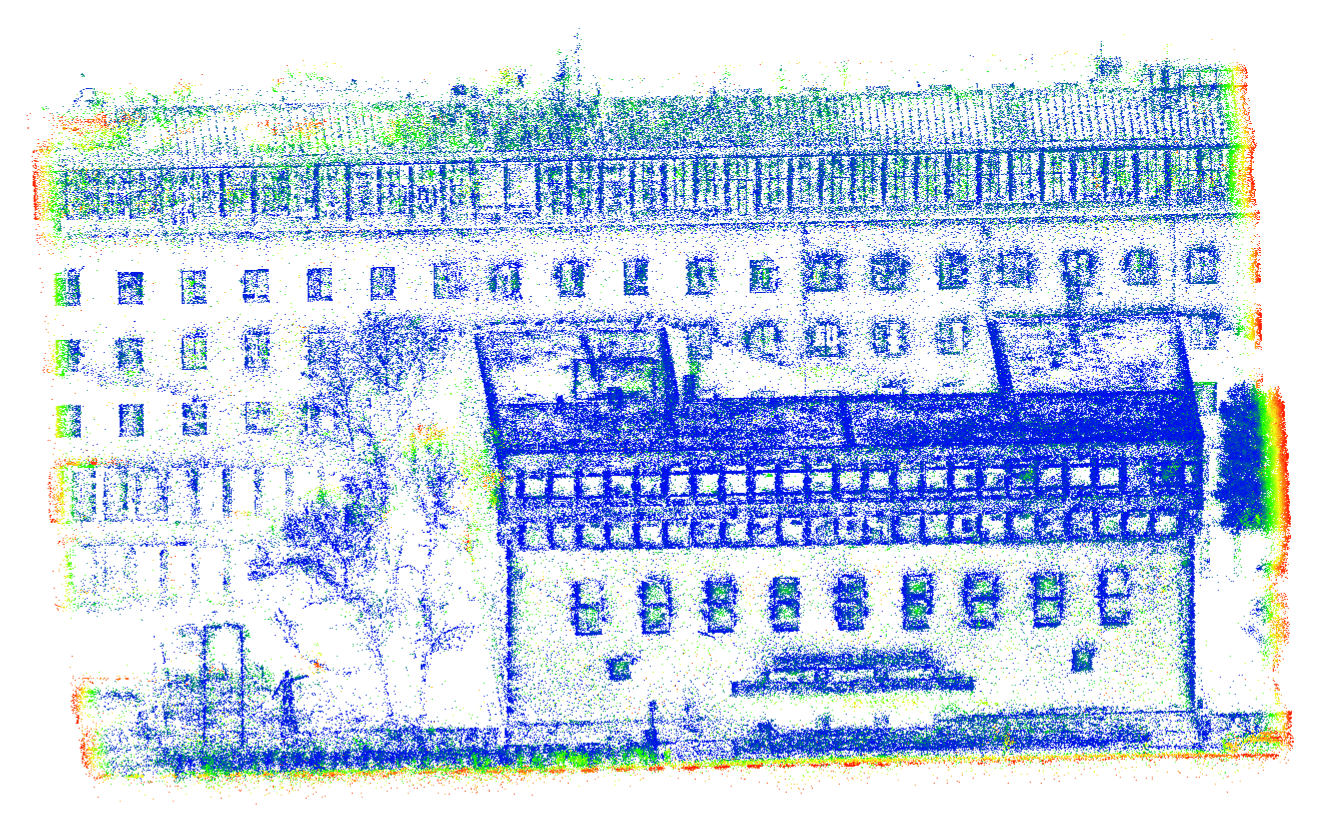} \\[2ex]
            \vspace{-5mm}
         \rotatebox{90}{\scriptsize\textbf{2DGS}} &
        \includegraphics[width=0.45\linewidth]{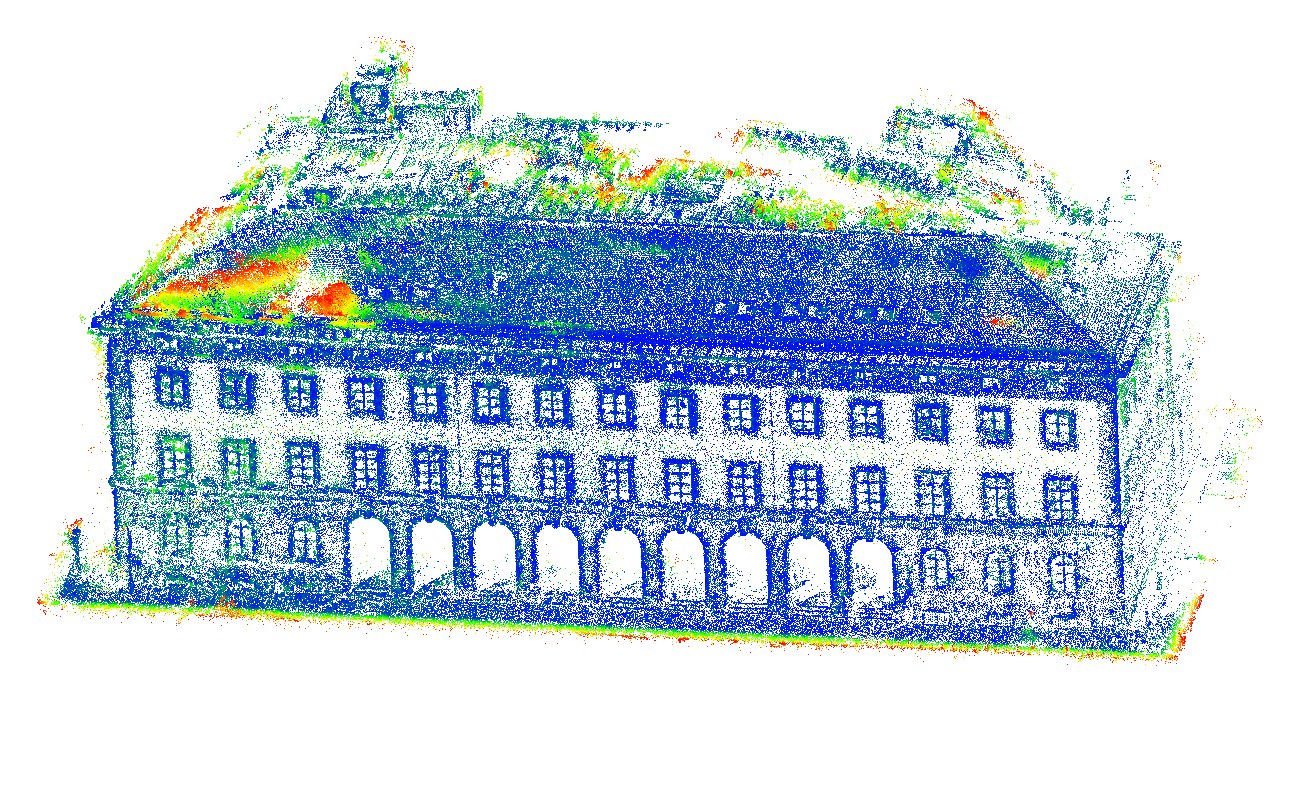} &
        \includegraphics[width=0.43\linewidth]{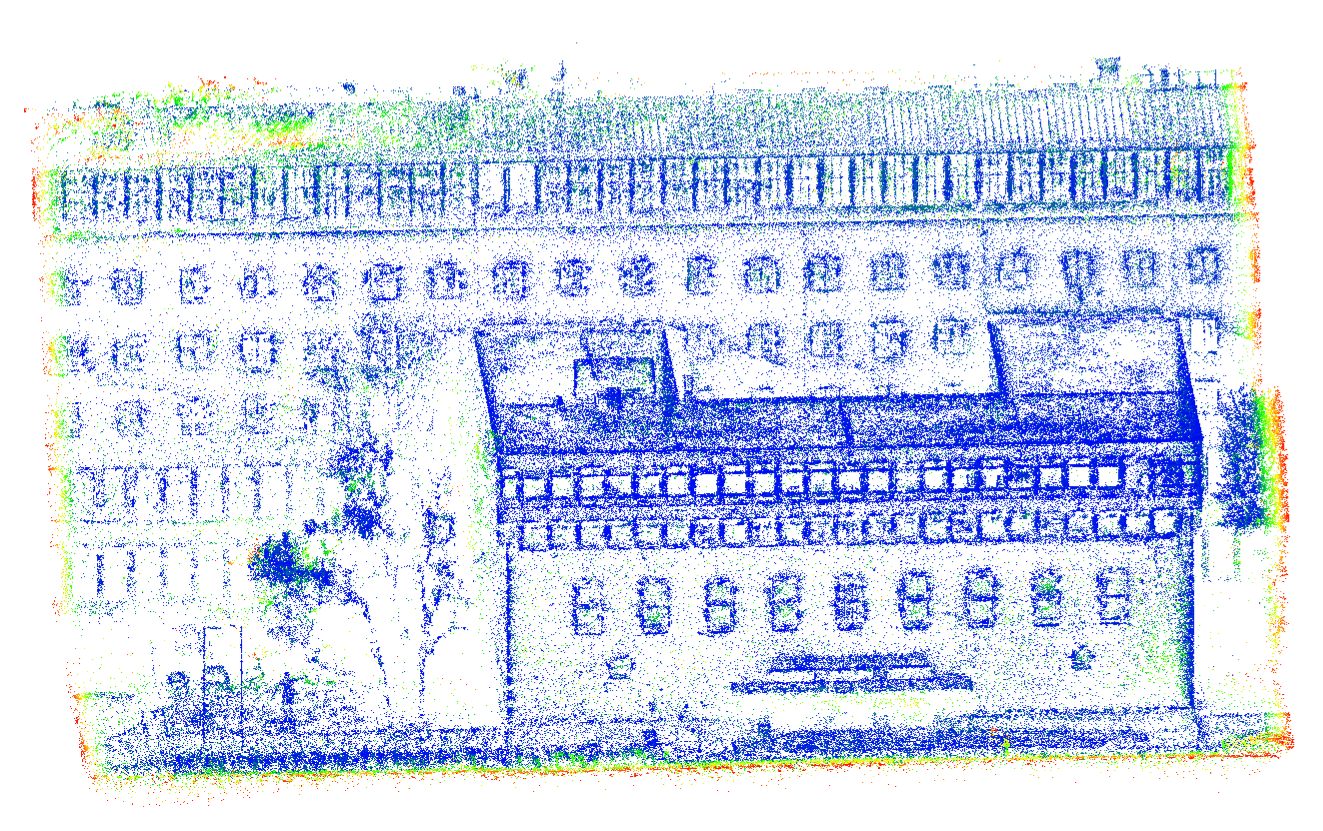} \\[2ex]
            \vspace{-5mm}
                 \rotatebox{90}{\scriptsize\textbf{PGSR}} &
        \includegraphics[width=0.45\linewidth]{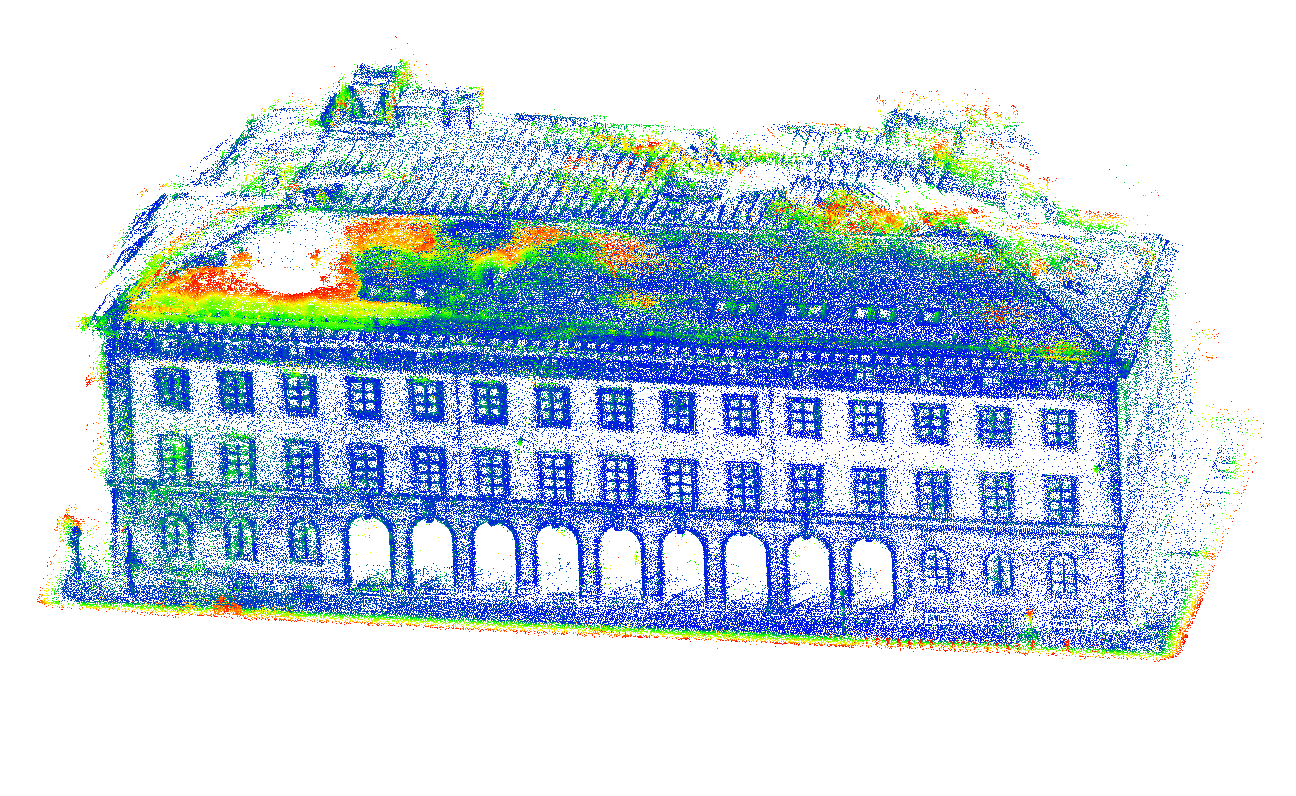} &
        \includegraphics[width=0.43\linewidth]{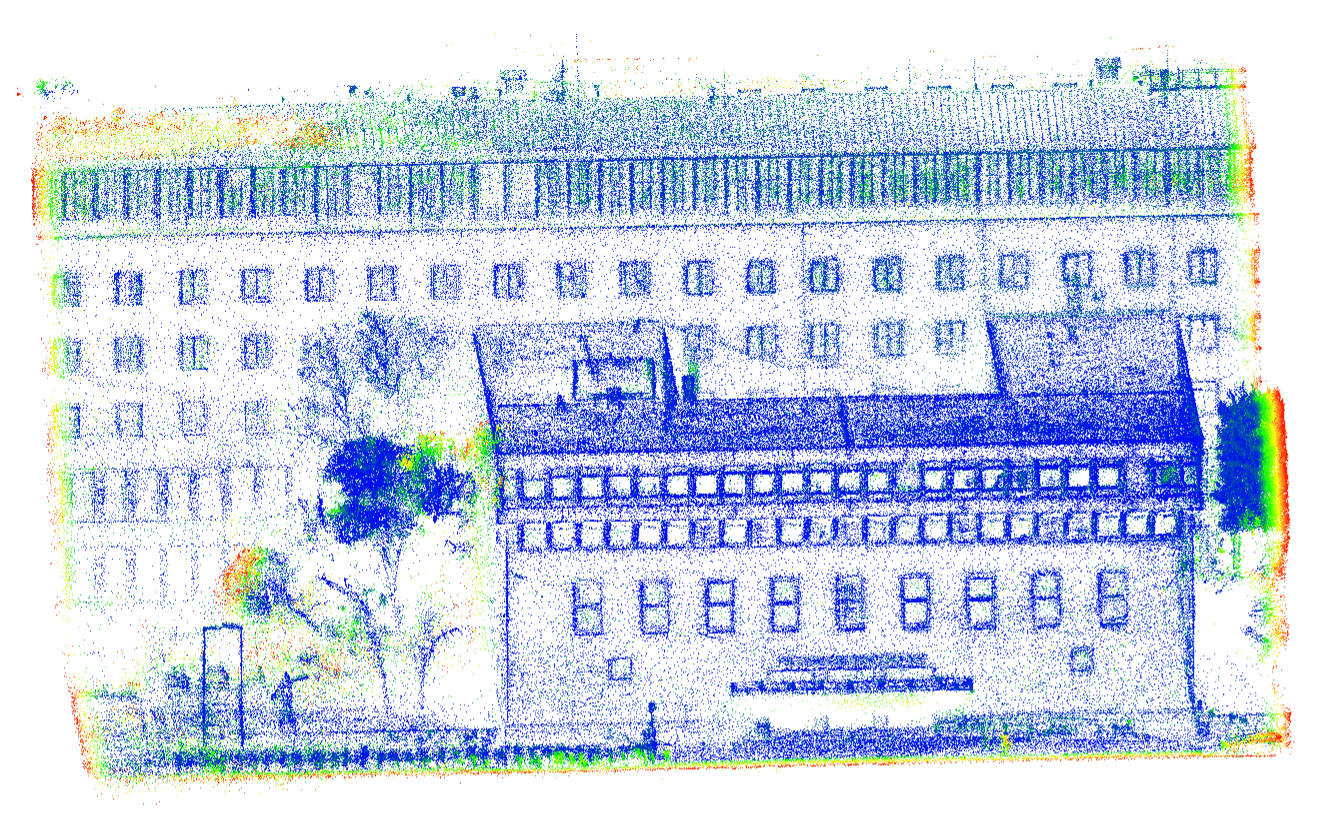} \\[2ex]
            \vspace{-5mm}
         \rotatebox{90}{\scriptsize\textbf{EntON}} &
        \includegraphics[width=0.45\linewidth]{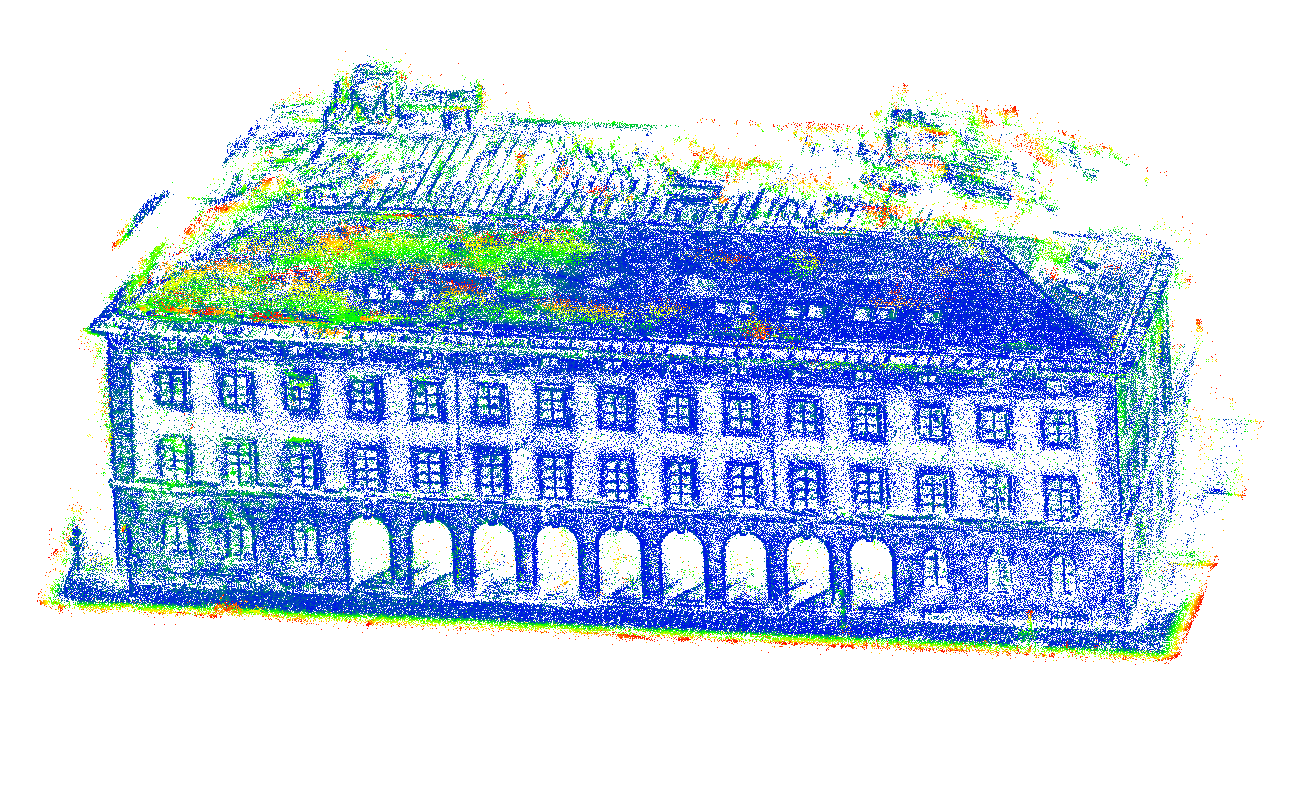} &
        \includegraphics[width=0.43\linewidth]{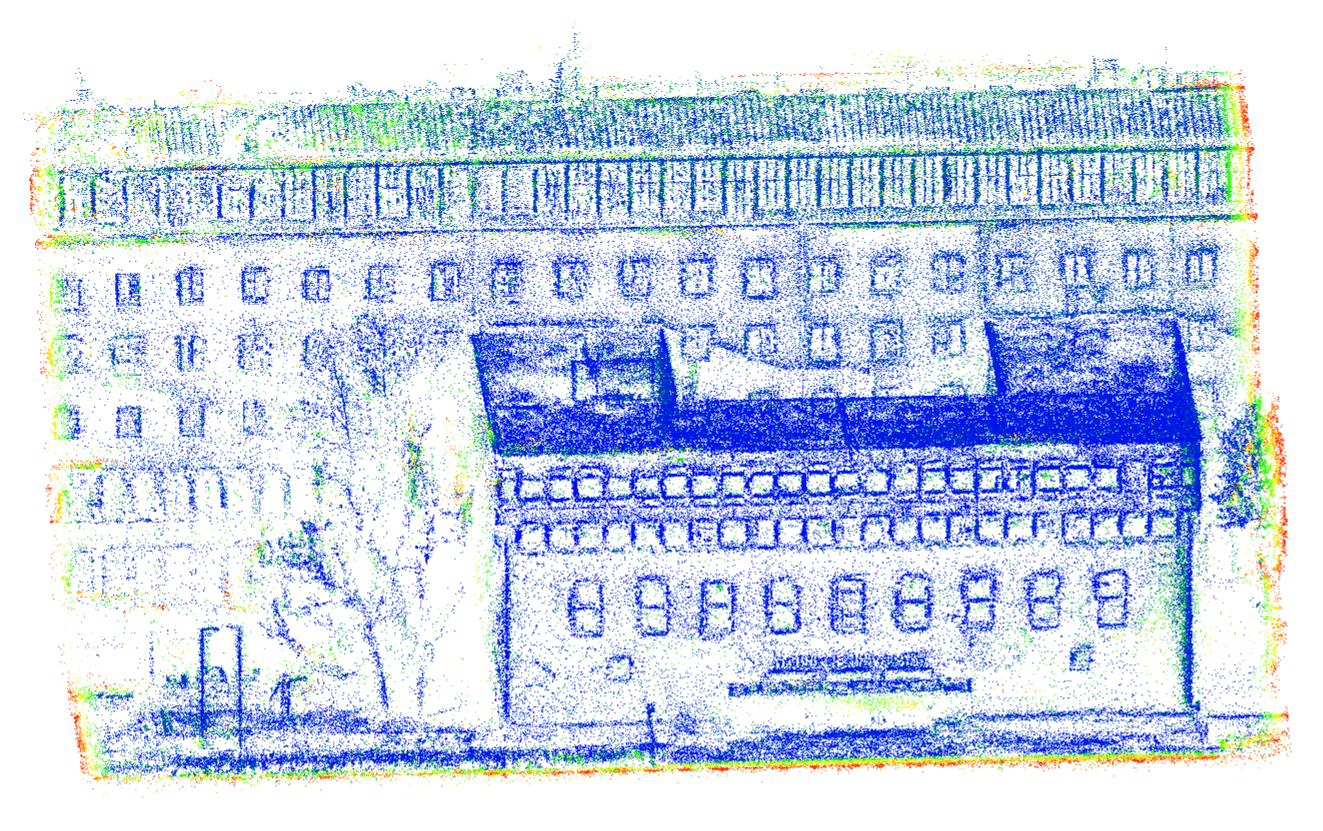} \\[2ex]
            \vspace{-5mm}
                 \rotatebox{90}{\scriptsize\textbf{GT}} &
        \includegraphics[width=0.45\linewidth]{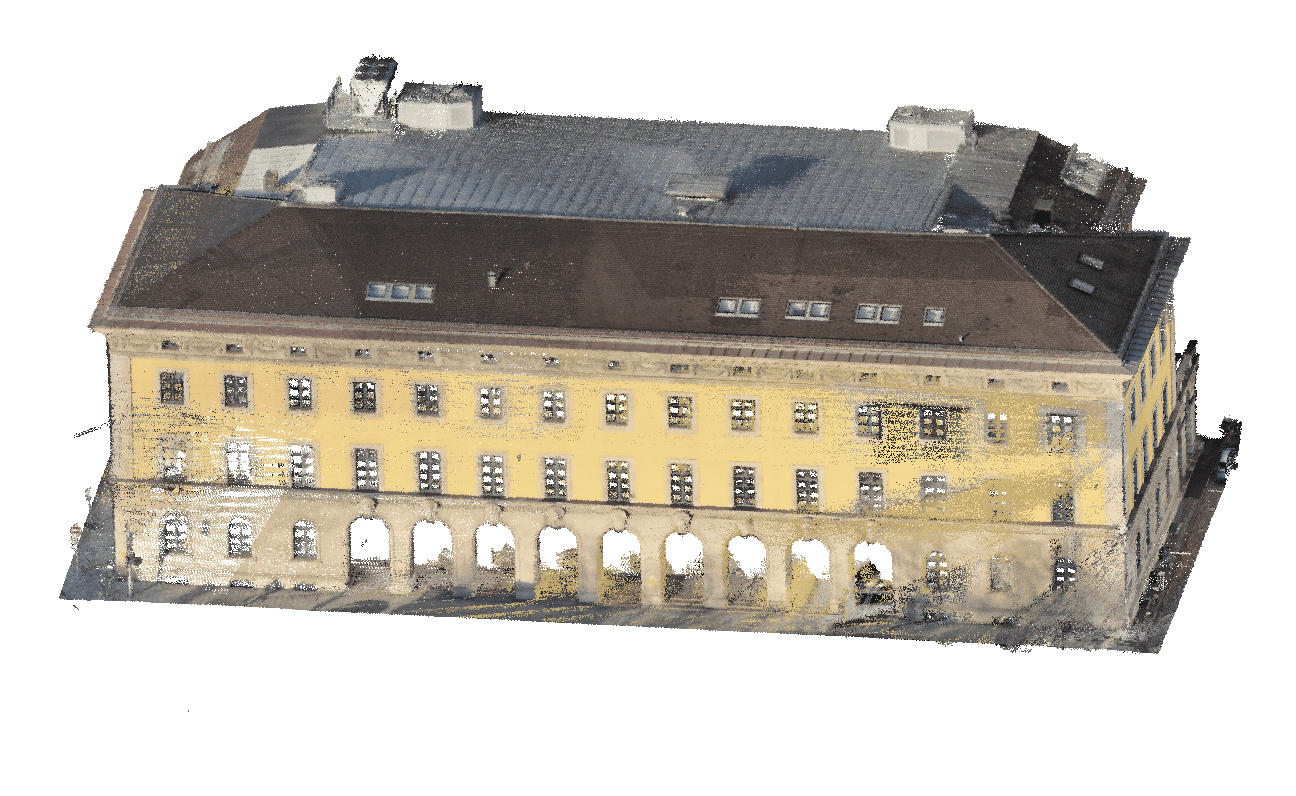} &
        \includegraphics[width=0.43\linewidth]{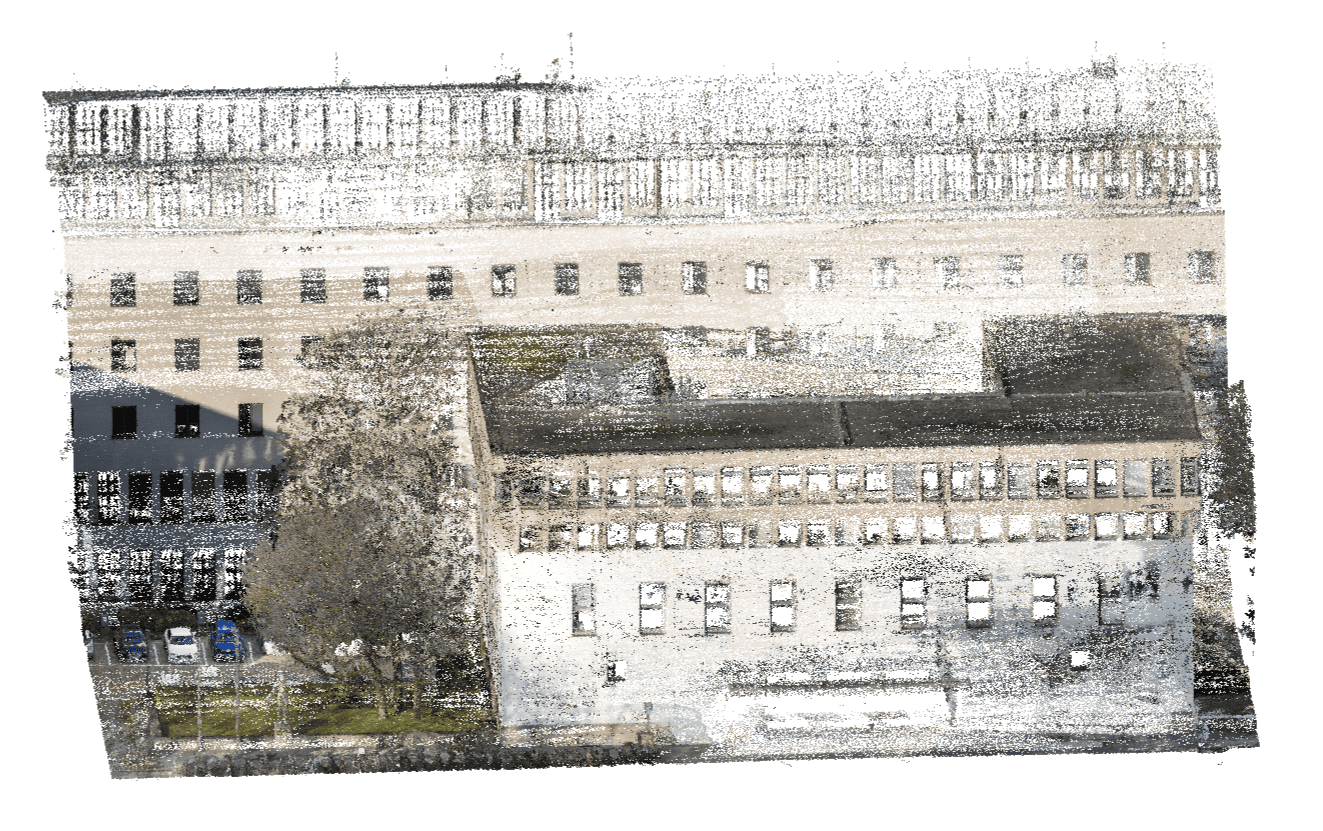} \\[2ex]
    \end{tabular}
    \includegraphics[width=0.3\textwidth]{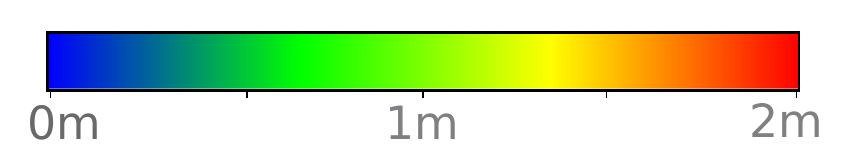} \vspace{-2mm}
    \caption{\textbf{Geometric accuracy} comparison of 3DGS, 2DGS, PGSR and EntON TUM2TWIN dataset with Chamfer cloud-to-cloud distances $\downarrow$. Color values are cropped at 2\,m distance.}
    \label{fig:Qualitative_c2c_tt}
\end{figure*}

The geometric accuracy of Gaussian centers on the TUM2TWIN large scale dataset, evaluated using the Chamfer cloud-to-cloud distance (Figure \ref{fig:Qualitative_c2c_tt}), demonstrate high surface accuracy for EntON and confirm the quantitative findings.

\paragraph{Rendering Quality}

\begin{figure*}[htbp]
    \vspace{-3.5cm}
    \centering
    \begin{tabular}{c c c}
        \textbf{} & \scriptsize\textbf{building1} & \scriptsize\textbf{building2} \\[1ex]
                \vspace{-5mm}
         \rotatebox{90}{\scriptsize\textbf{3DGS}} &
        \includegraphics[width=0.45\linewidth]{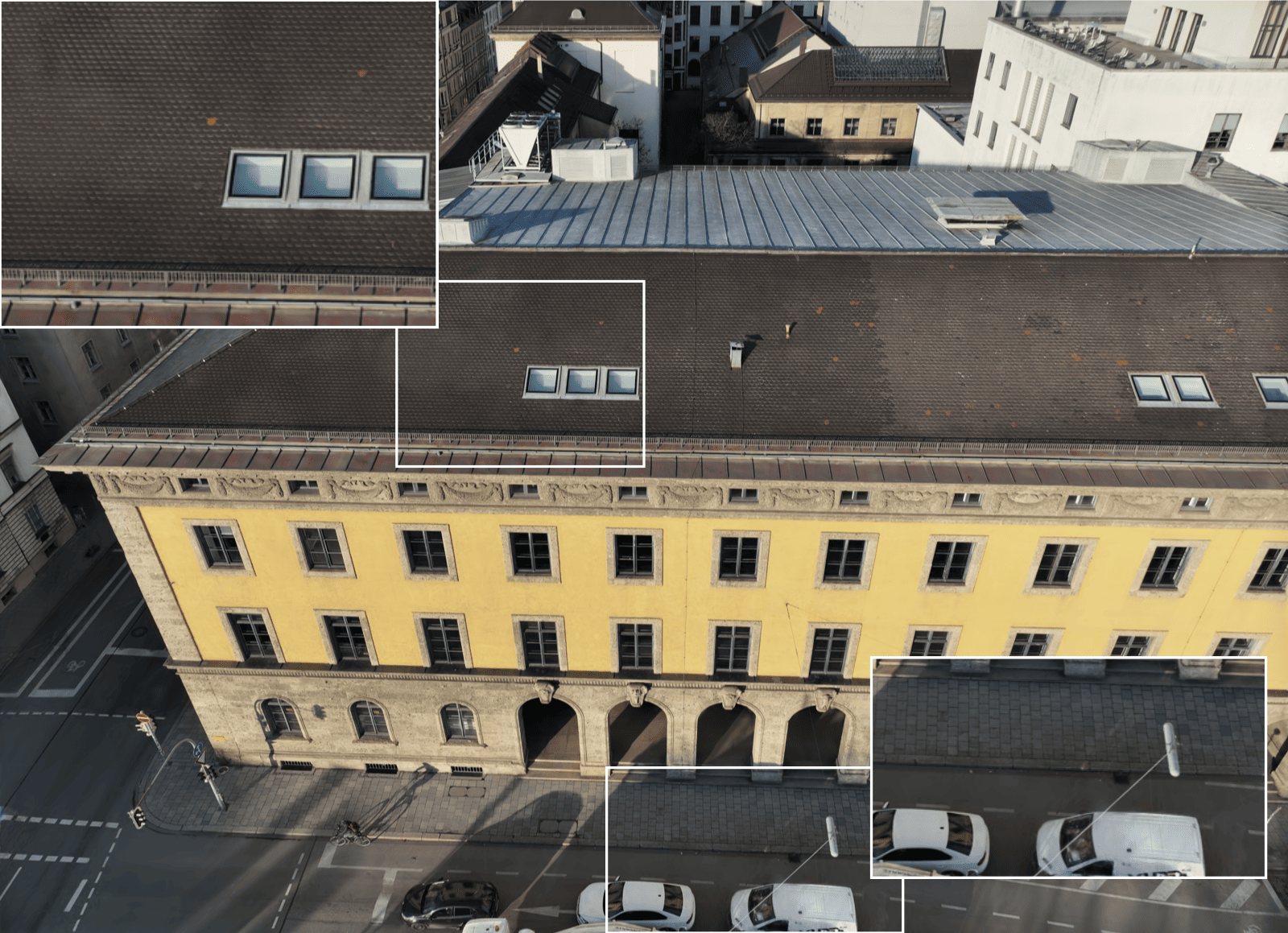} &
        \includegraphics[width=0.45\linewidth]{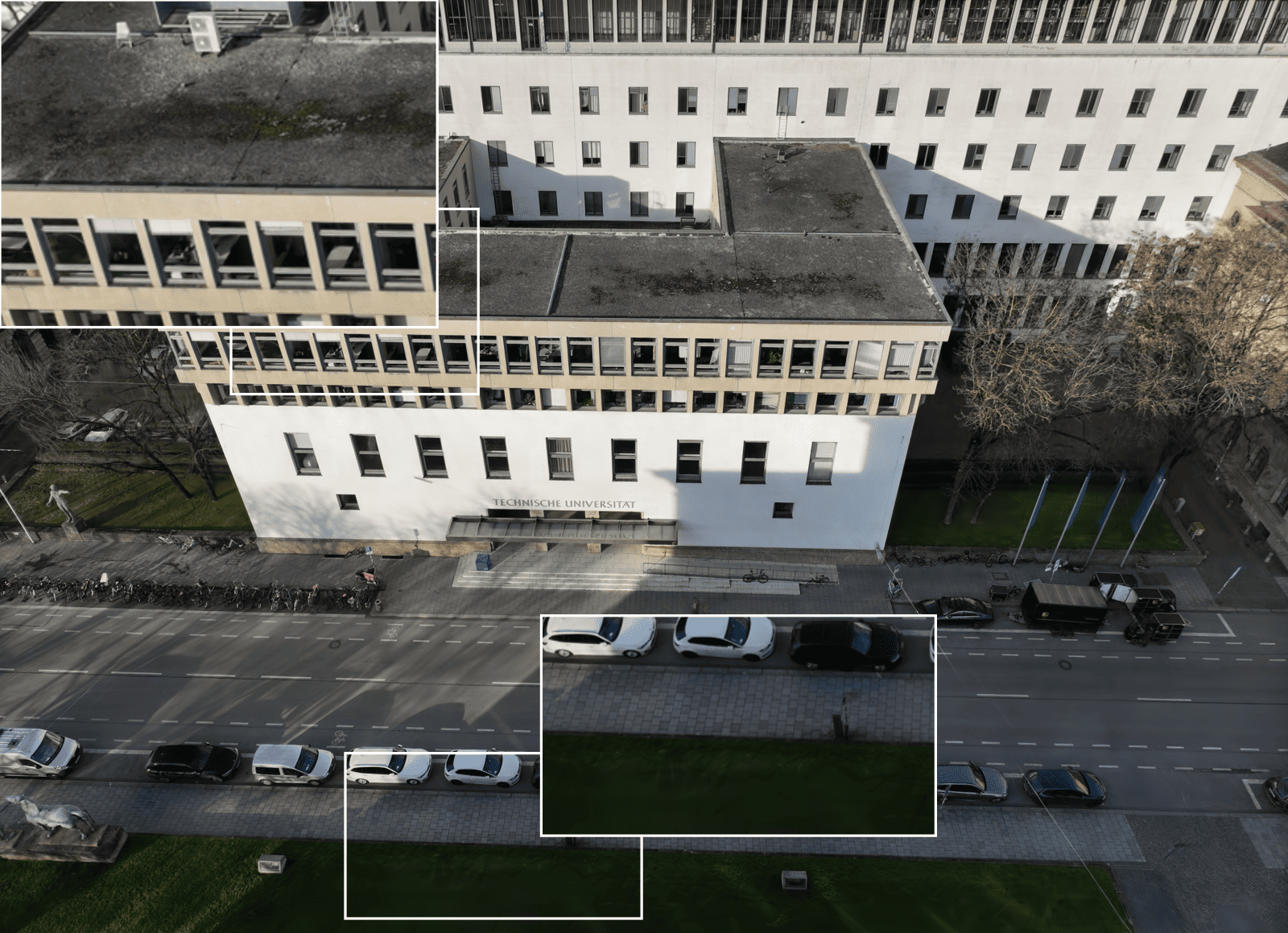} \\[2ex]
            \vspace{-5mm}
         \rotatebox{90}{\scriptsize\textbf{2DGS}} &
        \includegraphics[width=0.45\linewidth]{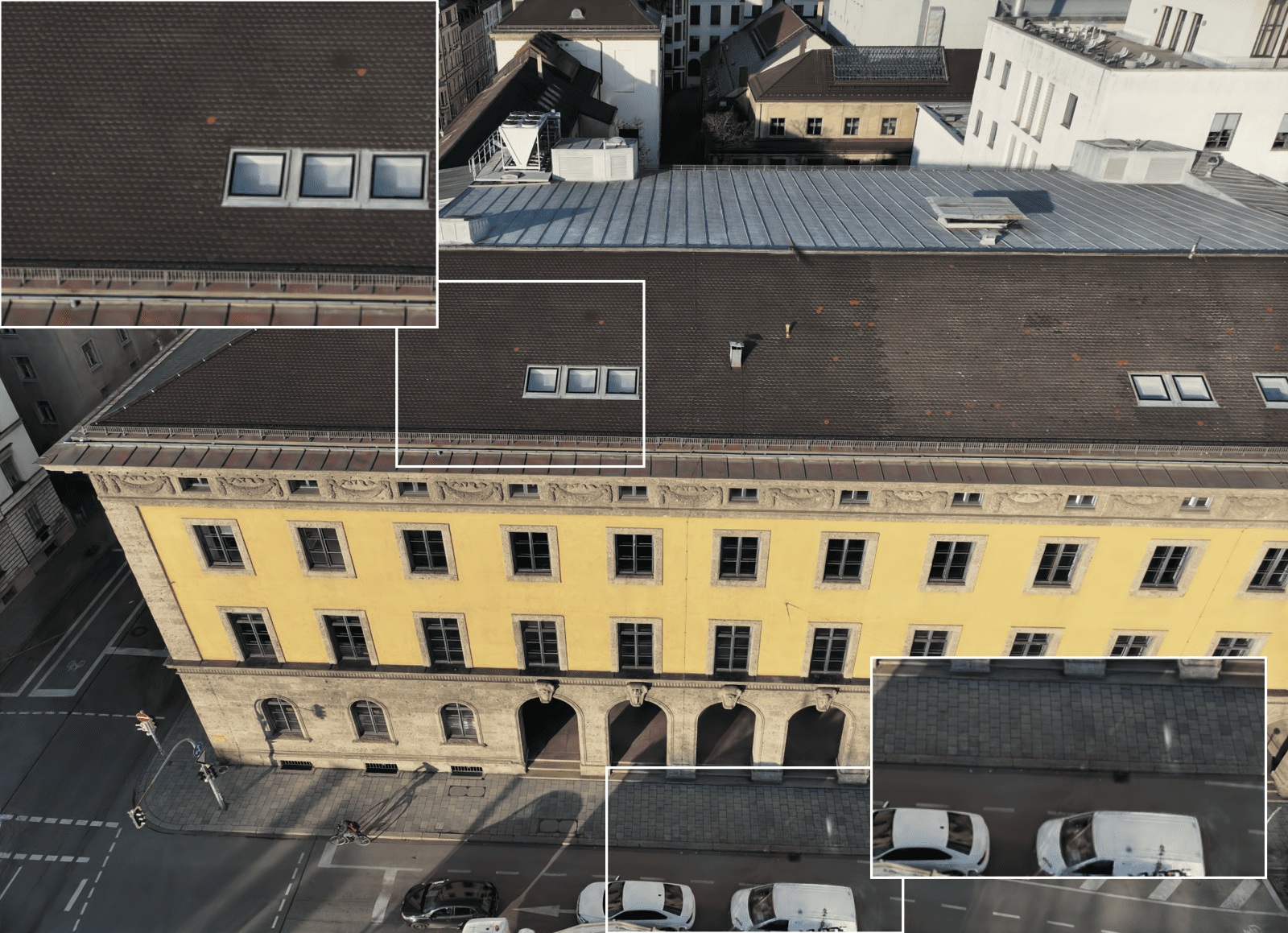} &
        \includegraphics[width=0.45\linewidth]{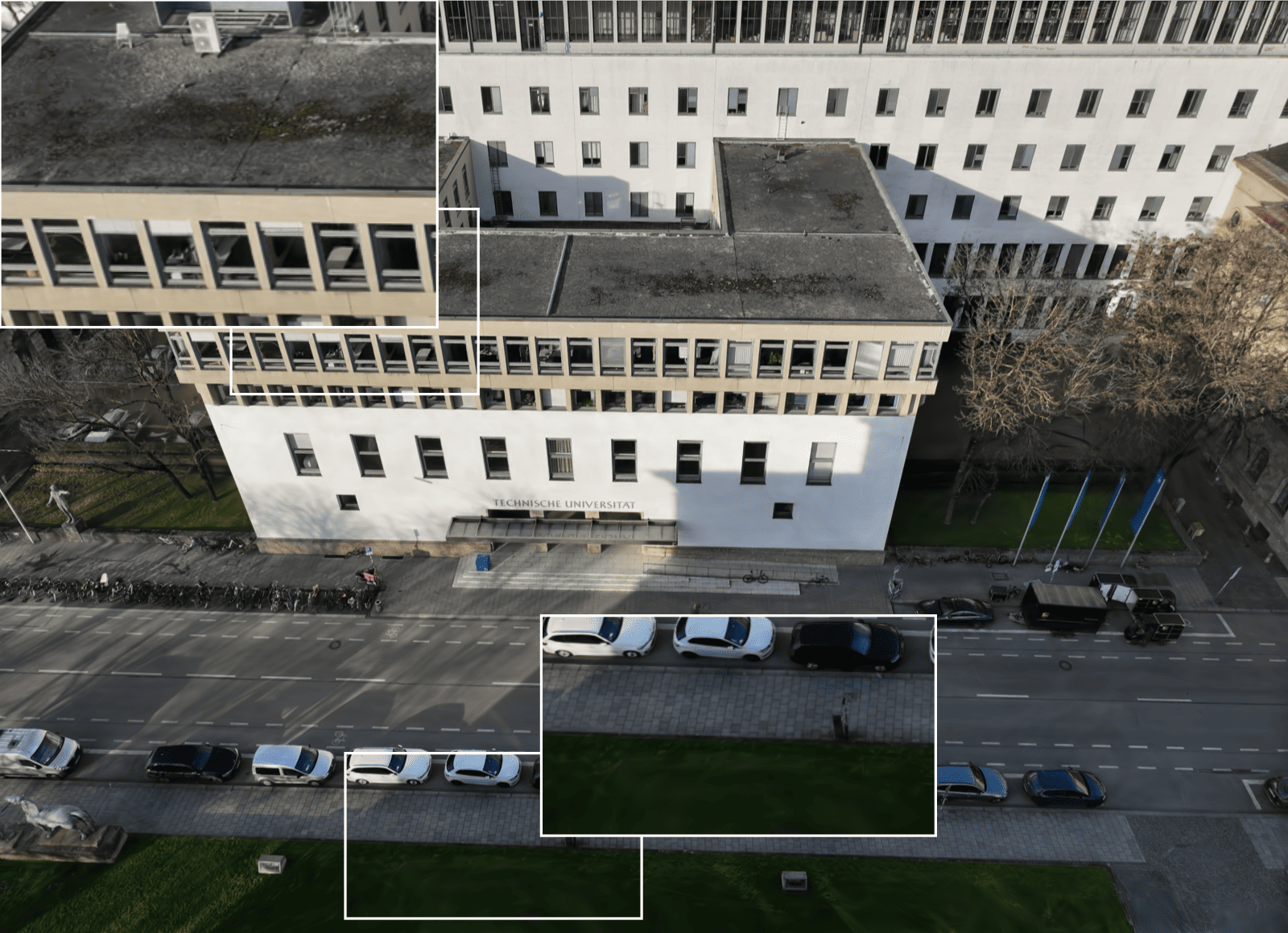} \\[2ex]
            \vspace{-5mm}
                 \rotatebox{90}{\scriptsize\textbf{PGSR}} &
        \includegraphics[width=0.45\linewidth]{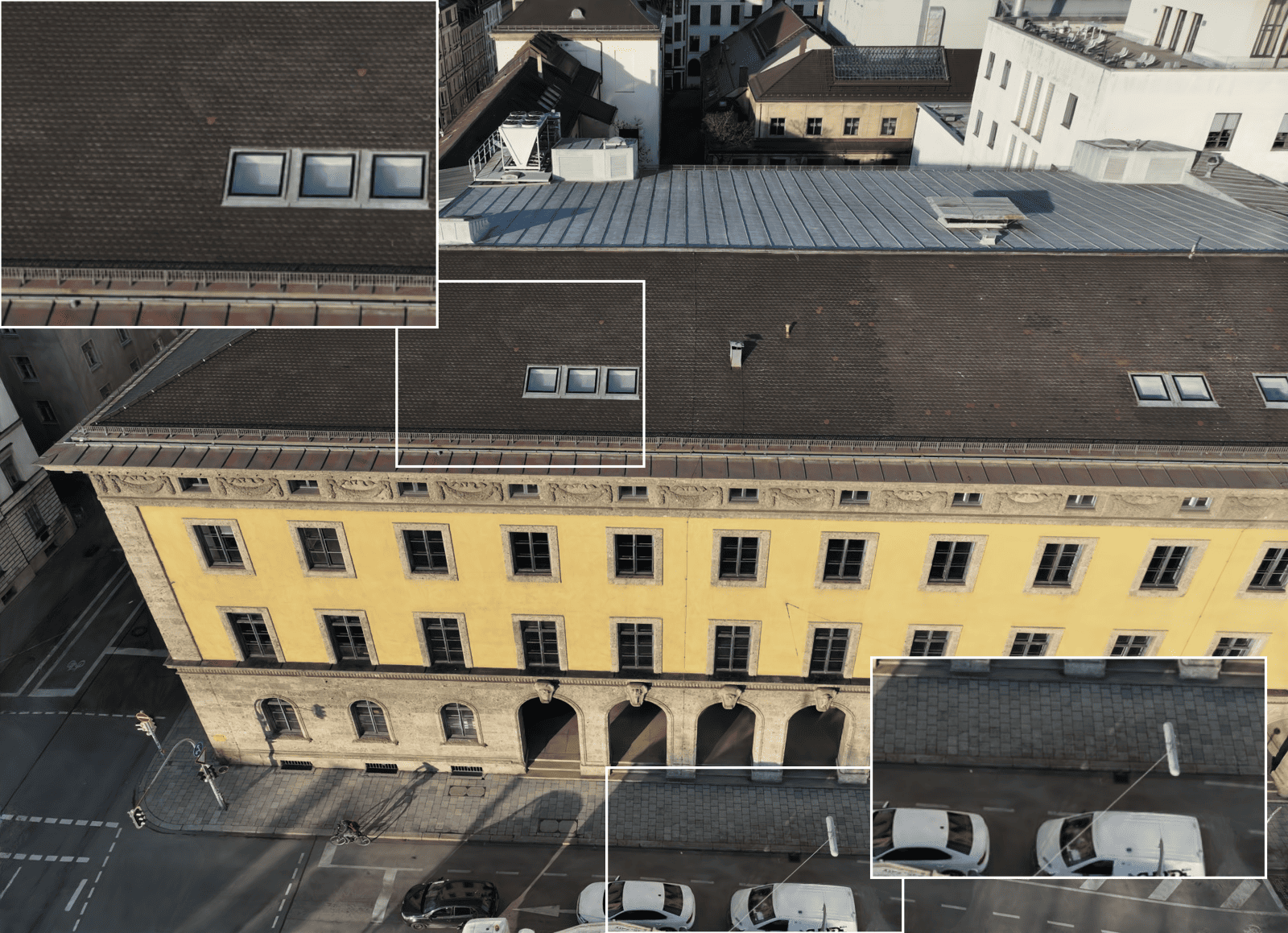} &
        \includegraphics[width=0.45\linewidth]{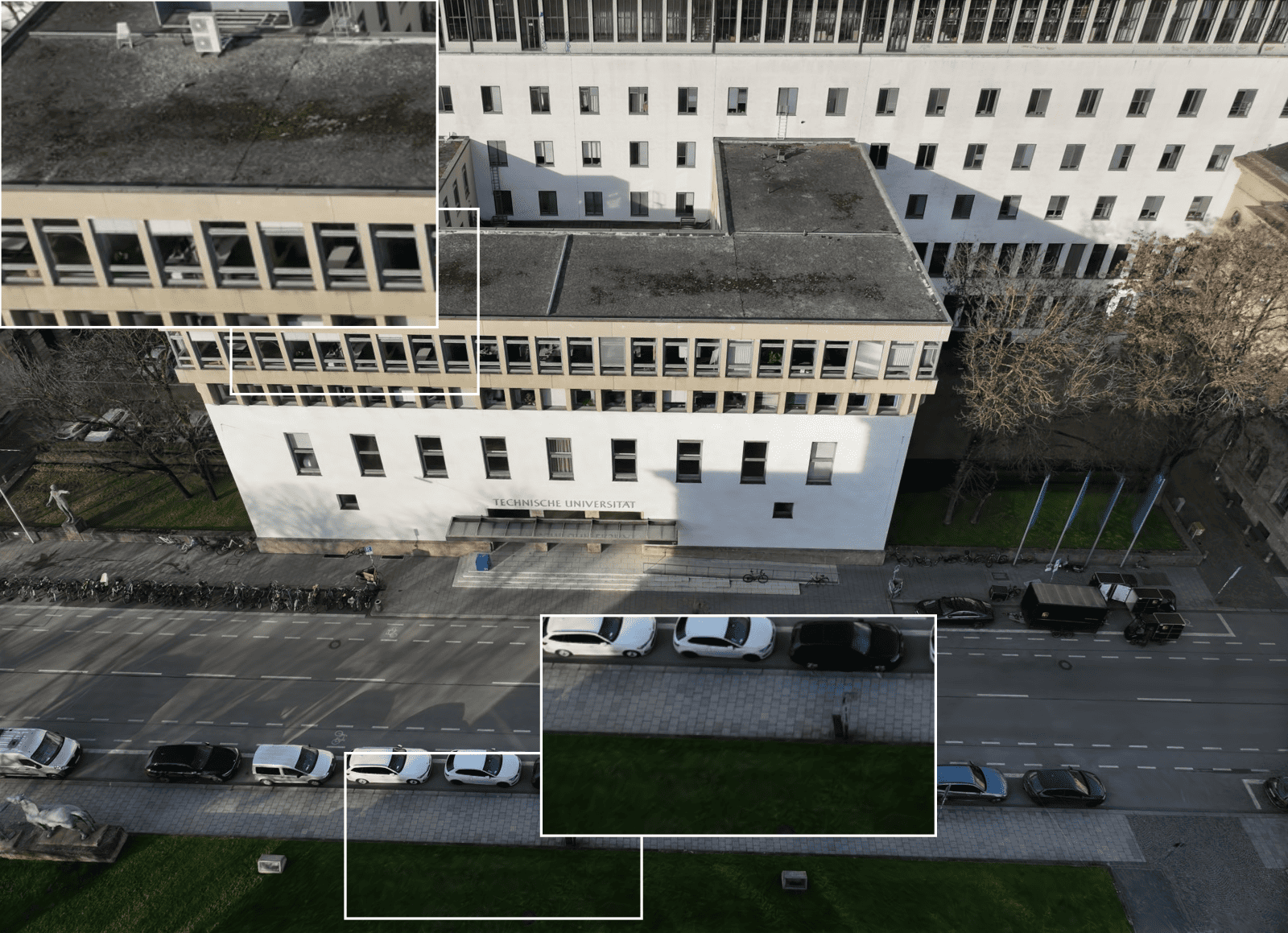} \\[2ex]
            \vspace{-5mm}
         \rotatebox{90}{\scriptsize\textbf{EntON}} &
        \includegraphics[width=0.45\linewidth]{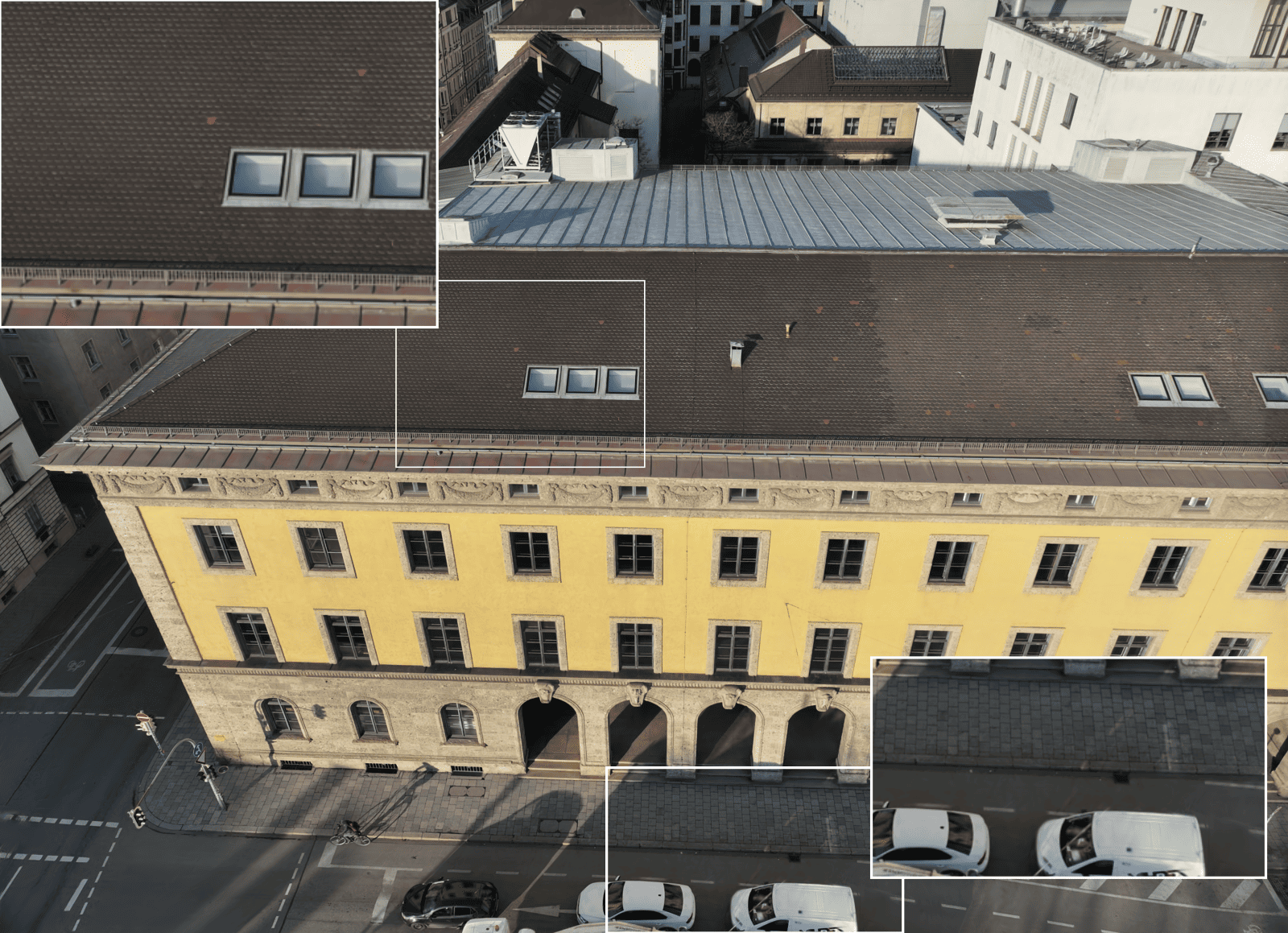} &
        \includegraphics[width=0.45\linewidth]{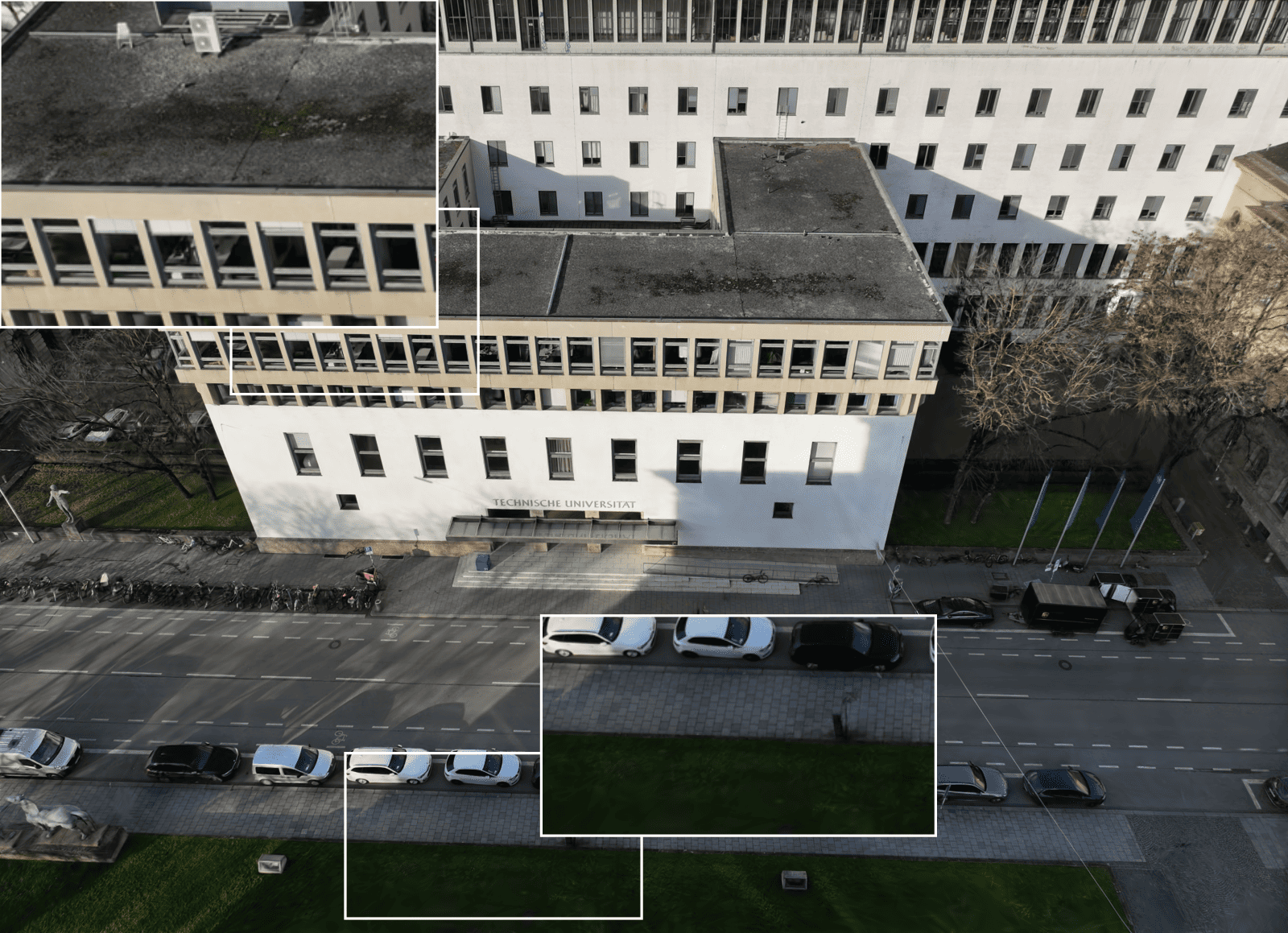} \\[2ex]
            \vspace{-5mm}
                 \rotatebox{90}{\scriptsize\textbf{GT}} &
        \includegraphics[width=0.45\linewidth]{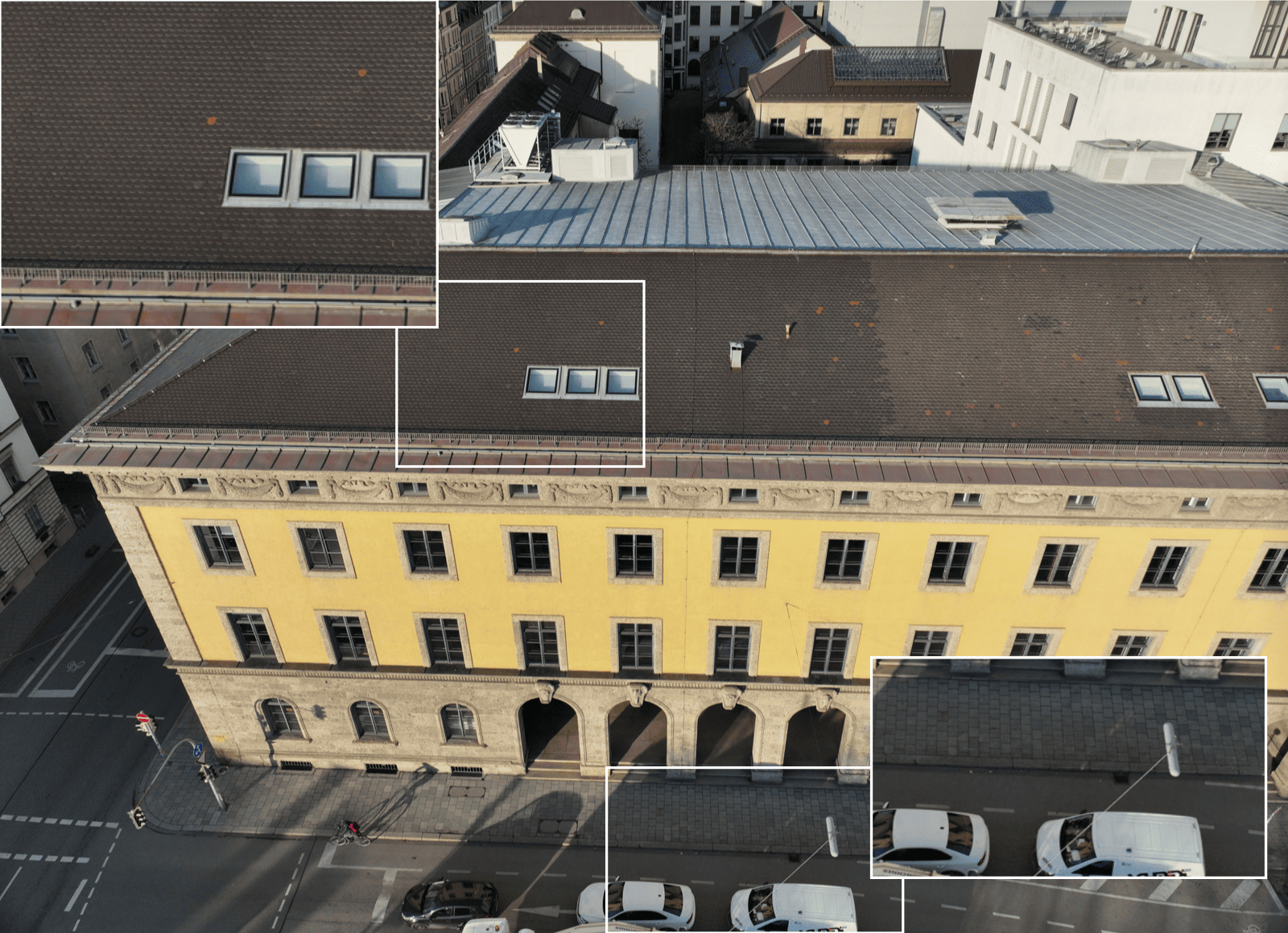} &
        \includegraphics[width=0.45\linewidth]{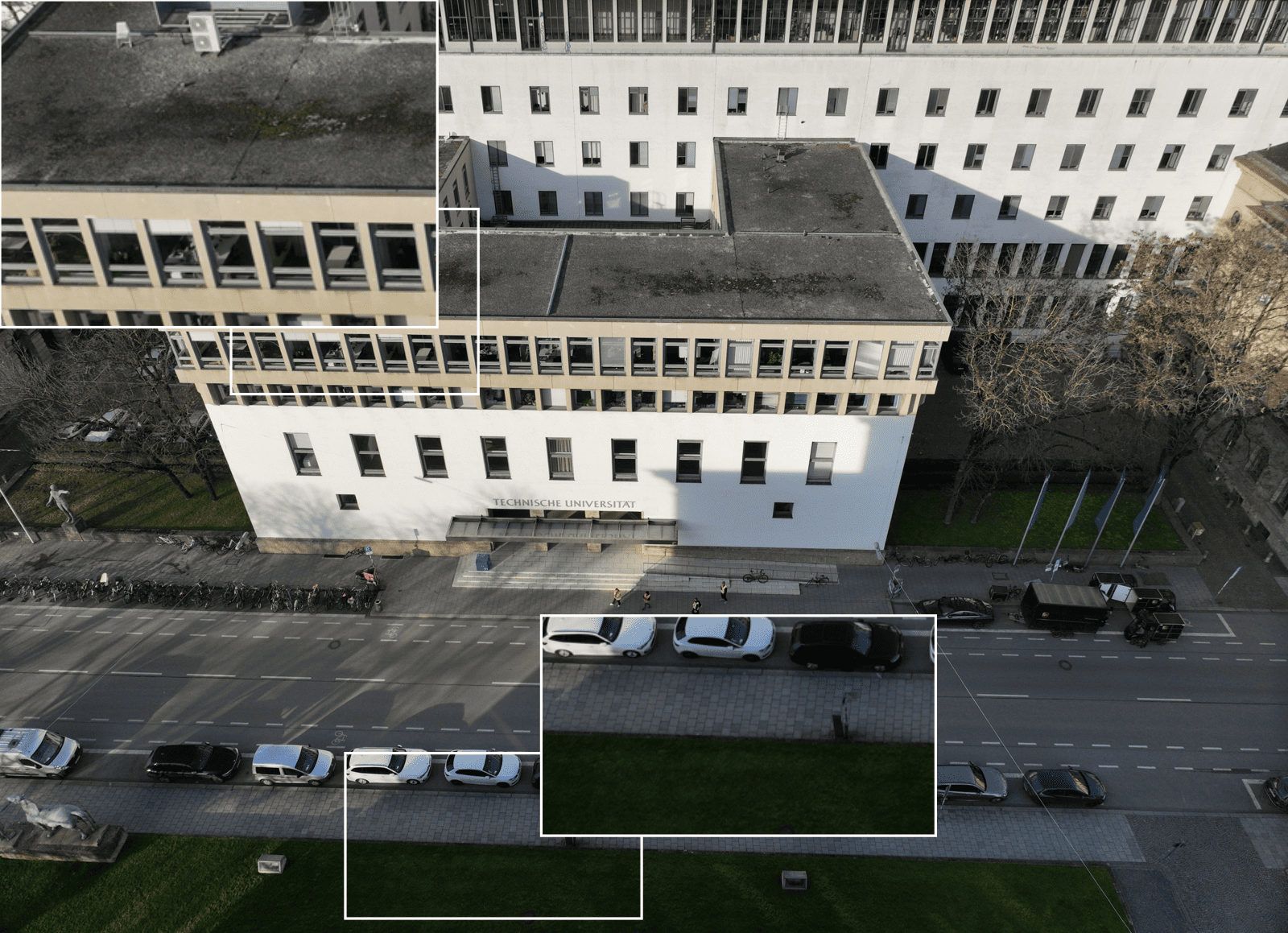} \\[2ex]
    \end{tabular}
    \caption{\textbf{Rendering quality} comparison of 3DGS, 2DGS, PGSR and EntON on the large-scale TUM2TWIN dataset, as well as ground truth (GT) images in original resolution.}
    \label{fig:Qualitative_rendering_tt}
\end{figure*}

The rendering results on the TUM2TWIN dataset (Figure \ref{fig:Qualitative_rendering_tt}) largely mirror the observations from the DTU small-scale dataset. EntON consistently reconstructs fine details more faithfully, avoiding the over-reconstructed areas that appear in 2DGS and, to a lesser extent, in 3DGS and PGSR. Flat surfaces, such as the roof and façade in building1 (see upper enlargement), are accurately captured, as are areas like the grass and roof in building2. On building1 (lower enlargement), EntO, and also 2DGS, prunes the Gaussiansof the street lamp, but EntON is able to reconstruct the vehicle and the underlying ground details with high fidelity regions that are less accurately represented in 2DGS and PGSR. Despite using fewer Gaussians, EntON achieves high rendering fidelity comparable to 3DGS, which produces similarly clean reconstructions but relies on a substantially higher number of Gaussians. 2DGS and PGSR, in contrast, use fewer Gaussians but generate partially blurred areas. Overall, the TUM2TWIN dataset confirms that EntON consistently balances rendering quality with compact and efficient scene representations, especially for man-made environments.

\section{Ablation Study}\label{sec:ablation}

This section presents an ablation study analyzing the Eigenentropy of Gaussians in outlier regions. We analyze the Eigenentropy of Gaussians in 3DGS after the training process based on 4 scenes, whose Gaussian centers lie more than 1\,mm from the reference point cloud to guide the choice of a pruning threshold. We than compute the mean Eigenentropy of these outlier Gaussians to guide the selection of the upper threshold for pruning in EntON.

\begin{figure*}[h!]
\centering
\begin{tabular}{c c c c c}
 & \scriptsize\textbf{scan40}
 & \scriptsize\textbf{scan55}
 & \scriptsize\textbf{scan106}
 & \scriptsize\textbf{scan122} \\[1ex]

\rotatebox{90}{\scriptsize\textbf{Eigenentropy}} &
\includegraphics[width=0.2\linewidth]{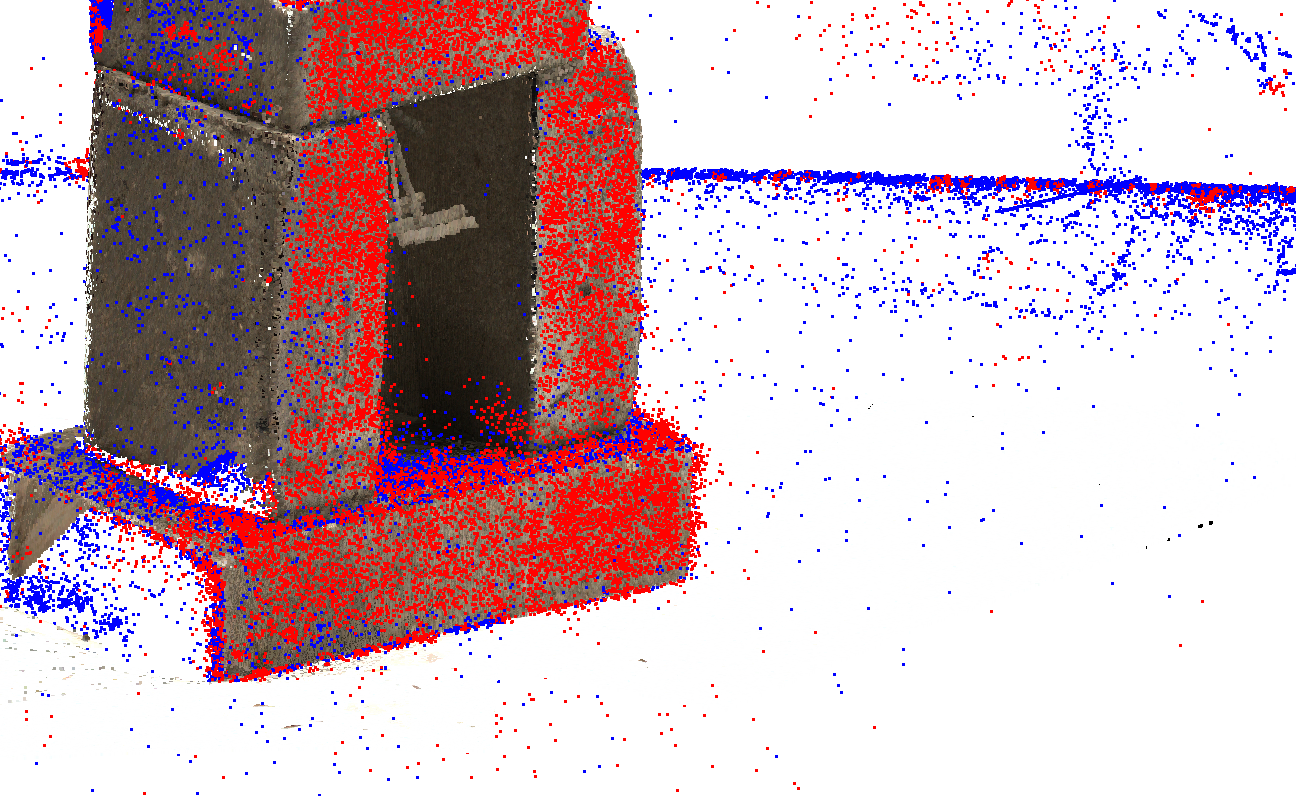} &
\includegraphics[width=0.2\linewidth]{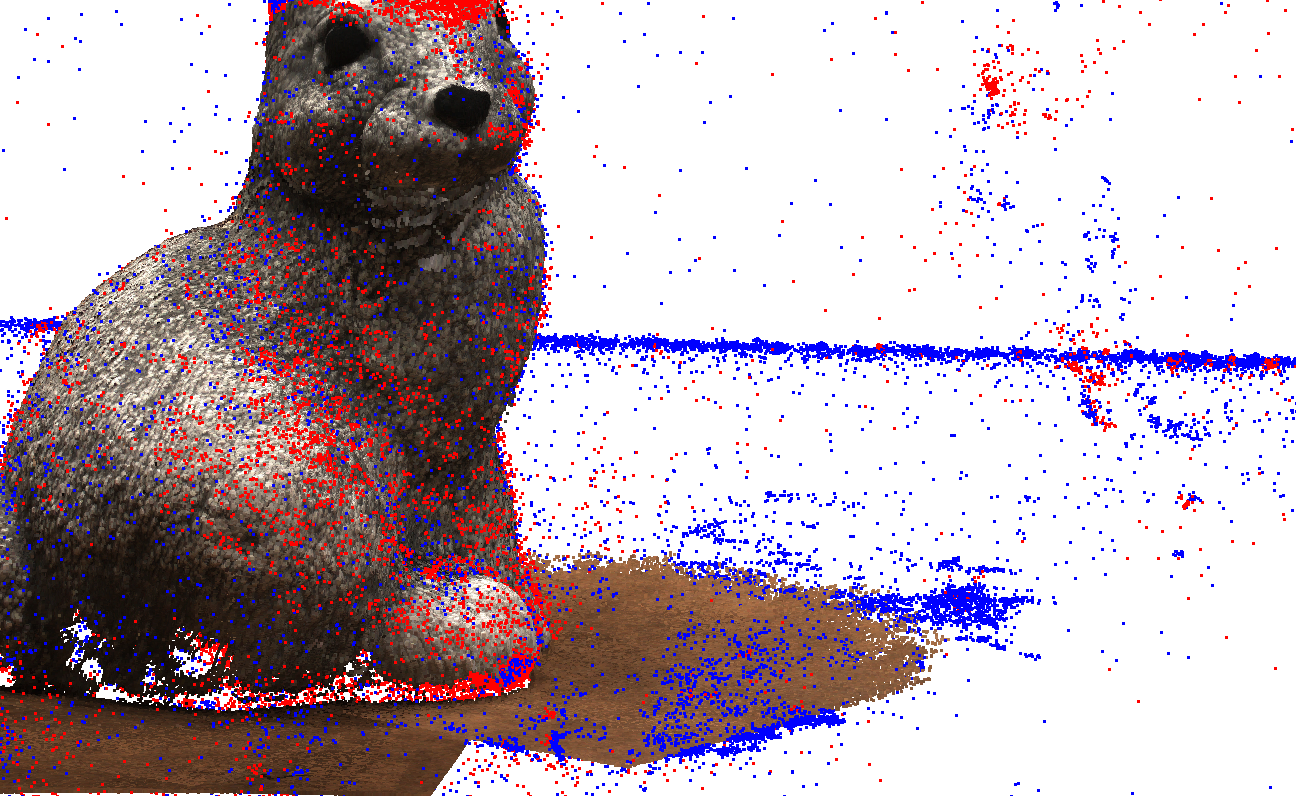} &
\includegraphics[width=0.2\linewidth]{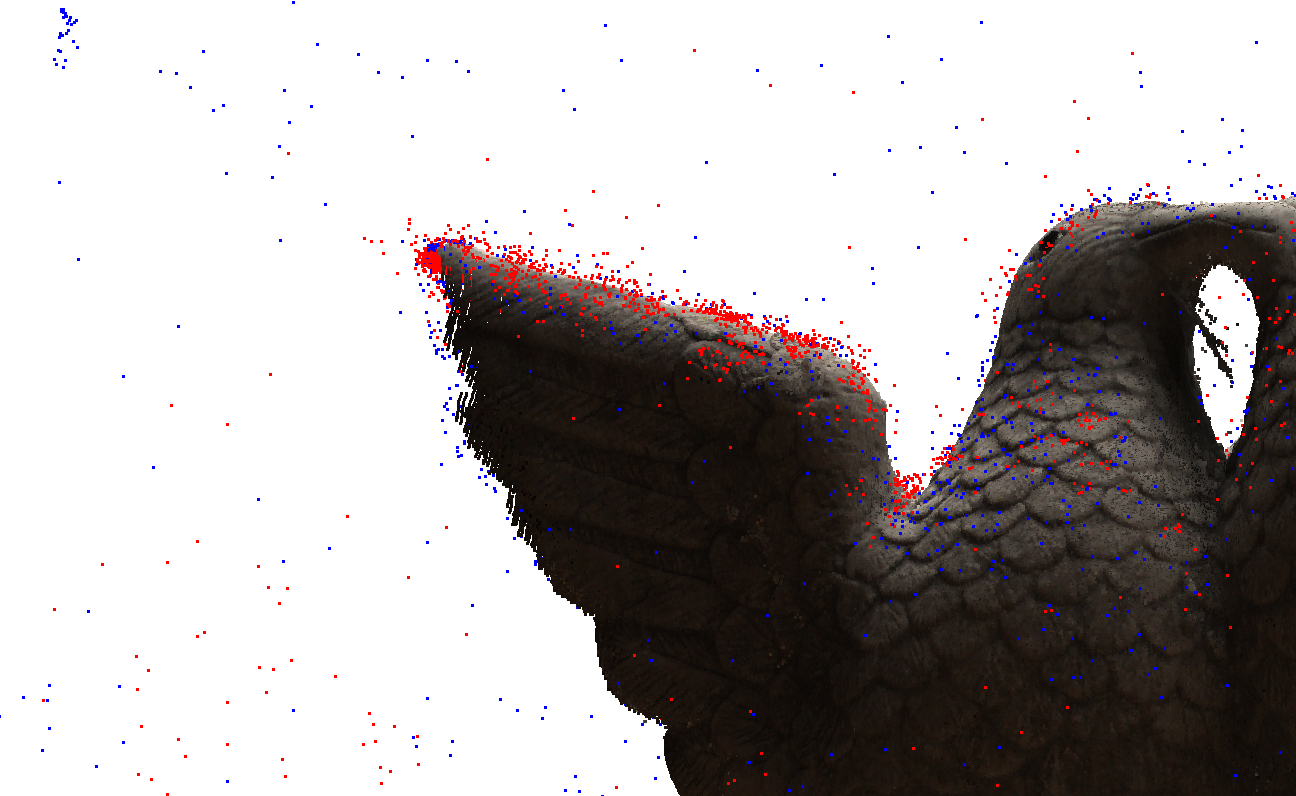} &
\includegraphics[width=0.2\linewidth]{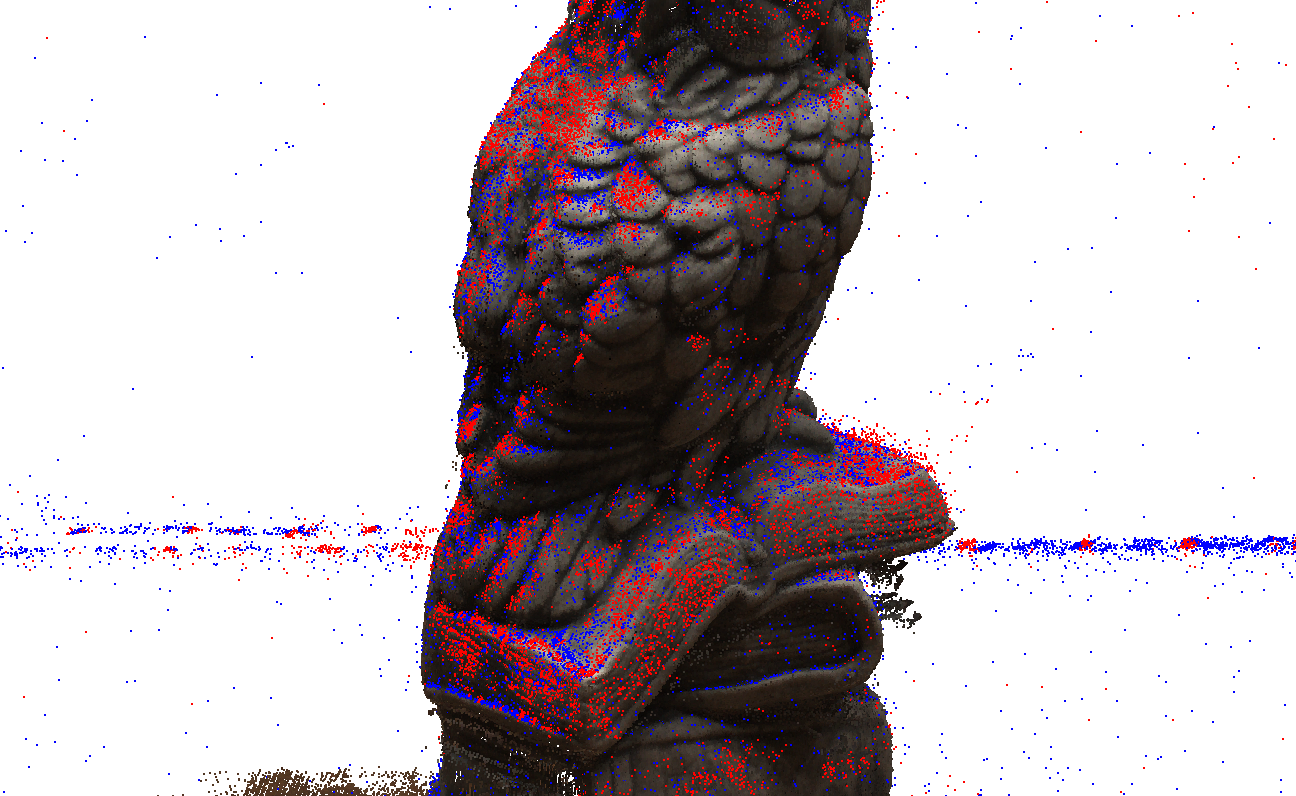} \\[2ex]
\end{tabular}
\caption{Comparison of Eigenentropy of points with a cloud-to-cloud distance (C2C) over 1mm to the reference point cloud. Blue points are below Eigenentropy $E=0.95$, red points above Eigenentropy $E=0.95$ and therefore candidates for pruning.}
\label{fig:Eigenentropy_artifacts}
\end{figure*}

\begin{table}[h!]
\centering
\small
\resizebox{\textwidth}{!}{
\begin{tabular}{lcccc}
\hline
Scene & scan40 & scan55 & scan106 & scan122 \\
\hline
Mean Eigenentropy (Gaussians, >1\,mm distance) & 0.9863 & 0.9617 & 0.9594 & 0.9531\\
\hline
\end{tabular}
}
\caption{Mean Eigenentropy of Gaussians in 3DGS corresponding to points with cloud-to-cloud distance over 1\,mm distance to the reference. Values guide the selection of the upper Eigenentropy threshold for pruning.}
\label{tab:mean_eigenentropy_outliers}
\end{table}
Table~\ref{tab:mean_eigenentropy_outliers} reports the mean Eigenentropy of outlier Gaussians across selected scans, showing values around Eigenentropy \(E=0.95\), which motivates the selection of an upper Eigenentropy threshold 
These observation is consistent with the theoretical characteristics of Eigenentropy discussed in Section~\ref{sec:geometry_feature_characteristics}. Specifically, Gaussians on well-structured surfaces typically correspond to near-planar distributions, where \(E \leq \log 2 \approx 0.693\), whereas outliers exhibit higher Eigenentropy, approaching values near 1, in line with locally isotropic or unstructured regions. Figure \ref{fig:Eigenentropy_artifacts} further visualizes these outlier Gaussians based on the choosed threshold: Gaussians exceeding the threshold \(E=0.95\) are marked in red and are considered candidates for pruning, while points below the threshold are shown in blue.

\section{Discussion}\label{sec:Discussion}

The proposed method, EntON, demonstrates clear superiority in terms of geometric reconstruction accuracy compared to 3D Gaussian Splatting (3DGS), while maintaining competitive rendering quality. These improvements are achieved through a targeted, Eigenentropy-aware densification and pruning strategy using local geometric information, that effectively exploits local geometric structure of Guassians under the Manhattan-World assumption predominant in man-made environments.
\paragraph{Eigenentropy Distribution}
A key insight provided by the experiments is the pronounced and systematic reduction of mean Eigenentropy across local Gaussian neighborhoods. EntON consistently achieves significantly lower final mean Eigenentropy values compared to the baseline. While 3DGS stabilizes at high Eigenentropy levels ($\sim$0.96) and even exhibits a slight increase over the course of training, EntON initiates a sharp decline shortly after the introduction of Eigenentropy-guided densification (around iteration 3000) and converges to a stable range of approximately 0.78 to 0.82. This behavior confirms that the proposed adaptive strategy successfully promotes lower-Eigenentropy, more ordered local neighborhoods.
Importantly, this reduction in Eigenentropy shows a strong correlation with improved geometric accuracy. Scenes and regions characterized by lower mean Eigenentropy consistently exhibit reduced Cloud-to-Cloud (C2C) errors, whereas high-Eigenentropy neighborhoods are associated with markedly larger geometric deviations. 
The qualitative results (Figure~\ref{fig:Eigenentropy_vs_C2C}) further reveal that EntON selectively achieves low Eigenentropy, particularly in structured, predominantly planar neighborhoods, precisely the areas where accurate surface representation is most critical under the Manhattan-World assumption.
Consequently, our strategy preferentially increases Gaussian density in these low-Eigenentropy (ordered, anisotropic, flat) areas by splitting to better capture fine geometric details on the object surface, while simultaneously pruning Gaussians in high-Eigenentropy, disordered, or spherically local neighborhoods. In contrast, 3DGS tends to retain, or even accumulate, redundant Gaussians in less structured areas, resulting in less efficient scene representations and lower geometric accuracy.
In summary, these findings strongly confirm the suitability of Eigenentropy as an effective guide for the densification and pruning criteria. By explicitly guiding towards low-Eigenentropy (ordered, anisotropic, flat) neighborhoods and pruning high-Eigenentropy (disordered, isotropic, spherical/scattered) regions, EntON achieves a systematically more structured and geometrically accurate Gaussian distribution while preserving high rendering quality. The results thereby validate the central hypothesis: Eigenentropy-aware control enables targeted resource allocation during optimization and delivers clear gains in geometric accuracy. Particularly in environments dominated by predominant flat structures.

As discussed, the Eigenentropy-aware densification and pruning strategy systematically promotes Gaussians in low-Eigenentropy (ordered, anisotropic, flat) local neighborhoods while suppressing those in high-Eigenentropy (disordered) regions. This targeted adaptation not only reduces the overall mean Eigenentropy, but directly translates into measurable benefits across three central performance metrics: geometric accuracy, rendering quality, and consumption of memory and training time.

\paragraph{Geometric Accuracy}

In detail, with respect to geometric accuracy, EntON achieves an average cloud-to-cloud (C2C) distance of $1.03\,\mathrm{mm}$ across all scenes of the DTU dataset. This performance is very close to the current state-of-the-art method PGSR ($1.00\,\mathrm{mm}$) and substantially outperforms both 3DGS ($1.609\,\mathrm{mm}$) and 2DGS ($1.331\,\mathrm{mm}$).
A scene-wise analysis highlights the strengths and limitations of the proposed Eigenentropy-aware strategy. In well-textured, predominantly planar regions (e.g., scan40, scan55, scan106), EntON achieves accuracy nearly identical to that of PGSR. Similarly strong performance is observed for scenes containing rough or diffuse materials (e.g., scan83, scan105). In contrast, for scenes dominated by reflective surfaces (e.g., scan63, scan97, scan110), EntON exhibits a slight degradation in accuracy compared to PGSR.
This scene dependency reveals a fundamental limitation of the neighborhood-based and point-density-dependent pruning and densification strategy. The computation of Eigenentropy is based on the covariance matrix derived from a local Gaussian neighborhood. However, in specular or texture-poor regions, the initial density of Gaussians is typically significantly lower, as fewer Gaussians are required to represent homogeneous color regions. With a fixed neighborhood size~$k$, the same number of Gaussian neighbors spans a substantially larger spatial extent in such low-density regions. As a consequence, the resulting covariance matrix tends to become more spherical, leading to an increased Eigenentropy, even when the underlying surface is inherently planar or low-Eigenentropy. These areas are thus erroneously interpreted by the method as disordered or high-Eigenentropy. As a result, splitting is insufficiently triggered, or not triggered at all, since the condition is (not) satisfied. At the same time, Gaussians with high Eigenentropy are systematically favored for pruning, further exacerbating the issue.
Together, these effects lead to a persistently reduced Gaussian density precisely in specular or view-dependent regions, where a higher density would in fact be beneficial to better enforce multi-view consistency and to reduce geometric errors, particularly for reflective or glossy surfaces. This behavior explains the slight but measurable degradation in geometric accuracy observed in such scenes compared to methods such as PGSR.
Nevertheless, EntON achieves near state-of-the-art accuracy with a significantly reduced number of Gaussians. The method deliberately concentrates Gaussians in regions where they are most critical for accurate geometric reconstruction, namely in low-Eigenentropy, structured, and planar areas, where object surface structure is likely to be present.

\paragraph{Rendering Quality}
Regarding rendering quality, EntON attains a mean PSNR of 34.91 dB over the entire small-scale DTU dataset, thereby slightly surpassing 3DGS (34.84 dB). 
Regarding rendering quality on the small-scale DTU dataset, EntON attains a mean PSNR of 34.39 dB across all neighborhood sizes, slightly below 3DGS (34.84 dB), but outperforming 2DGS (32.54,dB) and PGSR (32.32,dB). On the large-scale TUM2TWIN dataset, EntON achieves a mean PSNR of 31.44 dB across all neighborhood sizes, surpassing 2DGS (28.95 dB) and PGSR (28.87 dB) while remaining competitive with 3DGS (31.68 dB).

In well-textured scans, EntON typically matches or slightly exceeds 3DGS performance, while in low-texture scenes a minor drop in PSNR can be observed. 
This scene dependent rendering behavior stems directly from the Eigenentropy-aware strategy: As discussed in the context of geometric accuracy, lower initial point density in low-texture regions leads to a more isotropic local covariance structure and consequently higher Eigenentropy. This results in reduced Gaussian density (limited splitting and increased pruning), forcing both geometry and photometric appearance to be represented with even fewer primitives. Consequently, such regions may experience a slight loss of high-frequency details, manifesting in a marginally lower PSNR.

In contrast, in well-textured and structured regions the method benefits substantially from the high Gaussian density achieved there. This enables efficient and precise representation of dominant features such as sharp edges, texture, and fine surface details. Consequently, such scenes frequently exhibit rendering quality that is comparable or even superior to 3DGS, despite a significantly lower total number of Gaussians.
Compared to other methods such as 2DGS or PGSR, EntON overall preserves remarkably good photorealistic quality. The efficient resource allocation, achieved by concentrating Gaussians in structurally and low-entropy regions, thus provides an excellent trade-off between rendering quality and expansion of the scene. The approach sacrifices only minimal perceptual quality in extremely low-texture/ low-density regions, while benefiting strongly from reduced complexity in the majority of the scenes.

\paragraph{Efficiency}
EntON significantly reduces the mean number of Gaussians required for scene representation to 215\,k (compared to 392\,k in 3DGS, 232\,k in PGSR, and 209\,k in 2DGS). At the same time, training time for 15\,000 iterations ranges from 9.87 to 10.66\,min, substantially lower than that using 3DGS (12.88\,min), 2DGS (18.11\,min), or PGSR (32.89\,min) on the small-scale DTU dataset. 
On the large-scale TUM2TWIN dataset, EntON reduces the number of Gaussians compared to baselines, with a mean of 2.36–2.51 million across all neighborhood sizes (compared to 3.33M in 3DGS, 2.77M in 2DGS, and 2.56M in PGSR). However, the absolute number of Gaussians remains substantially higher than in DTU, which makes neighborhood calculations more expensive and explains why training times (19.30 to 24.98\,min) are not as low as on DTU, despite still being considerably faster than 2DGS (24.05\,min) and PGSR (39.23\,min).
This efficiency gain is a direct consequence of the Eigenentropy-aware strategy: focused densification in low-Eigenentropy regions, where object surface information is predominantly concentrated (i.e., structured, predominantly planar areas), combined with the systematic removal of redundant Gaussians in disordered/ spherical neighborhoods. Overall this results in a geometrically meaningful and compact distribution of the Gaussians in the scene. By eliminating unnecessary primitives, the method not only reduces memory footprint but also accelerates the training process, as fewer parameters overall need to be optimized, in particularly in non-surface regions. Consequently, EntON achieves high geometric and rendering quality at high efficiency regarding memory and training time, which is a key advantage for scalable applications and resource-constrained environments.\newline

\paragraph{Limitations}
Despite the strong performance of EntON across a wide range of scenes, the proposed method exhibits two principal limitations.
First, EntON is particularly designed for man-made scenes like in urban environments and relies on the assumption that meaningful geometric structure can be inferred from local 3D shape of Gaussian neighborhoods under a Manhattan-World prior. This assumption is well justified for scenes dominated by nearly-planar, anisotropic structures such as walls, floors, and architectural elements. However, it becomes less appropriate for environments characterized by highly curved, irregular, or scattered geometry, such as vegetation or organic objects. In such scenes, local Gaussian neighborhoods are inherently more isotropic and less aligned with dominant object surface, making the distinction between low- and high-Eigenentropy neighborhoods less indicative of true surface relevance. As a result, the EntON may suppress Gaussians in this scenes that are necessary to faithfully represent complex non-planar geometry. Second, the method is sensitive to variations in local Gaussian density due to its use of a fixed neighborhood size $k$ for Eigenentropy computation. In regions with low Gaussian density, commonly occurring on reflective or texture-less surfaces, the same number of neighbors spans a substantially larger spatial extent. This leads to more spherical covariance estimates and higher Eigenentropy values, even when the underlying surface is locally planar. Consequently, such regions may be misdefined as disordered, resulting in insufficient densification and increased pruning.
These limitations highlight that EntON is best suited for structured, predominantly planar scenes with sufficiently dense Gaussian distributions, and they point toward adaptive neighborhood strategies and geometry–appearance coupling as promising directions for future work. 

In summary, the Eigenentropy-aware densification and pruning strategy EntON successfully promotes a more structured and surface-aligned distribution of Gaussians. By preferentially densifying in low-Eigenentropy neighborhoods (indicating ordered, anisotropic, flat regions) and pruning Gaussians in high-Eigenentropy, disordered regions, EntON achieves substantially lower Eigenentropy compared to 3DGS, as evidenced by the rapid and stable decline starting early in training. This targeted adaptation translates into clear performance benefits across the key dimensions: high geometric accuracy, competitive to slightly superior rendering quality (preserving fine details despite reduced complexity), and markedly improved consumption of memory and training time efficiency (reduction of number of Gaussians count to present the scene and shorter training time). The observed scene dependent limitations, particularly the slight degradation on reflective/specular surfaces, are attributable to the Gaussian spatial density sensitivity of the computed Eigenentropy from a fixed number of neigherst neighbors, which can lead to under-densification in regions with a low spatial density of Gaussians. Nevertheless, the overall results strongly validate Eigenentropy as an effective guide for local geometric ordered densification, enabling more intelligent resource allocation without additional supervision or complex constraints.

\section{Conclusion}\label{sec:Conclusion}

We present EntON, a geometry-guided, Eigenentropy-aware alternating densification strategy for 3D Gaussian Splatting that leverages local 3D structural geometry to guide adaptive splitting and pruning. Gaussians in low-Eigenentropy (ordered, flat) neighborhoods are preferentially densified, while those in high-Eigenentropy (disordered, spherical) neighborhoods are pruned. By explicitly exploiting the local 3D neighborhood of Gaussians, EntON aligns the densification process with geometric regularities commonly found in man-made structures and urban scenes.
By alternating between standard gradient-based and Eigenentropy-aware densification, EntON captures fine geometric details with compact, surface-aligned Gaussian representations, while avoiding unnecessary scene expansion of Gaussians in high-Eigenentropy areas. As a result, it delivers high geometric accuracy, while maintaining photometric fidelity. Surface-aligned densification and systematic pruning lead to fewer Gaussians, lower memory usage, and shorter training time compared to conventional 3DGS and other baseline methods.
Experiments on two benchmark datasets demonstrate that by guiding densification toward low-Eigenentropy regions, EntON enables 3D scene reconstructions that are geometrically and photometrically accurate, memory- and computation-efficient, and suitable for both small- and large-scale scenes.


\bibliographystyle{cas-model2-names}
\bibliography{authors.bib}

\end{document}